\documentclass[letterpaper]{article}
\usepackage[draft]{aaai2026} %
\usepackage{times} %
\usepackage{helvet} %
\usepackage{courier} %
\usepackage[hyphens]{url} %
\usepackage{graphicx} %
\usepackage{graphicx} %
\usepackage{natbib} %
\usepackage{caption} %
\usepackage[utf8]{inputenc} %
\usepackage[T1]{fontenc}    %
\usepackage{nameref}
\usepackage{url}            %
\usepackage{booktabs}       %
\usepackage{amsfonts}       %
\usepackage{nicefrac}       %
\usepackage{microtype}      %
\usepackage{xcolor}         %

\usepackage{adjustbox}

\usepackage{subcaption}
\usepackage{graphicx}
\usepackage{todonotes}
\usepackage{algorithm}
\usepackage{algorithmicx}
\usepackage{algpseudocode}
\usepackage{amsmath}
\usepackage{amssymb}
\usepackage{caption}
\usepackage{amsthm}
\usepackage{thmtools}
\theoremstyle{definition}

\theoremstyle{remark}

\newcommand{\norm}[1]{\left\lVert#1\right\rVert}

\newcommand*{\boldone}{\text{\usefont{U}{bbold}{m}{n}1}}

\usepackage{minted}
\usepackage{amsmath}
\usepackage{tikz}
\usepackage{mathdots}
\usepackage{yhmath}
\usepackage{cancel}
\usepackage{color}
\usepackage{siunitx}
\usepackage{array}
\usepackage{multirow}
\usepackage{amssymb}
\usepackage{gensymb}
\usepackage{tabularx}
\usepackage{extarrows}
\usepackage{booktabs}
\usetikzlibrary{fadings}
\usetikzlibrary{patterns}
\usetikzlibrary{shadows.blur}
\usetikzlibrary{shapes}
\usepackage{environ}
\makeatletter
\newsavebox{\measure@tikzpicture}
\NewEnviron{scaletikzpicturetowidth}[1]{%
  \def\tikz@width{#1}%
  \begin{lrbox}{\measure@tikzpicture}%
  \BODY
  \end{lrbox}%
  \pgfmathparse{#1/\wd\measure@tikzpicture}%
  \BODY
}
\makeatother

\title{A Vector Symbolic Approach to Multiple Instance Learning}

\author {
    Ehsan Ahmed Dhrubo \textsuperscript{\rm 1},
    Mohammad Mahmudul Alam \textsuperscript{\rm 2},
    Edward Raff\textsuperscript{\rm 1,3},
    Tim Oates\textsuperscript{\rm 1},
    James Holt \textsuperscript{\rm 4}
}
\affiliations {
    \textsuperscript{\rm 1}University of Maryland, Baltimore County\\
    \textsuperscript{\rm 2}North South University\\
    \textsuperscript{\rm 3}CrowdStrike\\
    \textsuperscript{\rm 4}Datalytica\\
}

\begin{document}

\maketitle

\begin{abstract}
Multiple Instance Learning (MIL) tasks impose a strict logical constraint: a bag is labeled positive if and only if at least one instance within it is positive. While this iff constraint aligns with many real-world applications, recent work has shown that most deep learning-based MIL approaches violate it, leading to inflated performance metrics and poor generalization. We propose a novel MIL framework based on Vector Symbolic Architectures (VSAs), which provide a differentiable mechanism for performing symbolic operations in high-dimensional space. Our method encodes the MIL assumption directly into the model’s structure by representing instances and concepts as nearly orthogonal high-dimensional vectors and using algebraic operations to enforce the iff constraint during classification. To bridge the gap between raw data and VSA representations, we design a learned encoder that transforms input instances into VSA-compatible vectors while preserving key distributional properties. Our approach, which includes a VSA-driven MaxNetwork classifier, achieves state-of-the-art results \textit{for a valid MIL model} on standard MIL benchmarks and medical imaging datasets, outperforming existing methods while maintaining strict adherence to the MIL formulation. This work offers a principled, interpretable, and effective alternative to existing MIL approaches that rely on learned heuristics.

\end{abstract}

\section{Introduction}

Multiple Instance Learning (MIL) is an effective tool for many tasks because it imposes useful constraints on the model that maps well to real-world systems. In particular, a dataset $\mathcal{X}$ in the MIL formulation is composed of $N$ \textit{bags} $X_i$, where each bag contains $n_i$ different instances $\mathbf{x}_1, \mathbf{x}_2, \ldots, \mathbf{x}_{n_i}$ such that $\mathbf{x}_j \in \mathbb{R}^d$. The term bag is used to denote that the variable number of items $n_i$ are unordered (i.e., not a sequence) and that duplicates are allowed (i.e., not a set). The MIL problem is a binary classification task with label  $y_i \in \{-1, 1\}$. The useful constraint of the MIL problem is that given an instance-level classification $h(\mathbf{x}_j) = \hat{y}$, a bag-level classifier $g(X_i)$ should \textit{return a positive label if and only if there is at least one positive instance inside the bag}, that is to say: $g(X_i) = 1$ if and only if there exists a $h(x_j) = 1 \quad \forall x_j \in X_i$.

This constraint on the classifier necessarily reduces expressive power, but better aligns with many real-world use cases. For example, a mammogram is indicative of cancer if and only if cancerous cells are found. This makes the MIL model attractive to many similar tasks because it avoids spurious failure conditions (i.e., learning to default to positive and find negative indicators) that are physically or biologically impossible in a given context. Yet, as recently shown, almost all neural network approaches to MIL do not properly encode or respect this constraint \cite{raff_reproducibility_2023,jang_are_2024,liu_position_2025} and have over-inflated performance because they get to ``cheat'' by ignoring the iff constraint. 

Thus, there is currently a lack of deep learning approaches to the MIL problem that encode/abide by these constraints. Most rely on a relatively straightforward method from the 1990s in producing a linear response against the instance features, e.g., $h(\mathbf{x}):=\mathbf{x}^\top \mathbf{w}$, and taking the maximal response over all instances. Our work asks if there are alternative ways of designing the instance-level classifier that allow us to reason about the MIL constraints and improve performance.

By leveraging a Vector Symbolic Architecture (VSA)\footnote{Also termed hyper-dimensional computing or (HDC)}, we can answer this in the affirmative. VSAs provide binding/unbinding operations to manipulate vectors in such a way that one can encode first-order logic operations. Because these operations are differentiable, they are amenable to learning in a deep architecture. However, VSAs naturally work with random vectors as their inputs, where ``concepts'' are assigned to arbitrary vectors --- which would seemingly preclude use with dataset inputs with an unknown distribution. 
By encoding the MIL logic using VSAs, we get an approach that respects the MIL hypothesis by construction.\footnote{This should not be conflated to say that any such approach is perfect, and indeed noise is still a factor in the model and data.} We combine this with a novel approach to convert dataset inputs into a VSA-compatible representation, so that the symbolic manipulation properties of the VSA can be maintained. 

The rest of this article is organized as follows. We will review related work in Section\textit{~\nameref{sec:related_work}}, followed by a brief primer on VSAs for unfamiliar readers in Section\textit{~\nameref{sec:background}}. Then we will detail how VSAs can symbolically maintain the MIL assumptions, and our approach to convert instances into VSA-compatible representations in Section\textit{~\nameref{sec:methodology}}. After detailing our full architecture, we provide experiments on traditional MIL and medical imaging MIL problems to demonstrate our approach and its increased performance despite sticking to the MIL constraints in Section\textit{~\nameref{sec:results}}. Finally, we conclude in Section\textit{~\nameref{sec:conclusion}}. 

\section{Related Work}      \label{sec:related_work}

The primary impedes for or work is the results from \citet{raff_reproducibility_2023}, who demonstrated that seminal works like  \texttt{MIL pooling} ~\citet{pmlr-v80-ilse18a}, the Graph Neural Network based \texttt{GNN-MIL} ~\cite{tu_multiple_2019}, Transformer based \texttt{TransMIL} \cite{shao_transmil_2021}, and \texttt{Hopfield} MIL model of ~\cite{widrich_modern_2020} in fact perform \textit{set} classification, because they do not have any method to enforce the MIL framework. These approaches are foundational to most all modern ``MIL'' models, with the shared common error of an un-constrained non-linear layer after a pooling operation (i.e., variable sized bag to fixed length representation), which allows the model to learn invalid solutions. This is still true for MIL in sensitive domains like image classification \cite{attention,image_classification,multi_head_attention}, image segmentation \cite{image_segmentation}, and pathology \cite{survey_recent_papers,breast_cancer_highest_accuracy,DSMIL}. Many recently proposed MIL algorithms appear to have the issue of an unconstrained final layer~\cite{thandiackal_differentiable_2022,pal_bag_2022}, but nuanced inspection of each is time intensive and arbitrary, our interest is in designing new strategies to encode the MIL constraints in an easy-to-understand way. Other works have extended MIL to scenarios past the basic one we consider in this work like time-series and instance interactions, and appear to have done so without error~\cite{timemil,xmil}. We consider these beyond the scope of our article as we focus on more standard and simple MIL model, which now lacks modern techniques.

\subsection{MIL History}
Early approaches to MIL focused on the standard MIL assumption and constructed models that explicitly searched for discriminative instances. Classical algorithms such as axis-aligned rectangles \citep{dietterich1997solving}, diverse density \citep{maron1997framework}, and EM-DD \citep{zhang2001em} exemplify this view, modeling bag labels based on the presence of one or more key instances. These approaches were often derived by adapting single-instance algorithms, such as support vector machines \cite{NIPS2002_2232}, neural networks \citep{ramon2000multi}, and decision trees \citep{blockeel2005experimental}, to incorporate latent instance-level labels and constraints consistent with the standard assumption.

As MIL was applied to more diverse domains, including image and text classification, researchers proposed more flexible modeling assumptions. Weidmann et al.~\citep{weidmann2003two} introduced a hierarchy of presence-based, threshold-based, and count-based assumptions. Distance-based and embedding approaches—such as GMIL \citep{scott2005generalized}, DD-SVM \citep{chen2004kernel}, and MILES \citep{chen2006miles}—modeled bags using similarity to instance prototypes. The collective assumption \citep{xu2003statistical} treated bags as distributions over instances and computed predictions by aggregating instance-level probabilities. Later extensions introduced instance weighting schemes \citep{foulds2008learning}, graph-based bag structures \citep{zhou2009multi}, and propositionalization methods \citep{zhou2007multi}, further generalizing the MIL framework. 
\subsection{Vector Symbolic Architectures}
VSAs, also known as Hyperdimensional Computing (HDC), are a class of computational models that represent and manipulate data using high-dimensional distributed vectors called \emph{hypervectors}. VSAs aim to unify the strengths of symbolic representations (systematicity, compositionality, and expressiveness) with the robustness and similarity-based reasoning of distributed connectionist representations. They address challenges such as the “superposition catastrophe” \cite{rachkovskij2001binding},
and compositionality limitations \cite{jackendoff2002foundations} by employing algebraic operations that encode structure in a way that preserves semantic similarity. 

These operations include {\em binding} and {\em unbinding}.  Let $\mathbf{a}, \mathbf{b} \in \mathbb{R}^D$ be vectors of dimension $D$.  Binding is an operation $\circ$ that combines two vectors $\mathbf{c} = \mathbf{a} \circ \mathbf{b}$, and unbinding is the inverse operation $\oslash$ such that $\mathbf{a} = \mathbf{c} \oslash \mathbf{b} \quad \text{or} \quad \mathbf{b} = \mathbf{c} \oslash \mathbf{a}$.  
Different VSAs use different implementations of these two operations.

The VSA family encompasses a range of models that differ in representation space and operations. Tensor Product Representations (TPRs) use outer products for binding, resulting in growing dimensionality \cite{smolensky1990tensor}. Holographic Reduced Representations (HRRs) apply circular convolution \cite{plate1995holographic}, while Binary Spatter Codes (BSCs) use XOR \cite{kanerva1997fully}, and Multiply-Add-Permute (MAP) employs multiplicative binding with addition and permutation \cite{gayler1998multiplicative}. Other approaches include Sparse Binary Distributed Representations \cite{rachkovskij2001binding} and Modular Composite Representations \cite{snaider2014modular}. Many works have looked at ways to generically compare the pros/cons of different VSAs or improve core patterns~\cite{liu_linearithmic_2025,carzaniga_practical_2025}, which are beyond the scope of this article. 

These models enable compact, fixed-width vector representations of structured data with desirable properties such as similarity preservation, robustness to noise, and algebraic composability. Their ability to represent an exponential number of quasi-orthogonal symbols in fixed dimensionality, sometimes referred to as the “blessing of dimensionality” \cite{gorban2018blessing}, supports scalable learning and reasoning. The field, increasingly relevant to both symbolic AI and neuromorphic engineering, has seen applications in language processing \cite{sahlgren2008semantic}, cognitive modeling \cite{eliasmith2012large}, and unconventional computing architectures \cite{kleyko2021computing}.

The hybrid neuro-symbolic nature of VSAs and HDCs has been used to produce a long history of bespoke models to tackle specific issues, enabled by their symbolic manipulations. 
On the more symbolic AI side they have been used to solve problems like Raven's progressive matrices~\cite{hersche_neuro-vector-symbolic_2023}, subitization~\cite{alam_towards_2023}, and planning/path-finding~\cite{mcdonald_assembling_2024,stockl_local_2024,mcdonald_aspects_2021}. In general, VSAs have served as the basis for construction architectures that yield to theoretical analysis more easily~\cite{Frady2018}.
One line of history involves using VSA within other classical neural network architectures to provide mechanisms to alleviate specific limitations. This has been done for general architectures like CNNs~\cite{alam_holographic_2024}, RNNs~\cite{Danihelka2016,10.5555/2987061.2987066,saul_lempel-ziv_2022,tay_learning_2017} and Transformers~\cite{alam_recasting_2023} and to reduce model size in n-gram models~\cite{alonso_hyperembed_2021} and Extreme Multi-Label Classification~\cite{Ganesan2021}. 
Most relevant to our work thematically is using VSAs to design classifiers to have specific properties, such as in speech recognition ~\cite{Imani2017} for graph classification~\cite{nunes_graphhd_2022}, privacy~\cite{Alam2022}, unsupervised learning~\cite{osipov_hyperseed_2024}, set similarity~\cite{nunes_dothash_2023}. While these examples are not for the MIL problem, they share the same high-level goal: exploiting the symbolic structure of VSAs to design networks with known behavioral properties, in our case, respecting the MIL hypothesis.

\section{VSA Primer} \label{sec:background}

As VSAs are not widely known in the community today, we give a brief primer on them as relevant to our work. VSA is a computational framework that represents symbols using high-dimensional vectors capable of performing symbolic operations such as \emph{Binding} to combine concepts, \emph{Unbinding} to extract concepts, \emph{Bundling} to aggregate information, and more \cite{kanerva2009hyperdimensional}. These operations preserve the similarity of original components in a complex representation and support noise-tolerant computation. As such, VSAs are well-suited for representing a bag of instances in a high-dimensional vector using Bundling. Moreover, VSAs are resilient to noise and missing data, which are common in weakly labeled settings like MIL \cite{kleyko2021computing}.
\par 
The core operations of each VSA, Binding and Unbinding vary in definition, yet their underlying conceptual framework remains consistent. For instance, to construct a composite representation from a set of vector pairs $(x_n, y_n)$ using a Binding function $\mathcal{B}$, one can compute $\chi = \sum_{i=1}^{N} \mathcal{B}(x_i, y_i)$ for $N$ such pairs. Subsequently, the Unbinding function $\mathcal{B}^{*}$ can be applied with $y_i$ to retrieve the associated component via $\hat{x_i} = \mathcal{B}^{*}(\chi, y_i)$. The original and retrieved vectors, $x_i$ and $\hat{x_i}$ respectively, may then be compared using the inner product or cosine similarity to assess the presence and fidelity of the component within the composite representation.
\par  
The specific definitions of Binding and Unbinding differ across various VSAs. HRR \cite{plate1995holographic} employs circular convolution for these operations. Vector-derived Transformation Binding (VTB) \cite{vtb} utilizes matrix transformations. More recently, Hadamard-derived Linear Binding (HLB) \cite{hlb} introduced a linear VSA framework where Binding and Unbinding are performed through element-wise multiplication and division, respectively. This method leverages a specially designed initialization constraint known as the Mixture of Normal Distribution (MiND), defined in Equation~\ref{eq:init_cdn}. In this initialization scheme, a vector $x \in \mathbb{R}^d$ has an expected value of zero, i.e., $\mathrm{E}[x] = 0$, while their expected absolute mean is non-zero, i.e., $\mathrm{E}[|x|] = \mu$. The $L^2$ norm of the vectors is given by $\norm{x}_2 = \sqrt{\mu^2 d}$. Here, $\mathcal{N}$ and $\mathcal{U}$ denote the Normal and Uniform distributions, respectively, and $\mu$ represents the mean of the distribution.

\begin{equation} \label{eq:init_cdn}
\Omega(\mu, 1 / d) = \begin{cases}
\mathcal{N}(-\mu, 1 / d)  & \quad \textrm{if \quad} \mathcal{U}(0, 1) > 0.5 \\
\mathcal{N}(\ \mu, 1 / d) & \quad \text{else } \mathcal{U}(0, 1) \leq 0.5 \\
\end{cases}
\end{equation}

\section{Methodology} \label{sec:methodology}

\begin{figure}[!h]
    \centering
    \includegraphics[width=0.99\columnwidth]{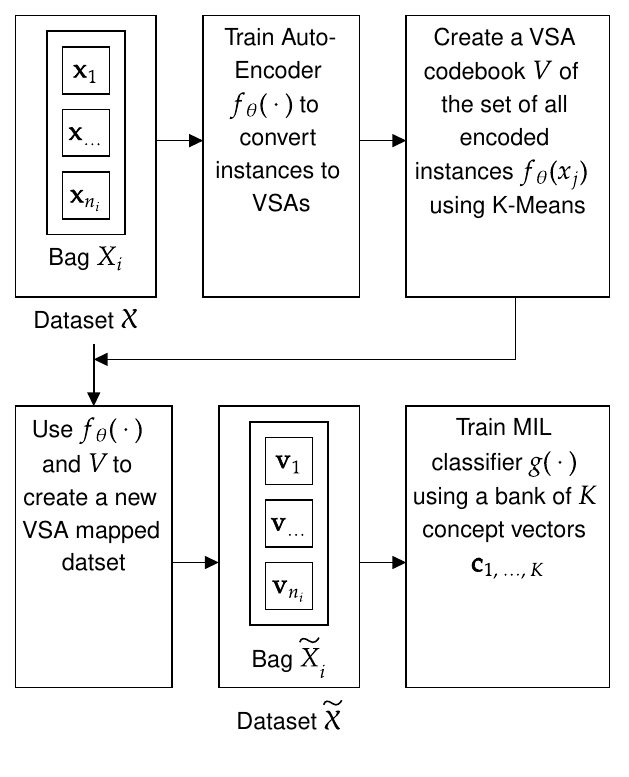}

    \caption{The overarching approach to VSA-MIL is presented in this flow. The dataset $\mathcal{X}$ is not a VSA, so it first goes through an Auto-Encoder to incentivize a representation that produces outputs that maintain the necessary conditions of the VSA in a semantically meaningful way. Clustering discretizes the continuous space via a codebook so that exact matches can occur. Then a VSA informed network can be used to train on the VSA compatible representation $\mathbf{v}_i$. }
    \label{fig:diagram}
\end{figure}

We will now describe our new approach for a MIL classifier, which has a high-level outline in Figure~\ref{fig:diagram}. First, we will detail how the VSA allows one to naturally frame the MIL problem given valid VSA vectors. For this work, we use the HLB vector~\cite{hlb}. However, an input dataset does not naturally map to VSAs, so we introduce an autoencoder approach to penalize a representation from diverging from the sufficient conditions for the VSA to operate correctly. Then we detail the full VSA-MIL algorithm. 

\subsection{Using VSAs to Perform MIL Classification}

The VSA approach enables the construction of a MIL model that inherently adheres to the fundamental prior of MIL models, thereby addressing the issues highlighted by \citet{raff_reproducibility_2023} , which suggests that recent deep learning literature has not adequately preserved the one-way relationship (i.e., the ability to learn negative classes).  We show below how to write the MIL problem in a simple first-order logic of VSA vectors, and thus respects the MIL constraint by construction. 

Suppose that there is a function $f(\cdot)$ that converts a bag $X$ of instances $\mathbf{x}_i$ into a set of valid VSA vectors $\mathbf{v}_i$.  Then there is a direct VSA-based method of implementing the MIL constraints that a positive prediction of $y=1$ should occur if and only if a positive indicator is observed, and return the negative prediction $y = -1$ otherwise.  That method is described next, and the function $f(\cdot)$ is defined thereafter.

Given $n$ VSA vectors ($\mathbf{v_1}, \mathbf{v_2}, \ldots , \mathbf{v_n}$) with $d$ dimensions,  a compressed "bag" representation of the VSAs can be computed as  $\mathbf{s} = \sum_{i=1}^{n} \mathbf{v_i}$. This will be our primary representation of the input. Now, let us assume that we have $K$ different "concepts", $\mathbf{c}_1, \ldots, \mathbf{c}_K$, that are VSAs from the same distribution as $\mathbf{v}_{1, \ldots, n}$ and are the positive indicators for which a MIL classifier should return a positive label. We wish to check if any of the $K$ concepts are a match to any of the $n$ VSAs, i.e., $\mathbf{c}_k \in \mathbf{s} \forall k \in [1, K]$. This check requires computing $\mathbf{c}_k^\top \mathbf{s}$, and we will show how this results in a positive or zero response for the present/absent case. 

First, we will assume that $\mathbf{c}_k$ is present inside of $\mathbf{s}$. Then the result will become:

\begin{equation}
     \mathbf{c_k}^{\top}\mathbf{s} = \sum_{j=1}^{n} \mathbf{c_k}^{\top} \mathbf{v_j}  = \norm{\mathbf{c_k}}_2^{2} + \sum_{j=1, \\ j\neq k}^{n} \mathbf{c_k}^{\top} \mathbf{v_j} \label{eq:1}
\end{equation}

Since the VSAs are nearly orthogonal to each other (as they are in expectation by construction), the second term will be zero for expectation. The equation \ref{eq:1} will become 

\begin{equation}
    \mathbb{E}[\mathbf{c_k}^{\top}\mathbf{s}] = \mathbb{E}[\norm{\mathbf{c_k}}_2^{2}] \label{eq:2}
\end{equation}

For HLB architecture we use, $\norm{\mathbf{c_k}}_2 = \sqrt{\mu^2d}$, where $\mu$ is the pre-specified absolute mean of the distribution that VSA vectors are sampled from. Simplifying this gives

\begin{equation}
    \mathbb{E}[\norm{\mathbf{c_k}}_2^{2}] = \mathbb{V}(\norm{\mathbf{c_k}}_2) + {\mathbb{E}[\norm{\mathbf{c_k}}_2]}^2 = 0 +  \mathbb{E}[\sqrt{\mu^2d}]^2 = \mu^2d  \label{eq:3}
\end{equation}

Thus, we know that the value that should be returned is a positive value, and the VSA's initial distribution determines the expected value precisely. 

If a VSA $\mathbf{c_k}$ is not in the $\mathbf{s}$, have the simple summation:

\begin{equation}
     \mathbf{c_k}^{\top}\mathbf{s} = \sum_{j=1, \\ \mathbf{c_k} \notin \mathbf{s}}^{n} \mathbf{c_k}^{\top} \mathbf{v_j} \label{eq:5}
\end{equation}

If we take the expectation of $\mathbf{c_k}^{\top}\mathbf{s}$, all terms will be in expectation zero given that all the VSAs in $\mathbf{s}$ are nearly orthogonal to $\mathbf{c_k}$. So, the equation \ref{eq:5} becomes $\mathbb{E}[\mathbf{c_k}^{\top}\mathbf{s}] = 0, \mathbf{c_k} \notin \mathbf{s} $. Then, if we wish to have a classification head $g(\cdot)$ that processes a bag of VSAs, the instance-level classifier $h(\cdot)$ is as simple as having a bias term $b$ to calculate $h(\mathbf{s}) = \max_k \mathbf{c}_k^\top \mathbf{s} $, which can be made computationally efficient by storing all $\mathbf{c}_k$ in a matrix and reducing over a matrix-vector product. Then the classification is $g(\mathbf{s}) = \boldone[h(\mathbf{s})  >b]$. This ensures that, similar to max-pooling, we retain a valid MIL prediction. But by incorporating multiple possible positive ``concepts'' via the VSA, we have greater flexibility in the output space than simple max-pooling, without succumbing to the error of using an unconstrained output layer that can invert the decision process \cite{jang_are_2024}.

\subsection{Converting Arbitrary Inputs into VSAs}

Now we must deal with the issue that VSAs are historically designed to work on vectors of stochastic origin, where semantic meaning is assigned to arbitrary vectors. However, our inputs already exist in some $d$-dimensional space and likely some assumed lower-dimensional manifold. This is the case for datasets that come as a ``bag'' of vectors naturally or in computer vision tasks where sub-regions of the image are treated as instances. Normal VSAs use the stochastic values to leverage the fact that, in high-dimensional space, random vectors are approximately orthogonal on average. 

Thus, we must determine a way to convert arbitrary and semantically meaningful inputs $\mathbf{x}_i$ into VSA-compatible vectors $\mathbf{v}_i$. Our key insight is that VSAs use stochastic sampling in order to satisfy sufficient conditions for the properties like quasi-orthogonality to hold, the specific random sampling is not a prerequisite in and of itself. For the HLB VSA in particular, quasi-orthogonality is not even a necessary condition. By training an auto-encoder that penalizes deviation from the sufficient conditions of the HLB vectors in the middle layer of an auto-encoder, we can learn a model $f_\theta(\cdot)$ to learn a conversion from input-to-VSA. 

In the case of the HLB VSA, there are three properties that are used to derive the binding and unbinding properties. Given an arbitrary dimension $k \in [1, d]$, these properties are:
(1) $\mathbb{E}[v_k] = 0$, (2) $\mathbb{E}[|v_k|] = \mu$, and (3) $\mathbb{E}[\|\mathbf{v}\|_2] = \sqrt{\mu^2 d}$. 

The original article obtained these properties by sampling from a bimodal normal distribution $v_k \sim \{\mathcal{N}(-\mu, 1/d),~\mathcal{N}(\mu, 1/d)\}$. Instead, we define an encoder $f_\theta(\cdot)$ as a neural network with a corresponding decoder $f^{-1}_{\theta'}(\cdot)$. Combined, the loss of our auto-encoder becomes Equation~\ref{eq:hlb_autoencoder_loss} using a target response of $\mu = 1/2$, and would be averaged over a batch of samples. We do not find any hyperparameter tuning necessary in our approach, but it could be added if desired. 

\begin{align} \label{eq:hlb_autoencoder_loss}
\begin{split}
\underbrace{\left\|\mathbf{x}_i - f^{-1}_{\theta'}(f_\theta(\mathbf{x}_i))\right\|_2}_{\text{Reconstruction}} + 
\underbrace{\frac{1}{d} \sum_{j=1}^d f_\theta(\mathbf{x}_i)_j}_{\text{HLB Property 1, } \mathbb{E}[v_k] = 0} + \\
\underbrace{\left(\frac{1}{2}-\frac{1}{d} \sum_{j=1}^d |f_\theta(\mathbf{x}_i)_j|\right)^2}_{\text{HLB Property 2, } \mathbb{E}[|v_k|] = \mu}+
\underbrace{\left(\|f_\theta(\mathbf{x}_i)\| - \sqrt{d/4}\right)^2}_{\text{HLB Property 3, } \mathbb{E}[\|\mathbf{v}\|_2] = \sqrt{\mu^2 d}}
\end{split}
\end{align}

In this way, we learn a model that should learn a useful hidden representation (via the first reconstruction term) while also producing vectors that should approximately behave like VSA vectors (via the three additional penalties). Although this cannot enforce perfect HLB correspondence, the VSA process is inherently noisy due to the multiple vectors bound to a finite-dimensional space. Our empirical results will demonstrate that this works in practice to obtain good accuracy and does not fail empirical tests of the MIL properties.

\subsection{VSA-MIL}

We now have the full components to describe the details of our approach as outlined in Figure~\ref{fig:diagram}, with more details in Algorithm~\ref{algo:vsa-mil}. Given a dataset $\mathcal{X}$ we train an auto-encoder using ReZero \cite{rezero} skip connections, such that it learns a mapping $f_\theta(\cdot)$ to convert vectors into VSA-compatible representations. Because we wish to be able to have strong ``present/absent'' responses, we further discretize the continuous output space by applying k-Means clustering to the encoded dataset. This will produce a new finite set of $k$ VSA vectors $V \in \mathbb{R}^{d,k}$ which is used as a codebook. Any bag that comes in is first translated at an instance level via $f_\theta$ and then replaced with its nearest exemplar in $V$.  A pooling step is used for the final predictor $g(\cdot)$ which has a bias term to minimize the binary cross entropy loss $\ell(\cdot, \cdot)$.

\begin{algorithm}[!h]
\caption{VSA-MIL Algorithm} \label{algo:vsa-mil}
\begin{algorithmic}[1]
\Require Dataset $\mathcal{X} = \{\mathbf{x}_i\}_{i=1}^N$, batch size $B$, number of clusters $K$
\State Train an encoder $f_\theta(\cdot)$ and decoder $f^{-1}_{\theta'}(\cdot)$ on batches of size $B$ using reconstruction Equation~\ref{eq:hlb_autoencoder_loss}
\State Encode the full dataset: $\tilde{\mathbf{v}}_i = f_\theta(\mathbf{x}_i), \quad \forall \mathbf{x}_i \in \mathcal{X}$
\State Apply $k$-Means clustering on $\{\tilde{\mathbf{v}}_i\}$ to obtain cluster centroids $V \in \mathbb{R}^{d,k}$, where $V_k$ represents the $k$'th centroid
\State Create and clustered dataset $\tilde{\mathcal{X}}$ by replacing, for all $\mathbf{x}_i \in X \in \mathcal{X}$ with $\mathbf{v}_i = \min_k \|V_k -f_\theta(\mathbf{x}_i)\|$
\State Initialize subnetwork $h(\cdot)$ with vectors $\mathbf{c}_1, \dots, \mathbf{c}_K$
\State Let $h(\mathbf{v}) = \max_k(\mathbf{v}^\top \mathbf{c}_k)$ and $g(\tilde{X}) = \min_j\{ h(\mathbf{v}_j) : \mathbf{v}_j \in \tilde{X} \} + b$
\For{each bag $\tilde{X} = \{\mathbf{v}_j\}_{j=1}^{n_i}$ in $\tilde{\mathcal{X}}$}
    \State Compute binary cross entropy loss $\ell(y_i, g(\tilde{X}))$ and accumulate gradients/updates using optimizer. 
\EndFor
\end{algorithmic}
\end{algorithm}

By construction, our approach retains the fundamental assumptions of the MIL model: positive identification of a vector is needed to return a positive label, enforced by the max and min pooling steps. By using an auto-encoder, we enforce (approximately) the properties that the VSA needs for its theory to work, which handles the MIL constraints discussed at the beginning of this section. The use of multiple VSA vectors $\mathbf{c}_k$ that are learned via gradient descent (along with the bias term $b$) allows the model flexibility to determine the prediction. We will determine the number of concepts and codebook size via a hyperparameter tuning, detailed in the results section. 

We calculate the dot product between the VSA embeddings and concept vectors, then take the maximum value across all instances from each bag to identify the most representative and aligned instance from each bag for each concept. After we do this step for all concept vectors, we take the minimum across all concept vectors for each bag to capture the least satisfied concept. This enforces that concept vectors must be used, as otherwise a single dominant concept could prevent any learning from other vectors due to the lack of gradient when using max operations. For the same reason, a single bias term is used across all possible outputs to encourage a loose normalized score between them.

\section{Experimental Results} \label{sec:results}

In this article, we are concerned with the performance of valid MIL models that respect the MIL hypothesis. Most prior methods in deep learning, and even still published ones, are not valid MIL models but actually \textit{set} classification models~\cite{raff_reproducibility_2023}. For this reason, many methods that get higher test performance are not considered in our experiments, because they get to ``cheat'' and use verboten signals from the training data. This is a philosophical requirement for why we use the MIL model to begin with, the belief that a non-MIL constrained model may look better, but perform worse outside of a lab environment or over time, because it may be using known-to-be-invalid signals to obtain its apparently good performance. This is not to say such set approaches and non-MIL models may not be better under some circumstances, but we are interested specifically in obtaining the best possible performance under the conservative scenario of the MIL hypothesis. 

In line with the recommendation from \cite{raff_reproducibility_2023}, we use all test cases they provide, and our method passes all test cases, which can be found in Appendix Table \ref{table:cmp_algorithm_test}. 
We now detail the experiments to validate our approach. Two groups of datasets will be considered. First are \textit{traditional MIL benchmark datasets} like “Elephant”, “Protein”, “MUSK1”, “MUSK2”, “Ucsb breast cancer”, “Birds brown creeper”, “Web recommendation 1”, and “Corel dogs”. These are available as pre-vectorized representations $\mathbf{x}_i$ to apply a desired MIL algorithm to. 
Second, we consider four medical image datasets CAMELYON16 (Cancer Metastases in Lymph Nodes Challenge 2016), The Cancer Genome Atlas (TCGA) lung cancer dataset, RSNA Screening Mammography Breast Cancer (RSNA-SMBC), Brain Tumor MRI Dataset (BTMD), Prostate cANcer graDe Assessment (PANDA) 
 and RSNA Intracranial Hemorrhage Detection
(RSNA-ICH). We use Resnet18 as a ``backbone'' to vectorize the image, and take sub-regions (or ``patches'') of the image to be the instances in a bag. While using a more advanced architecture would likely yield better results, we use ResNet18 to directly compare with prior work that used the same architecture. Changing the backbone before comparison would lead to an unfair advantage to our approach and has caused replication issues in the past that we aim to avoid~\cite{Musgrave2020}. 

We perform the training with one NVIDIA GeForce RTX $2080$ Ti. For training our autoencoder and the $h()$ and $g()$ models, we use a batch size of 16, learning rate = 0.1, with the AdamW optimizer, and 50 epochs of training. We use the Optuna library~\cite{Akiba:2019:ONH:3292500.3330701} to perform hyperparameter tuning for the number of ReZero blocks, the number of layers in each block, the number of clusters in our autoencoder, and the number of concept vectors and the learning rate for $h()$. We varied them to find the best combination of these variables, yielding the best results for each dataset using a validation set split from the training data or the dataset's provided validation set.

Since the focus of this work is on deep learning methods for MIL classification, and it has been recently identified that most papers on this topic do not actually respect the MIL hypothesis, we will have limited baselines to compare against. Our primary deep learning approach is the CausalMIL baseline that has shown improved results on traditional MIL tasks and uses a Bayesian approach to guarantee the MIL assumptions~\cite{zhangMultiInstanceCausalRepresentation2022}, was validated in testing by \cite{raff_reproducibility_2023}, and has readily available code. For the medical imaging datasets, we will also compare to the proposed method of the DSMIL paper~\cite{DSMIL} --- with the cautionary note that in the article~\cite{DSMIL} , \textit{an approach known to be invalid because it will learn MIL violating predictions is used}, given these prior works an unfair advantage in the task. Finally, we will test the miSVM~\cite{NIPS2002_2232} classifier that is of historical note, respects the MIL assumptions, and provides a baseline to apply on top of our auto-encoded representation as a method of confirming that we provide an additional improvement using VSAs. 

\subsection{Traditional Results}

We first look at the hyperparameter tuning results for each dataset, which are found in Table~\ref{table:parameters_optuna_MIL_benchmark}. Given that many of these older datasets are small, it is unsurprising that a small number of layers and blocks are considered. However, the number of clusters and concept vectors is particularly informative, and we remind the reader that all tasks are binary prediction problems. By having multiple latent concepts represented via the $\mathbf{c}_i$ VSAs, we can interpret the number of cluster/concept vectors as loose proximal signals about the number of underlying types of content in the dataset. 

\begin{table}[!h]
\caption{Parameters that give the best AUROC score for the benchmark MIL datasets using Optuna}
\label{table:parameters_optuna_MIL_benchmark}
\centering
\adjustbox{max width=\columnwidth}{%
\begin{tabular}{lcccccccc}
\toprule
Dataset & Block & Layer & Cluster & Concept \\

 & number & number & number & vector \\
\midrule
Elephant & 1 & 2 & 4 & 18 \\
Protein & 2 & 1 & 9 & 18 \\
MUSK1 & 1 & 3 & 3 & 1 \\
MUSK2 & 1 & 2 & 10 & 1 \\
UCSB Breast Cancer & 1 & 3 & 3 & 12  \\
Birds brown creeper & 1 & 4 & 9 & 5 \\
Web Recommendation1 & 2 & 1 & 10 & 17 \\
Corel dogs & 1 & 4 & 10 & 20 \\

\bottomrule
\end{tabular}
}
\end{table}

\begin{figure}[!h]
    \centering
    \includegraphics[width=0.98\columnwidth]{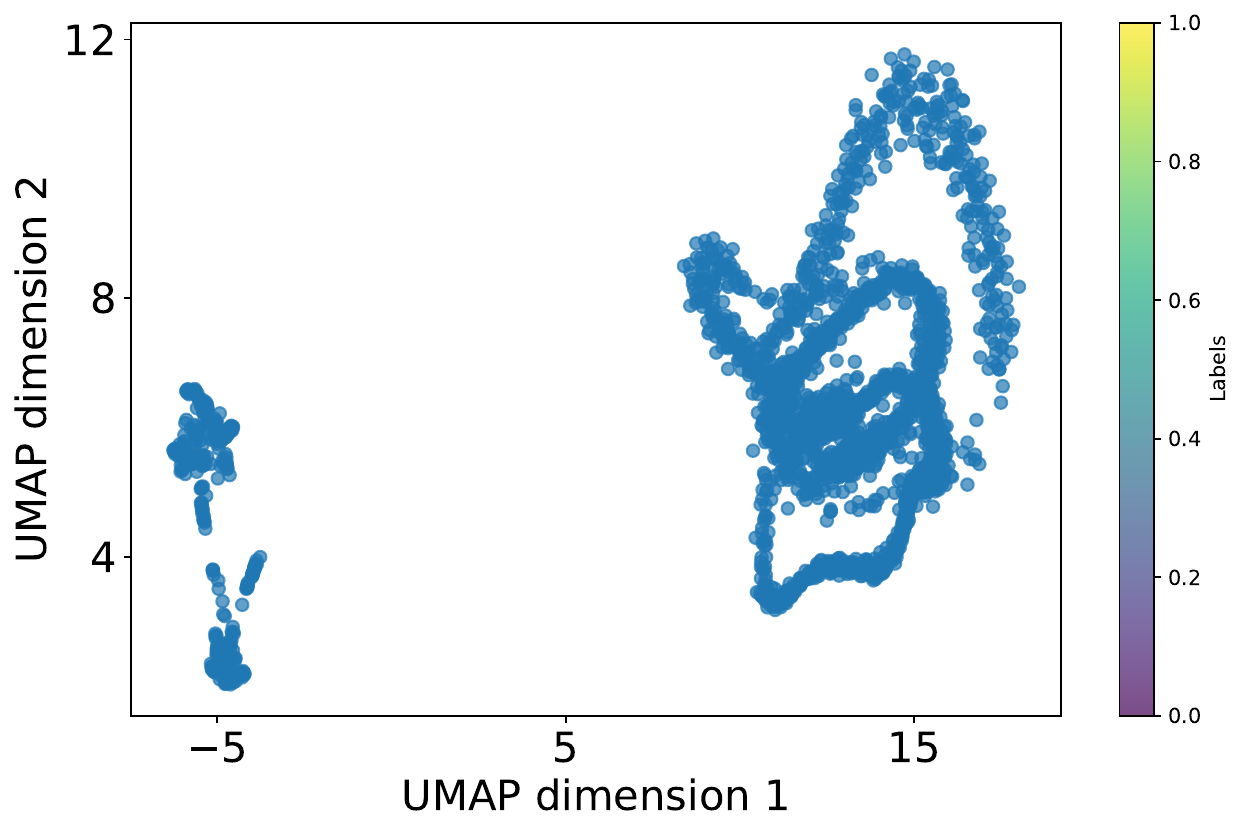}
    \caption{UMAP projection of VSA embeddings into 2D space from the training dataset of MUSK1.}
    \label{fig:umap_projection}
\end{figure}

One may wonder if the number of clusters is reasonable and surprisingly small given the task at hand. We can explore this factor by using the UMAP algorithm~\cite{McInnes2018,Nolet2020} to project the learned encoding $f_\theta(\cdot)$ into a two-dimensional space. This is done on the MUSK1 dataset which determined only 3 clusters in Figure~\ref{fig:umap_projection}. In the UMAP plot we can see the evidence for approximately 3 clusters, and shows that the determination is reasonable. 

It is also noteworthy that the rightmost cluster of the plot has far greater variance/asperity than the tight groups on the left. This highlights why our k-means clustering step is useful: it helps to avoid the large variance factors that are potentially meaningless in the semantics of the content, but would meaningfully alter the response of the VSA's similarity calculation.

Looking at the results in accuracy and AUC in Table~\ref{table:cmp_MIL_benchmark}, we see that our VSA-MIL significantly outperforms the CausalMIL option and the miSVM. This demonstrates the high utility of having multiple latent concepts that can map to the positive label, whereas CausalMIL has a single weight vector applied after pooling and thus is a more limited hypothesis space. 

\begin{table}[!h]
\caption{Comparison of performance between different methods for the benchmark MIL datasets}
\label{table:cmp_MIL_benchmark}
\adjustbox{max width=\columnwidth}{%
\begin{tabular}{lcccccccc}
\toprule
Dataset & \multicolumn{2}{c}{misvm} & \multicolumn{2}{c}{CausalMIL} & \multicolumn{2}{c}{VSA-MIL} \\
 & Acc. & AUROC & Acc. & AUROC & Acc. & AUROC \\
\midrule
Elephant & 0.825 & 0.878 & 0.810 & 0.881 & \textbf{0.900} & \textbf{0.954} \\ 
Protein & 0.871 & 0.773 & 0.872 & 0.741 & \textbf{0.948} & \textbf{0.988} \\
MUSK1 & 0.736 & 0.727 & 0.737 & 0.807 & \textbf{0.894} & \textbf{0.965} \\
MUSK2 & 0.666 & 0.981 & 0.800 & 0.872 & \textbf{0.952} & \textbf{1.000} \\
Ucsb Breast Cancer & 0.666 & 0.657 & 0.667 & 0.743 & \textbf{0.833} & \textbf{0.942} \\ 
Birds Brown Creeper & 0.918 & \textbf{0.987} & 0.918 & 0.964 & \textbf{0.945} & 0.984 \\
Web Rec. & 0.800 & 0.733 & 0.866 & 0.927 & \textbf{0.933} & \textbf{0.964} \\
Corel Dogs & \textbf{0.930} &  0.850 & \textbf{0.930} & 0.836 & \textbf{0.930} & \textbf{0.941} \\

\bottomrule
\end{tabular}
}
\end{table}

\begin{figure}[!h]
    \centering
    \includegraphics[width=0.98\columnwidth]{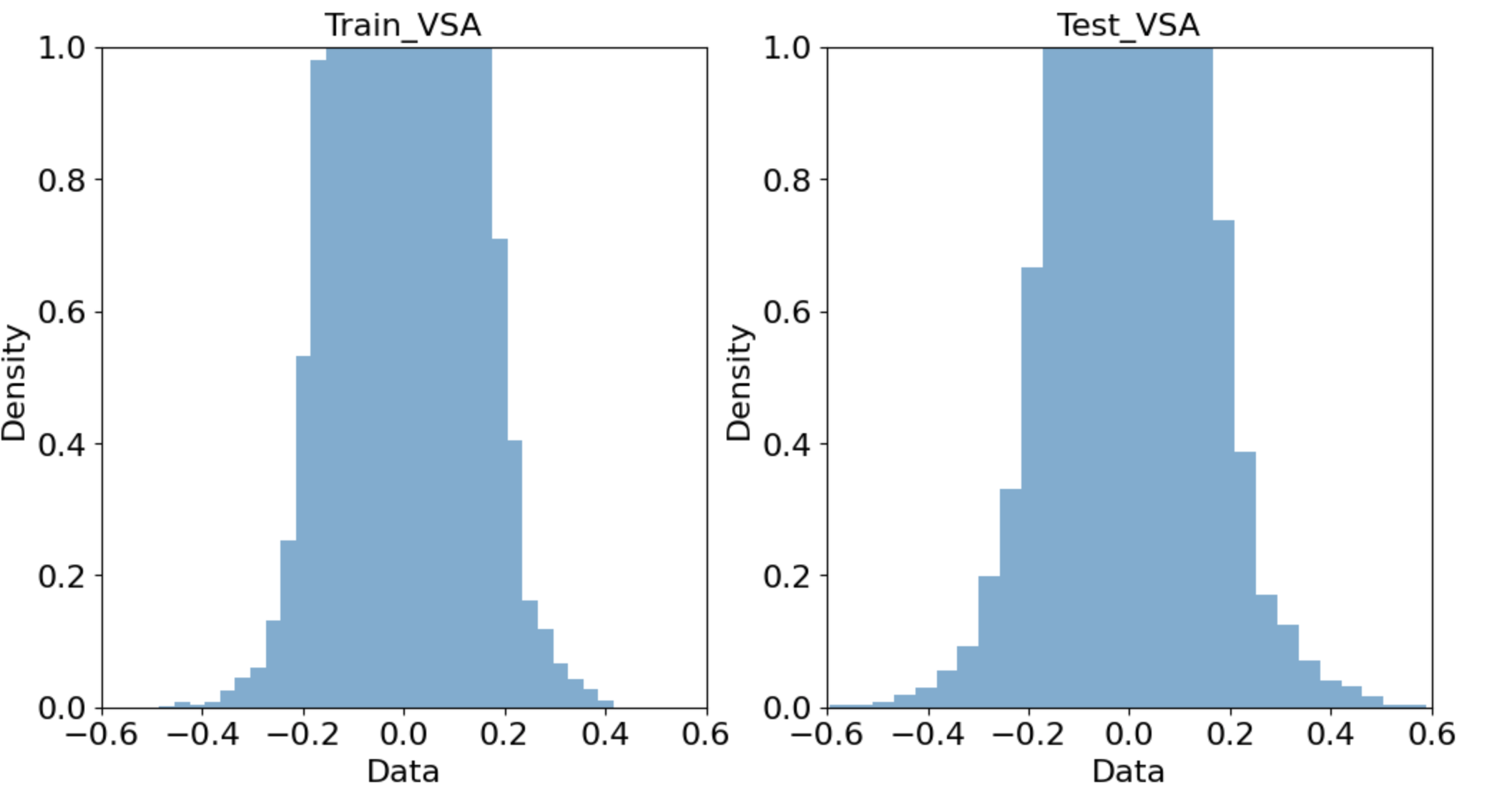}
    \caption{Histogram density distribution of VSA from the training dataset of MUSK1.}
    \label{fig:vsa_distribution}
\end{figure}

The success of our approach in producing these classifications is dependent on making the input instances $\mathbf{x}$ closely match the target distribution properties of the VSA. We provide an example of this in Figure~\ref{fig:vsa_distribution} showing the histogram of the training and testing data instance dimensions. As can be seen visually, despite not originally existing in a normalizer form, our auto-encoder $f_\theta(\cdot)$ learns to successfully transform the distribution to maintain the desired properties from Section\textit{~\nameref{sec:methodology}}. 

\subsection{Medical Results}

We now turn to evaluating our approach on the medical imaging tasks, which have become a highly desirable application space for MIL due to the correspondence in pathology (e.g., cancer requires cancer cells). We similarly do not expect a large number of non-linearities to be necessary, as the ResNet18 backbone being used to convert patches of the image into ``bags'' is already performing a significant non-linear operation. 

The more interesting result is to look at the predictive performance of each method, where miSVM will have a significant relative advantage due to the ResNet18 backbone performing a large amount of non-linear operations and being used across all approaches. The results are found in Table~\ref{table:cmp_medical_data}, where the last row has the state-of-the-art (SOTA) results that are \textit{invalid in respecting the MIL hypothesis}. In this perspective, we expected the paper's method to have the ``best'' results because it is not actually a valid MIL model, but in three metrics over two datasets, we see that VSA-MIL actually outperformed the invalid approach.

\begin{table}[!h]
\caption{Camparison of performance between different methods for medical image datasets}
\label{table:cmp_medical_data}
\adjustbox{max width=\columnwidth}{%
\begin{tabular}{@{}cccccc@{}}
\toprule
Dataset                         & Metric   & misvm     & CausalMIL       & VSA-MIL           & DSMIL (Invalid approach)  \\  \midrule
\multirow{2}{*}{CAMELYON16}     & Accuracy & 0.698     & 0.690           & \textbf{0.813}    & 0.868          \\
                                & AUROC    & 0.638     & 0.740           & \textbf{0.856}    & 0.894          \\ \midrule
\multirow{2}{*}{TCGA}           & Accuracy & 0.875     & 0.876           & \textbf{0.958}    & 0.957          \\
                                & AUROC    & 0.949     & 0.954           & \textbf{0.992}    & 0.981          \\ \midrule
\multirow{2}{*}{RSNA-SMBC}      & Accuracy & 0.638     & 0.960           & \textbf{0.970}    & 0.950          \\
                                & AUROC    & 0.641     & 0.701           & \textbf{0.820}    & 0.860          \\ \midrule
\multirow{2}{*}{BTMD}           & Accuracy & 0.741     & \textbf{0.906}  & \textbf{0.906}    & 0.997          \\
                                & AUROC    & 0.615     & 0.722           & \textbf{0.870}    & 0.984          \\ \midrule
\multirow{2}{*}{PANDA}          & Accuracy & 0.735     & 0.749           & \textbf{0.853}    & 0.818          \\
                                & AUROC    & 0.798     & 0.857           & \textbf{0.934}    & 0.872          \\ \midrule
\multirow{2}{*}{RSNA-ICH}       & Accuracy & 0.632     & 0.756           & \textbf{0.816}    & 0.824          \\
                                & AUROC    & 0.880     & 0.907           & \textbf{0.922}    & 0.960          \\
                                
                                \bottomrule
\end{tabular}
}
\end{table}

Compared to miSVM and CausalMIL, which use the same ResNet18 representation as VSA-MIL, we see that VSA-MIL obtains significantly improved results in all cases, by 10-17\% in Accuracy and AUROC in most cases. This shows that VSA-MIL is a highly effective and competitive approach to using the MIL modeling approach, and the value of approaching the MIL task from new perspectives like VSAs.

\begin{figure}[!h]
    \centering
    \includegraphics[width=0.8\columnwidth]{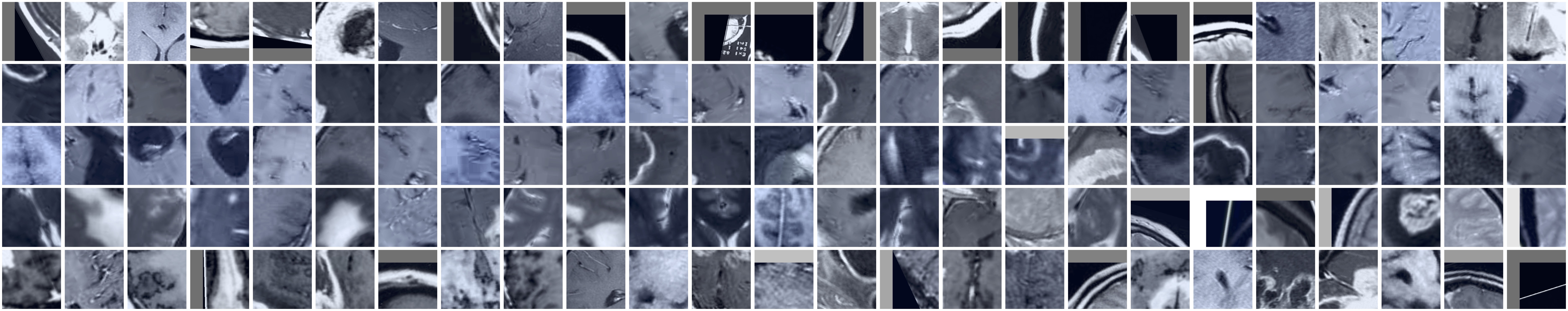}
    \caption{Extracting image patches for representing concept vectors with 25 images from training set of BTMD dataset.}
    \label{fig:image_patch_brain_tumor}
\end{figure}

\begin{figure}[!h]
    \centering
    \includegraphics[width=0.8\columnwidth]{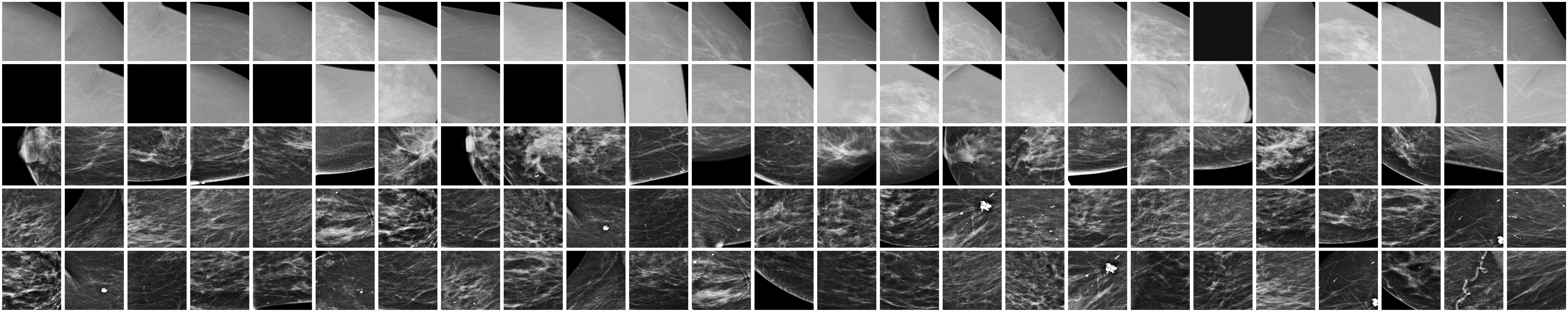}
    \caption{Extracting image patches for representing concept vectors with 25 images from the training set of the RSNA-SMBC dataset.}
    \label{fig:image_patch_rsna_breast}
\end{figure}

A further benefit of our approach is that the concept vectors $\mathbf{c}_k$ can be used to gain an interpretable component about what the model is learning to look for. We can do this by showing example patches that correspond to the $k$'th concept vector activating and having the maximal response, i.e., making the classification determination. We provide an example for the first five concept classes on the BTMD dataset in Figure~\ref{fig:image_patch_brain_tumor} and the RSNA-SMBC dataset in Figure~\ref{fig:image_patch_rsna_breast}. In both cases, the row corresponds to a concept, and 25 columns show 25 winning exemplars. 

The BTMD dataset has considerably more visual variation in the images, making them more challenging to inspect as non-medical experts, but visual themes are generally discernible across each row in Figure~\ref{fig:image_patch_brain_tumor}. We note that there is no mechanism to force one concept vector to suppress other concept vectors for similar activations, and so we see some cross-pollination between the concept vectors. 
The RSNA dataset has less visual asperity, and visual patterns are more clearly observed in the rows of Figure~\ref{fig:image_patch_rsna_breast}.

\begin{figure}[!h]
    \centering
    \includegraphics[width=\columnwidth]
    {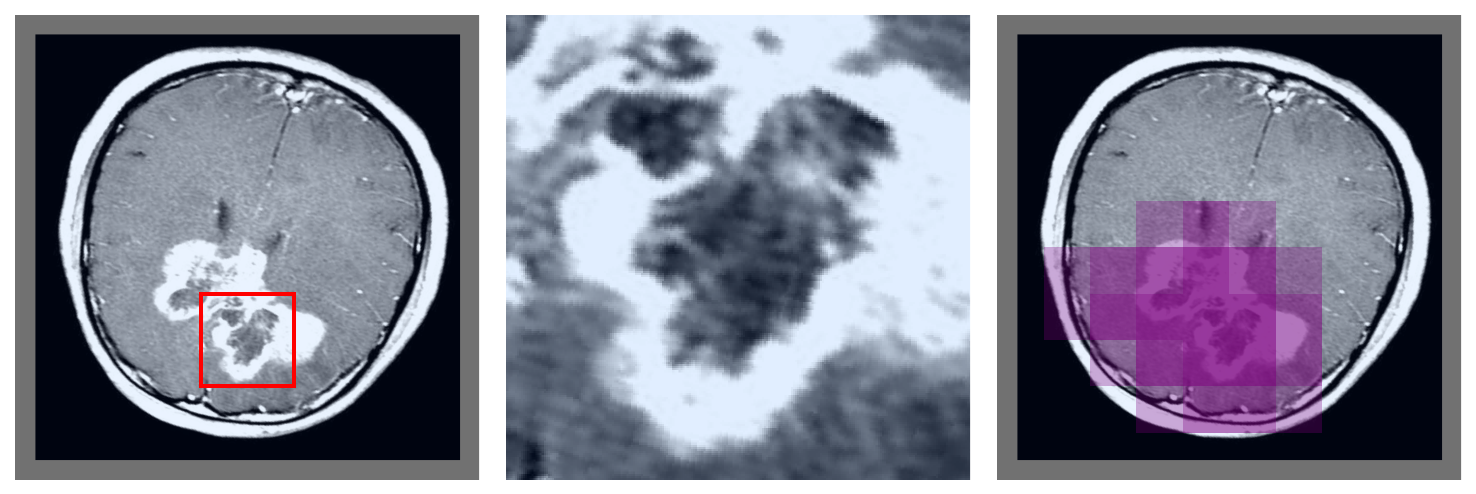}
    \includegraphics[width=\columnwidth]
    {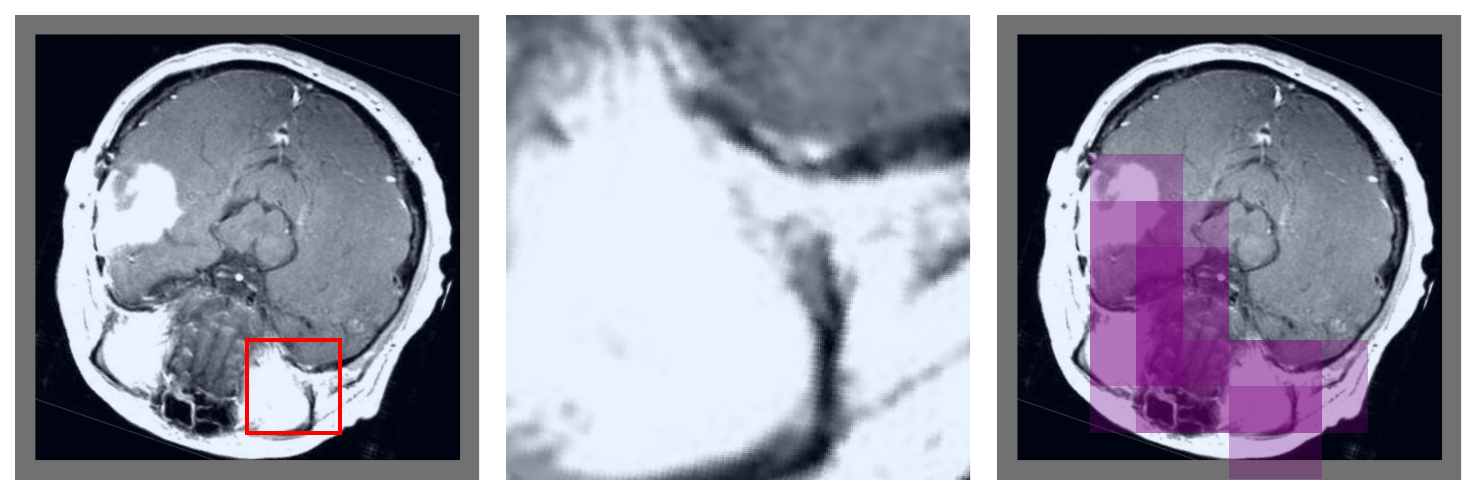}
    \includegraphics[width=\columnwidth]
    {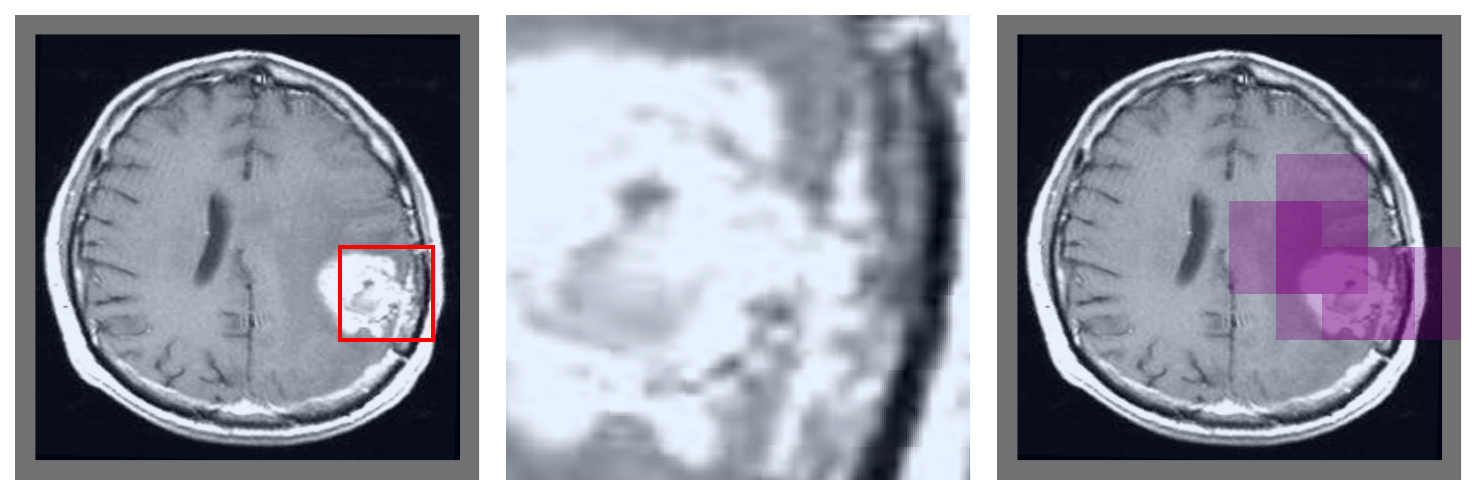}

    \caption{
    From the BTMD, we see on the left side the most positive region of the image detected by VSA-MIL (which is zoomed in on the center image), demonstrating our method's ability to identify the positive items. On the right, we show all positive instances in the ``bag'' as detected by VSA-MIL, which covers all three tumors in the brain. 
    }
    \label{fig:image_patch_brain_tumor_important_patches_overlap1_single}
\end{figure}

Finally, we can also visualize the results of our MIL model in the same manner as prior approaches, highlighting either the most positive, or all positive, regions to allow interpretable inspection and verification. This is demonstrated in Figure \ref{fig:image_patch_brain_tumor_important_patches_overlap1_single} using the BTMD dataset, in which tumors are recognizable to non-medical experts. The appendix has additional random samples from the other dataset. 

\subsubsection{Ablating Backbone and Overlap}

In our experiments we used the ResNet18 architecture as a ``backbone'' to embed patches into a feature-vector, as it was the same architecture used in prior papers. This showed that, on equal representation, our method achieved better performance in an apples-to-apples comparison with prior papers that had an unfair advantage in being larger than the MIL hypothesis space.

\begin{table}[!h]
\caption{Ablating the impact of changes in the percentage of overlap between patches in bag construction, and the choice of embedding backbone architecture, across RSNA-SMBC (first rows) and BTMD (second rows), PANDA (third rows), and RSNA-ICH (fourth rows) datasets. Our results show further improvement is possible by using more modern backbones and increasing the overlap.} \label{tbl:backbone_overlap_ablation}
\adjustbox{max width=\columnwidth}{%
\begin{tabular}{@{}lcccccc@{}}
\toprule
\multicolumn{1}{c}{}          & \multicolumn{2}{c}{0\% Overlap} & \multicolumn{2}{c}{50\% Overlap} & \multicolumn{2}{c}{75\% Overlap} \\ \midrule
\multicolumn{1}{c}{Back-Bone} & Acc.           & AUROC          & Acc.           & AUROC           & Acc.           & AUROC           \\ \midrule
ResNet18                      & 0.970          & 0.820          & 0.980          & 0.829           & 0.980          & 0.840            \\
EfficientNet B$0$             & 0.975          & 0.869          & 0.980          & 0.883           & 0.980          & 0.898            \\
DINOv$2$ ViT                  & 0.975          & 0.846          & 0.980          & 0.871           & 0.980          & 0.884            \\ \midrule
ResNet18                      & 0.906          & 0.870          & 0.881          & 0.866           & 0.909          & 0.882            \\
EfficientNet B$0$             & 0.909          & 0.887          & 0.916          & 0.921           & 0.934          & 0.947            \\
DINOv$2$ ViT                  & 0.912          & 0.889          & 0.919          & 0.924           & 0.931          & 0.939            \\ 
\midrule
ResNet18                      & 0.853          & 0.934          & 0.861          & 0.945           & 0.875          & 0.961            \\
EfficientNet B$0$             & 0.857          & 0.938          & 0.862          & 0.947           & 0.881          & 0.963            \\
DINOv$2$ ViT                  & 0.857          & 0.938          & 0.865          & 0.949           & 0.880          & 0.961            \\ \midrule
ResNet18                      & 0.816           & 0.922         & 0.837          & 0.937           & 0.842          & 0.940            \\
EfficientNet B$0$             & 0.815           & 0.922         & 0.839          & 0.938           & 0.841          & 0.940            \\
DINOv$2$ ViT                  & 0.819           & 0.925         & 0.847          & 0.941           & 0.847          & 0.941            \\ \bottomrule
\end{tabular}
}
\end{table}

To further emphasize the quality of our approach, we ablate the choice of back-bone, and the amount of overlap between patches. We test EfficientNet and DINOv2 as two choices that balance between light-weight/fast versus larger and slower, respectively. Similarly, we test 0\%, 50\%, and 75\% overlap between patches for speed vs coverage tradeoffs. The results are found in Table \ref{tbl:backbone_overlap_ablation}, showing we can further improve results with better architectures, and still perform well even with low overlap.

\section{Conclusion} \label{sec:conclusion}

In this work, we have developed a new approach to Multiple Instance Learning (MIL) using Vector Symbolic Architectures (VSAs). VSAs allow us to enforce the MIL assumptions in our design, avoiding the pitfall of many prior studies in designing neural networks with no means of enforcing the \textit{positive if and only if} constraint of the MIL hypothesis space. By designing an approach to learn a translation from feature inputs to a VSA-compatible representation that maintains the sufficient conditions of the VSA, we design an approach that yields significant gains in accuracy compared to prior valid MIL options. 

\bibliography{refs,referencesZoteroRaff}

\clearpage
\onecolumn
\section{Appendix}

As reported by \cite{raff_reproducibility_2023}, many MIL algorithms fail to work properly and this can be tested with ``poisoned'' datasets that have two solutions: one which respects the MIL hypothesis, and a second which violates the MIL and has an easier signal to learn. A broken implementation, or poorly designed algorithm, will fail the tests. We refer the reader to \cite{raff_reproducibility_2023} for the details, but as shown in Table \ref{table:cmp_algorithm_test} each test is passed by obtaining a $\geq 0.5\%$ AUROC on the test set. In fact, our method is the only one to receive a non-trivial positive accuracy in all three tests. 

\begin{table}[!h]
\caption{Camparison of performance of VSA-MIL approach with Algorithm 1, 2, and 3 from \cite{raff_reproducibility_2023} paper}
\label{table:cmp_algorithm_test}
\centering
\adjustbox{max width=\columnwidth}{%
\begin{tabular}{cccccc}
\toprule
Algorithm & Train accuracy & Train AUROC & Test accuracy & Test AUROC \\
\midrule

1 & 0.952 & 0.995 & 0.897 & 0.987   \\
2 & 0.918 & 0.950 & 0.824 & 0.873     \\
3 & 0.943 & 0.978 & 0.877 & 0.915   \\

\bottomrule
\end{tabular}
}
\end{table}

\begin{figure}[!h]
    \centering
    \includegraphics[width=0.8\columnwidth]{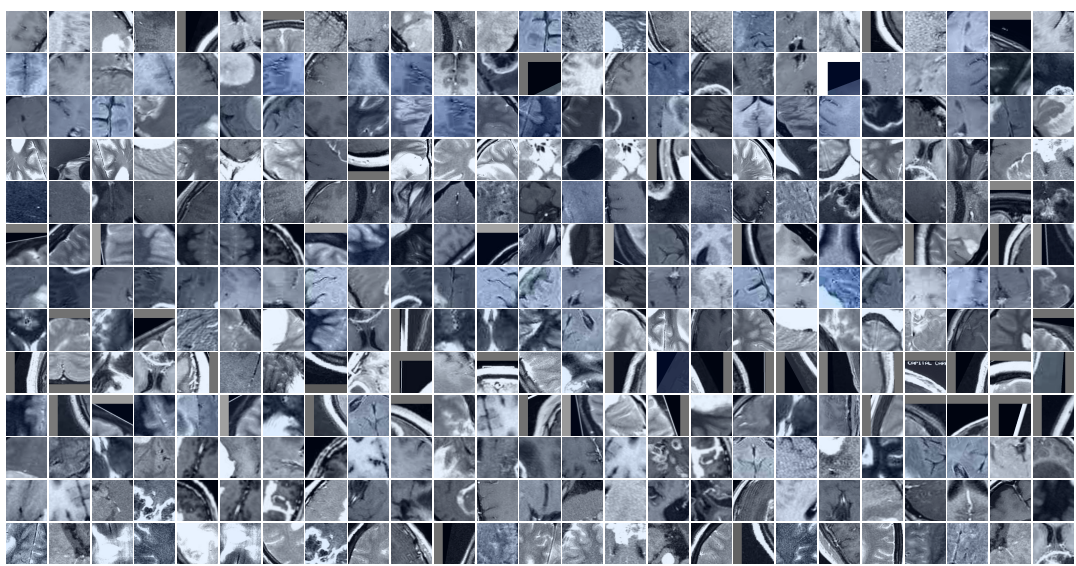}
    \caption{Extracting image patches for representing all concept vectors with 25 images from training set of BTMD dataset.}
    \label{fig:image_patch_brain_tumor_all}
\end{figure}

\begin{figure}[!h]
    \centering
    \includegraphics[width=0.8\columnwidth]{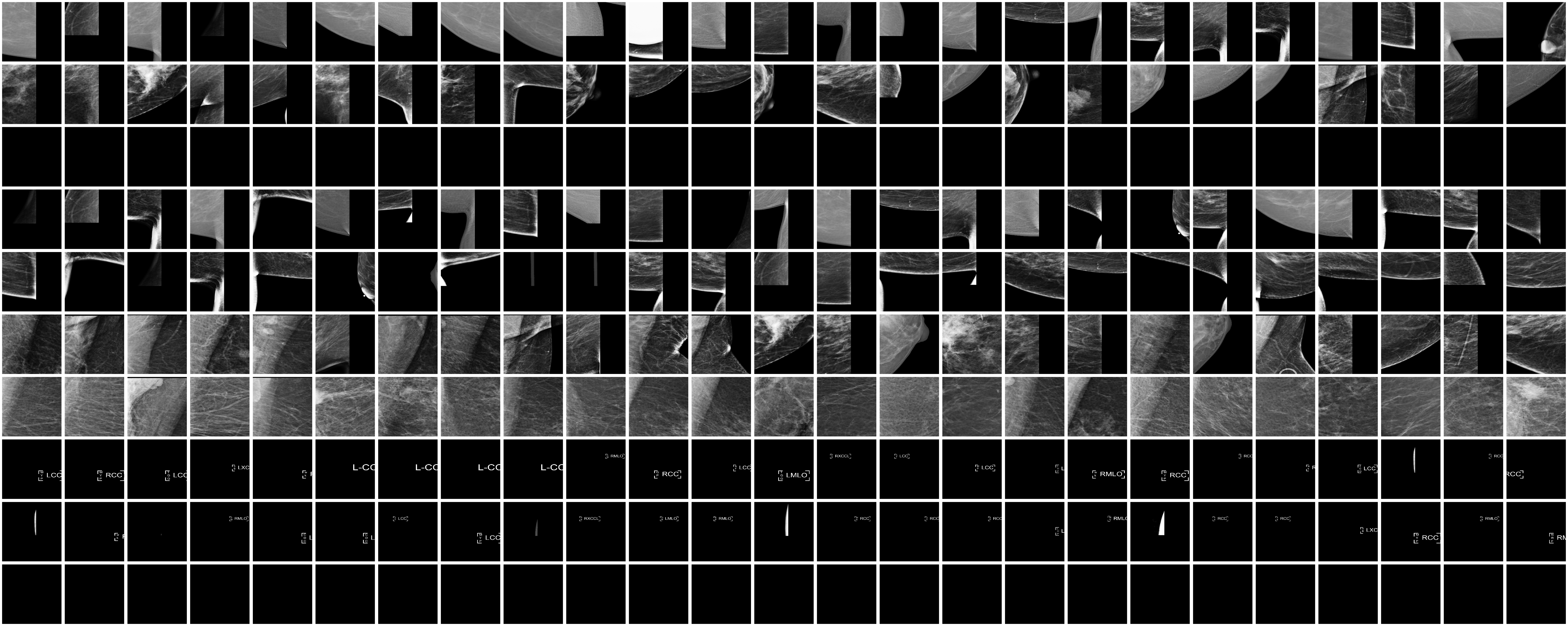}
    \caption{Extracting image patches for representing all concept vectors with 25 images from training set of RSNA-SMBC dataset.}
    \label{fig:image_patch_rsna_breast_all}
\end{figure}

In Figures \ref{fig:image_patch_brain_tumor_all} and \ref{fig:image_patch_rsna_breast_all}, we display more representative patches selects that correspond highly to a given concept vector (one concept per row), highlighting our our method automatically learns sub-regions of the larger image space. 

\begin{figure}[!h]
    \centering
    \begin{subfigure}[b]{0.45\textwidth}
        \centering
        \includegraphics[width=\linewidth]{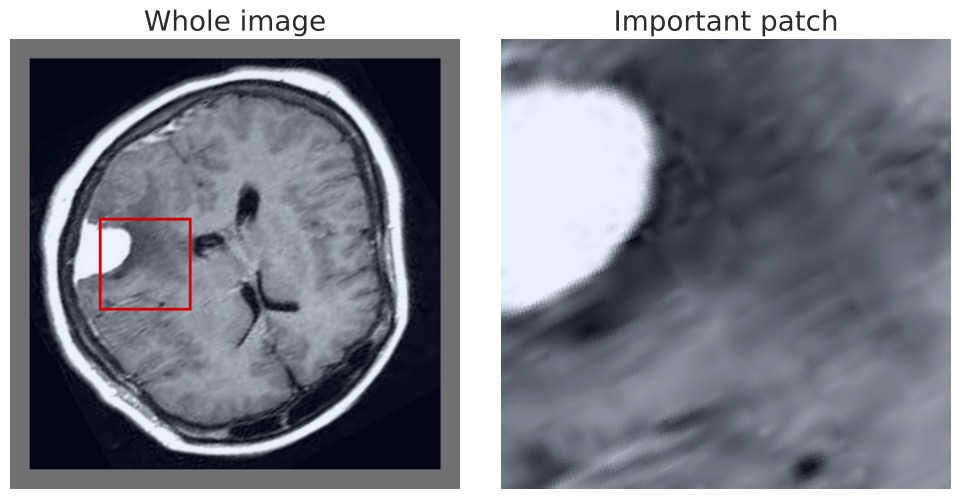}
        \caption{Image patch 1}
    \end{subfigure}
    \hfill
    \begin{subfigure}[b]{0.45\textwidth}
        \centering
        \includegraphics[width=\linewidth]{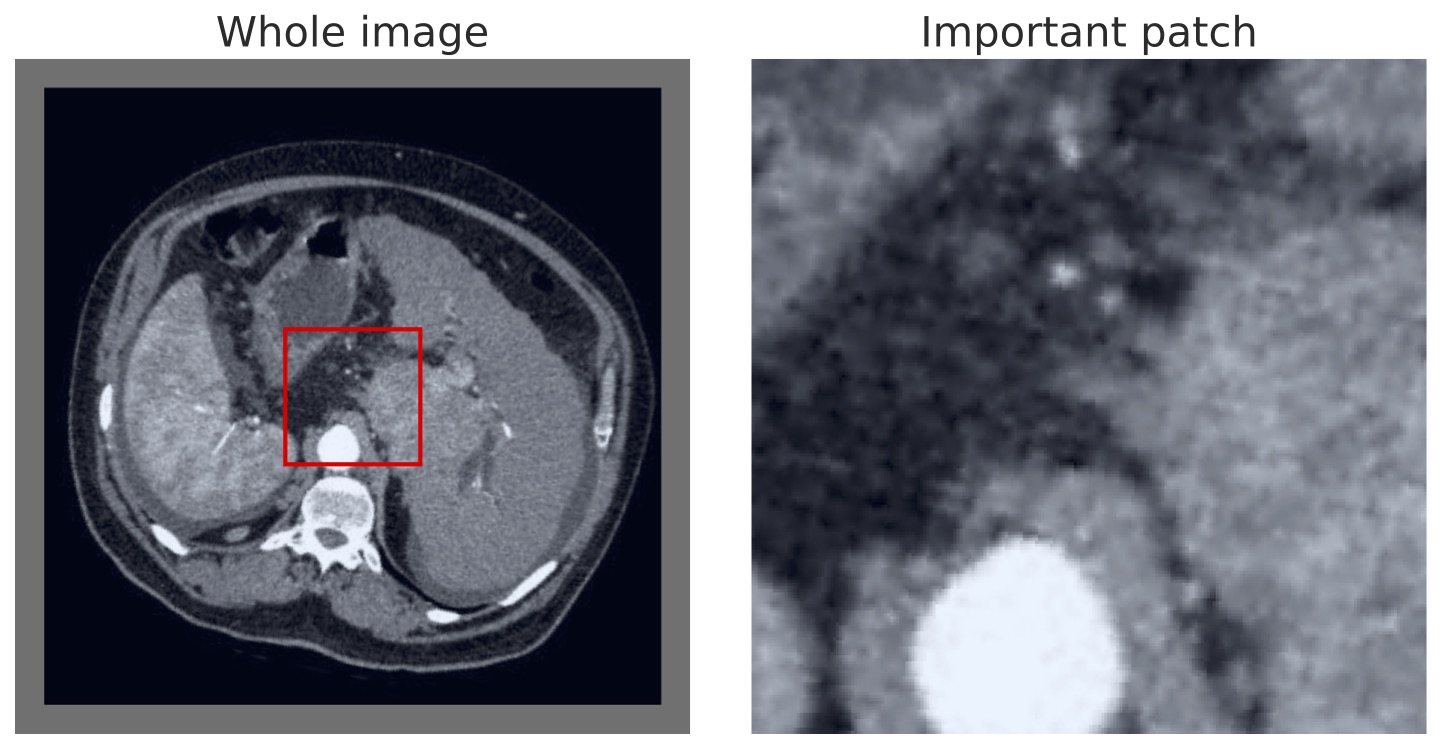}
        \caption{Image patch 2}
    \end{subfigure}

    \vspace{1em} %

    \begin{subfigure}[b]{0.45\textwidth}
        \centering
        \includegraphics[width=\linewidth]{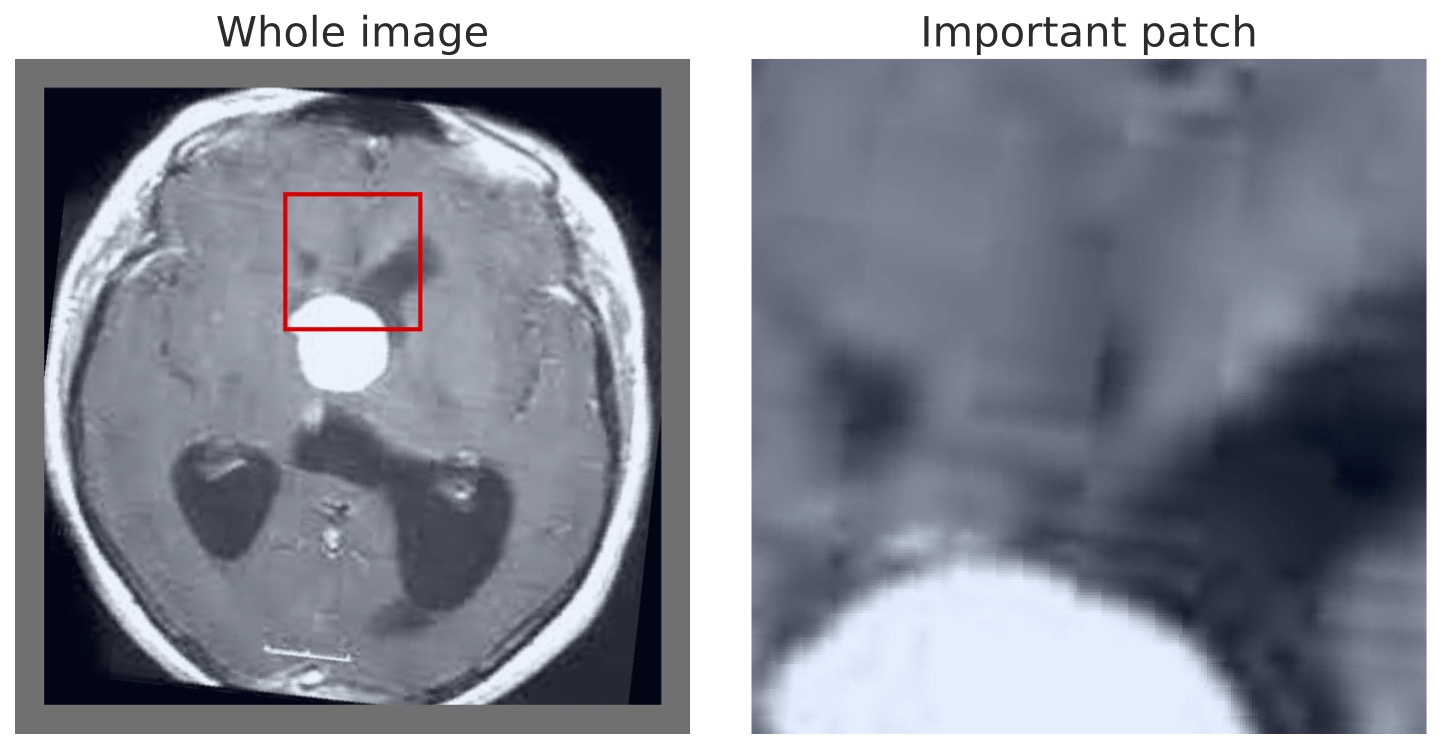}
        \caption{Image Patch 3}
    \end{subfigure}
    \hfill
    \begin{subfigure}[b]{0.45\textwidth}
        \centering
        \includegraphics[width=\linewidth]{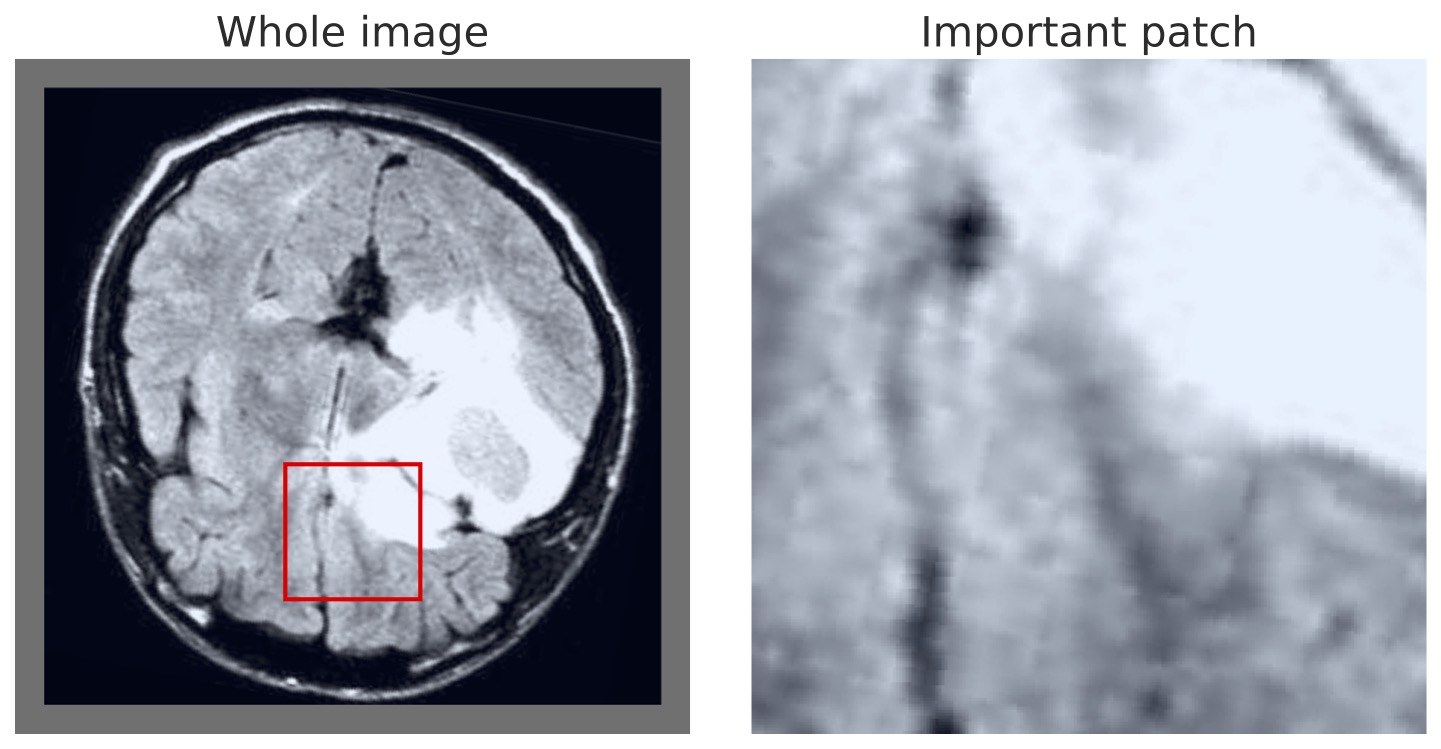}
        \caption{Image Patch 4}
    \end{subfigure}

    \vspace{1em}

    \begin{subfigure}[b]{0.45\textwidth}
        \centering
        \includegraphics[width=\linewidth]{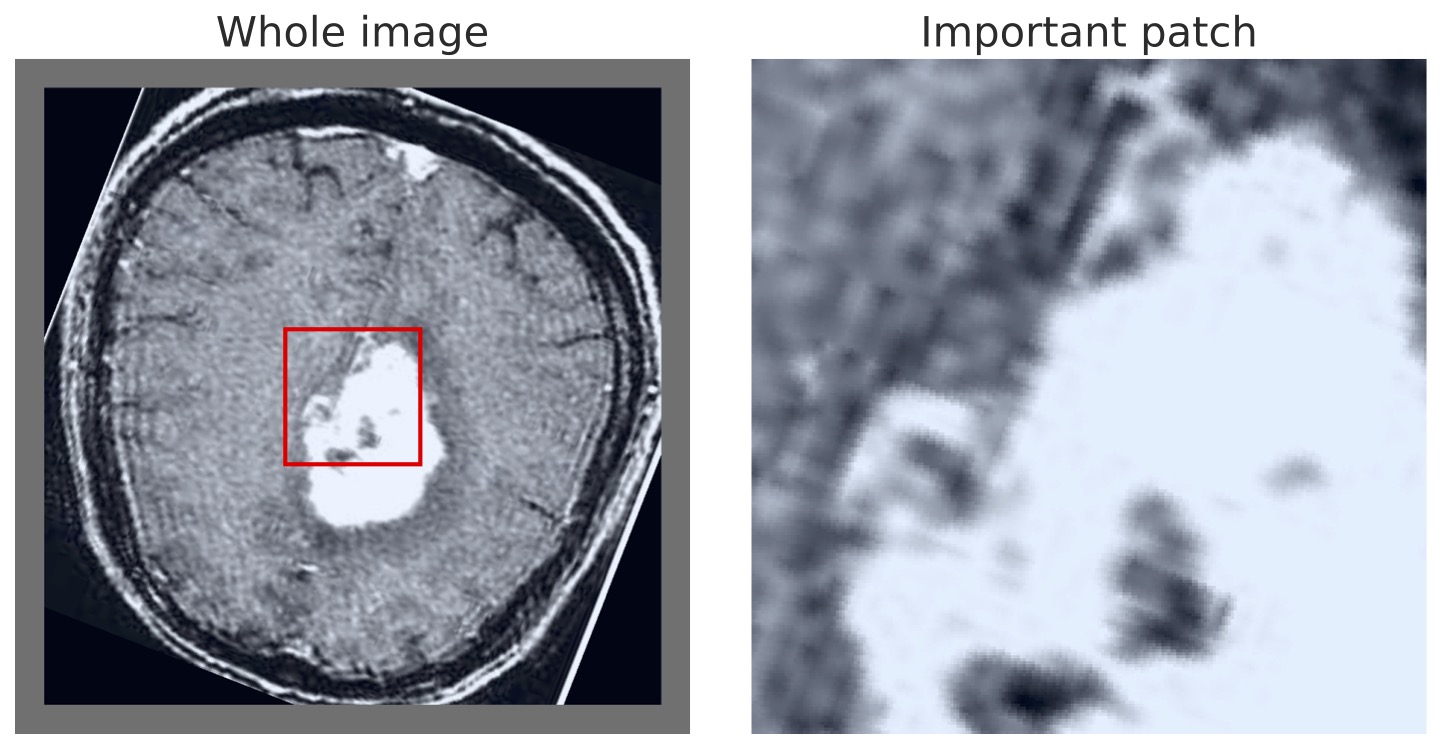}
        \caption{Image patch 5}
    \end{subfigure}
    \hfill
    \begin{subfigure}[b]{0.45\textwidth}
        \centering
        \includegraphics[width=\linewidth]{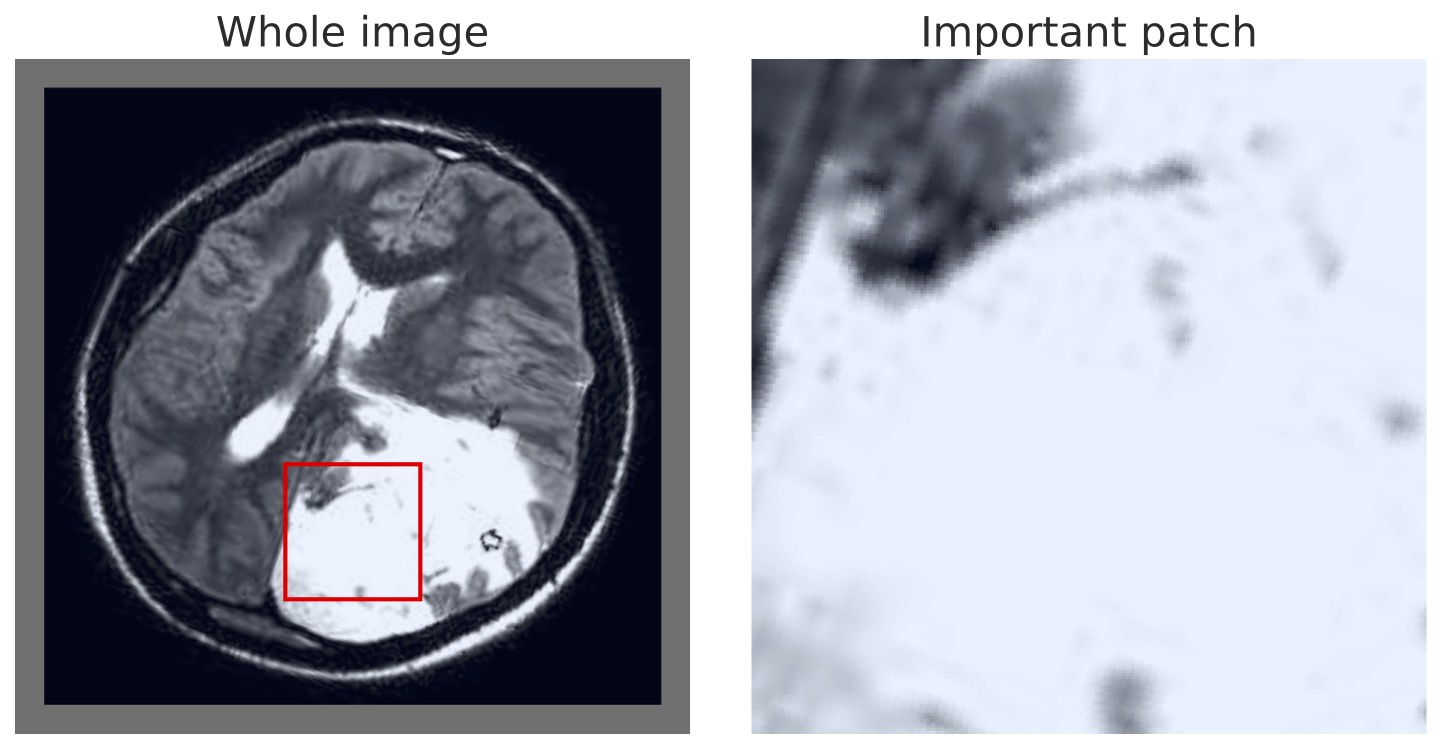}
        \caption{Image patch 6}
    \end{subfigure}

    \caption{Indicating image patches influential for detecting tumor cell of a whole slide image from training set of BTMD dataset.}
    \label{fig:image_patch_brain_tumor_important_patch}
\end{figure}

In Figure \ref{fig:image_patch_brain_tumor_important_patch} we show multiple BTMD images randomly selected and the maximal positive detection region from each image. In each case it is clear that the model successfully focuses on the malignant portions. 

\begin{figure}[!h]
    \centering
    \begin{subfigure}[b]{0.45\textwidth}
        \centering
        \includegraphics[width=\linewidth]{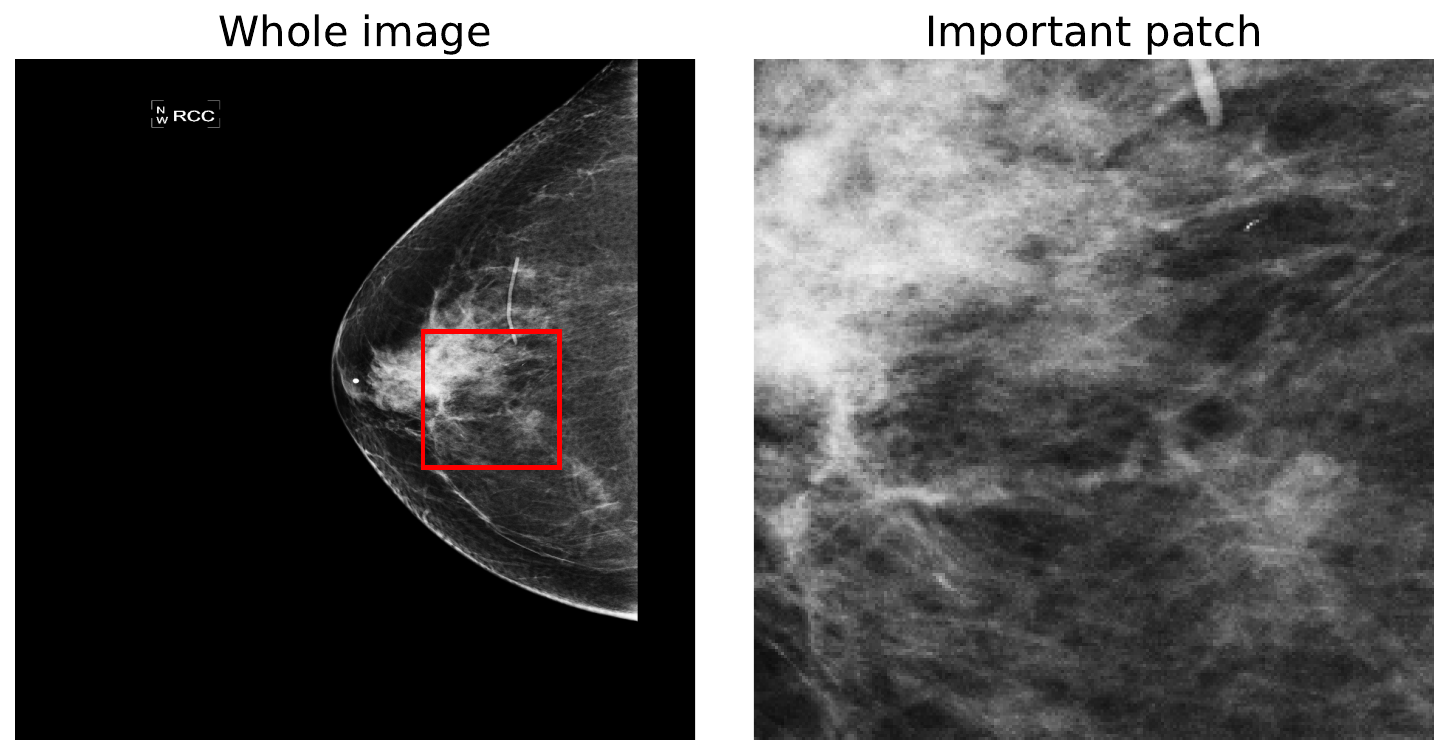}
        \caption{Image patch 1}
    \end{subfigure}
    \hfill
    \begin{subfigure}[b]{0.45\textwidth}
        \centering
        \includegraphics[width=\linewidth]{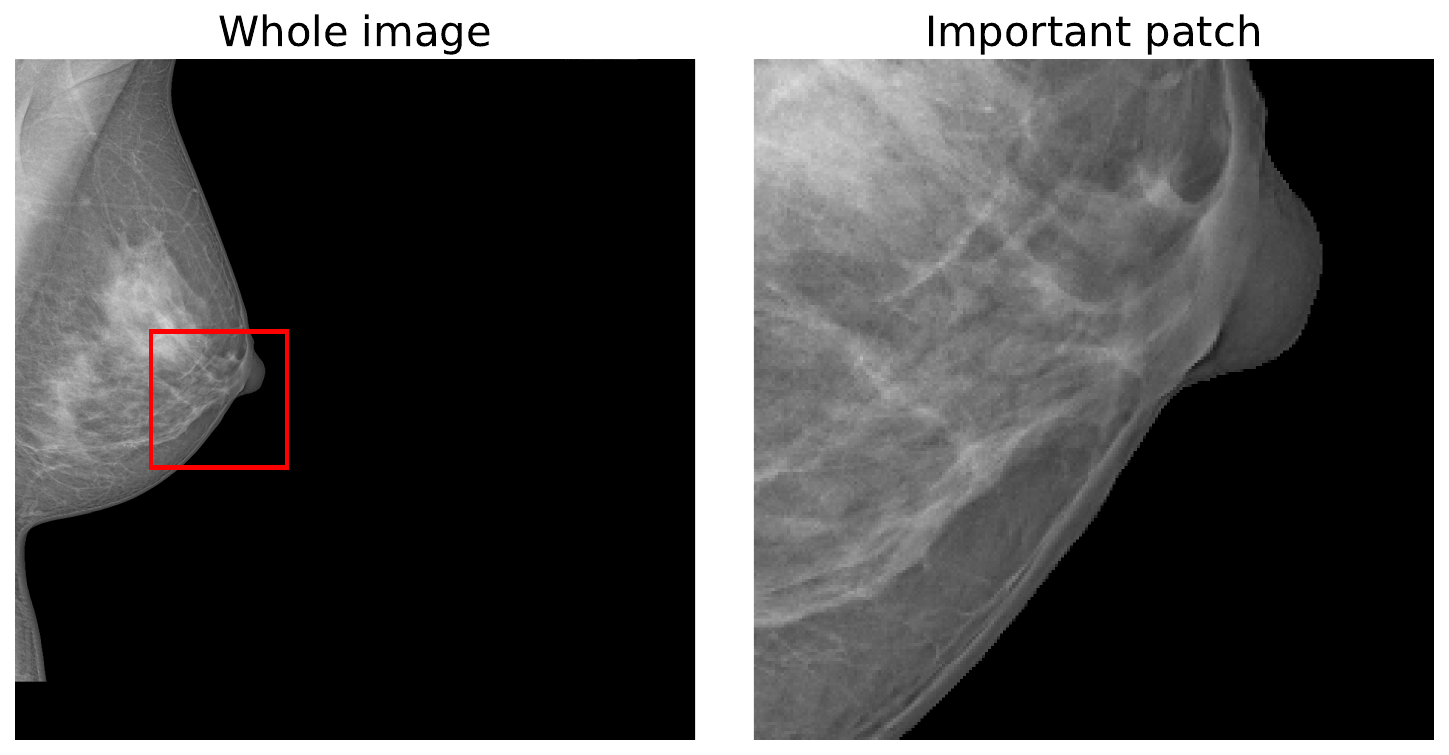}
        \caption{Image patch 2}
    \end{subfigure}

    \vspace{1em} %

    \begin{subfigure}[b]{0.45\textwidth}
        \centering
        \includegraphics[width=\linewidth]{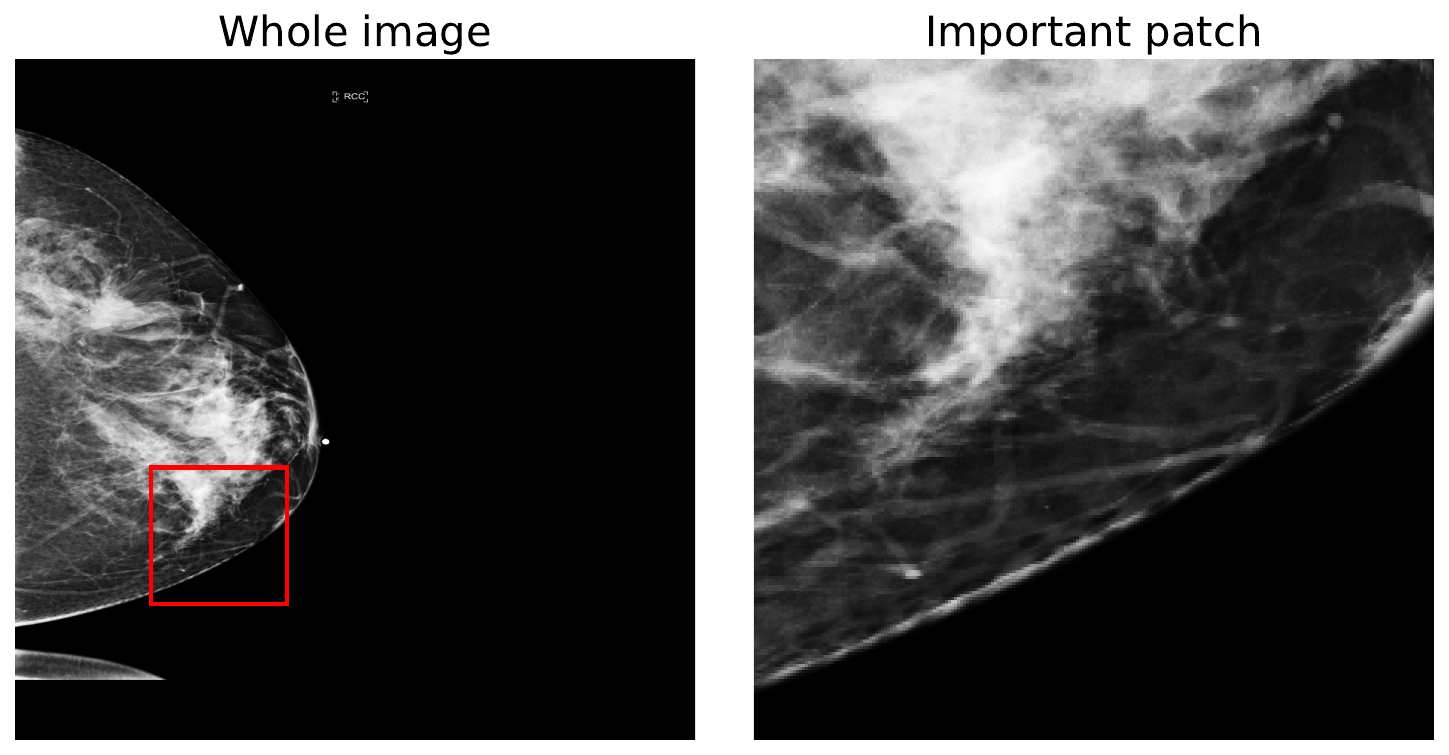}
        \caption{Image Patch 3}
    \end{subfigure}
    \hfill
    \begin{subfigure}[b]{0.45\textwidth}
        \centering
        \includegraphics[width=\linewidth]{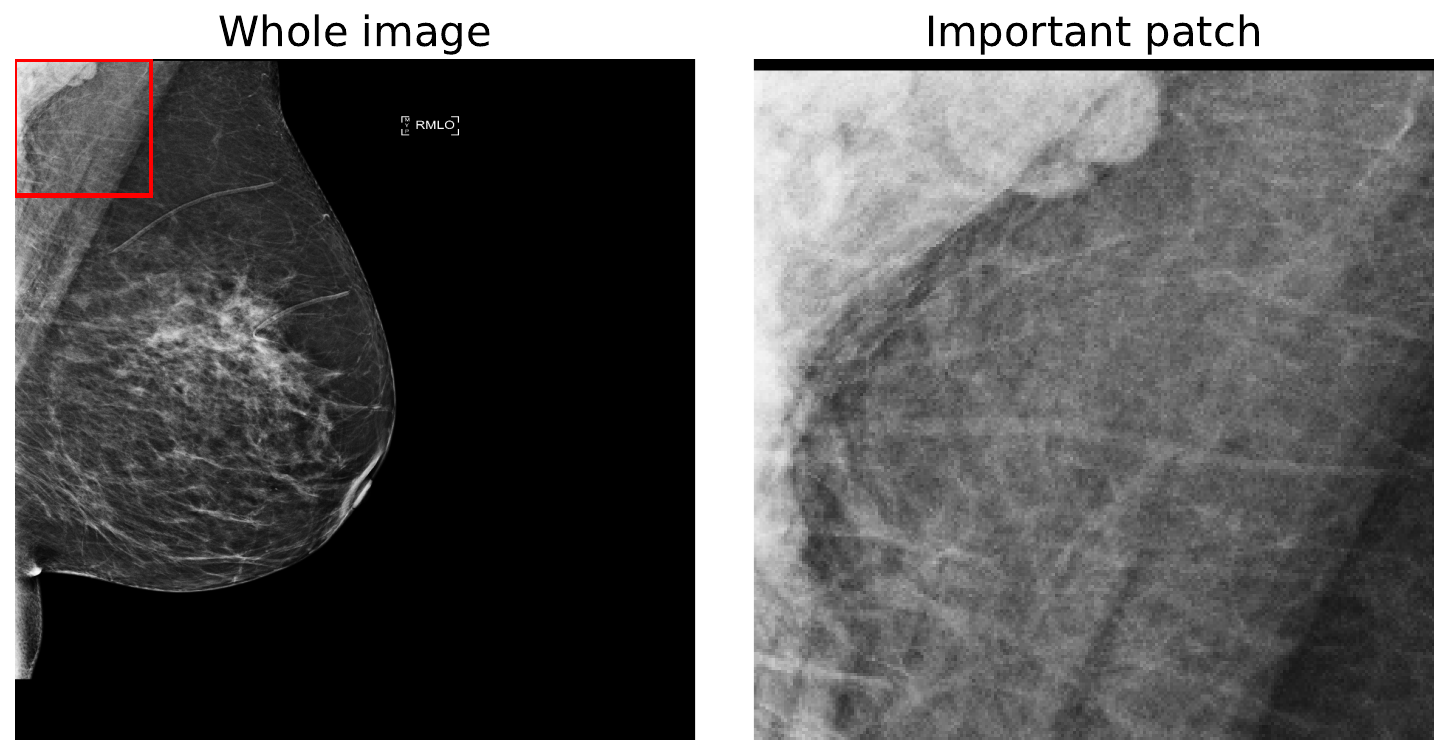}
        \caption{Image Patch 4}
    \end{subfigure}

    \vspace{1em}

    \begin{subfigure}[b]{0.45\textwidth}
        \centering
        \includegraphics[width=\linewidth]{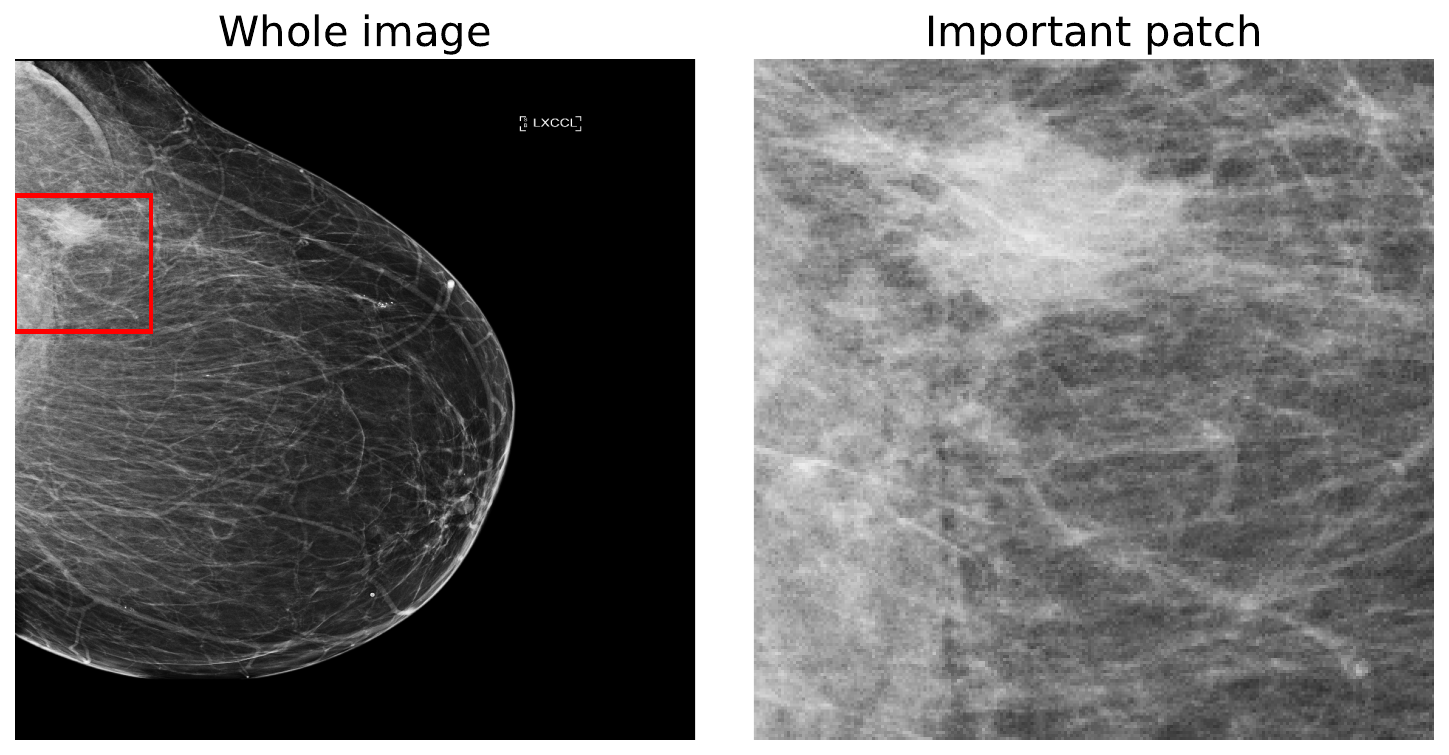}
        \caption{Image patch 5}
    \end{subfigure}
    \hfill
    \begin{subfigure}[b]{0.45\textwidth}
        \centering
        \includegraphics[width=\linewidth]{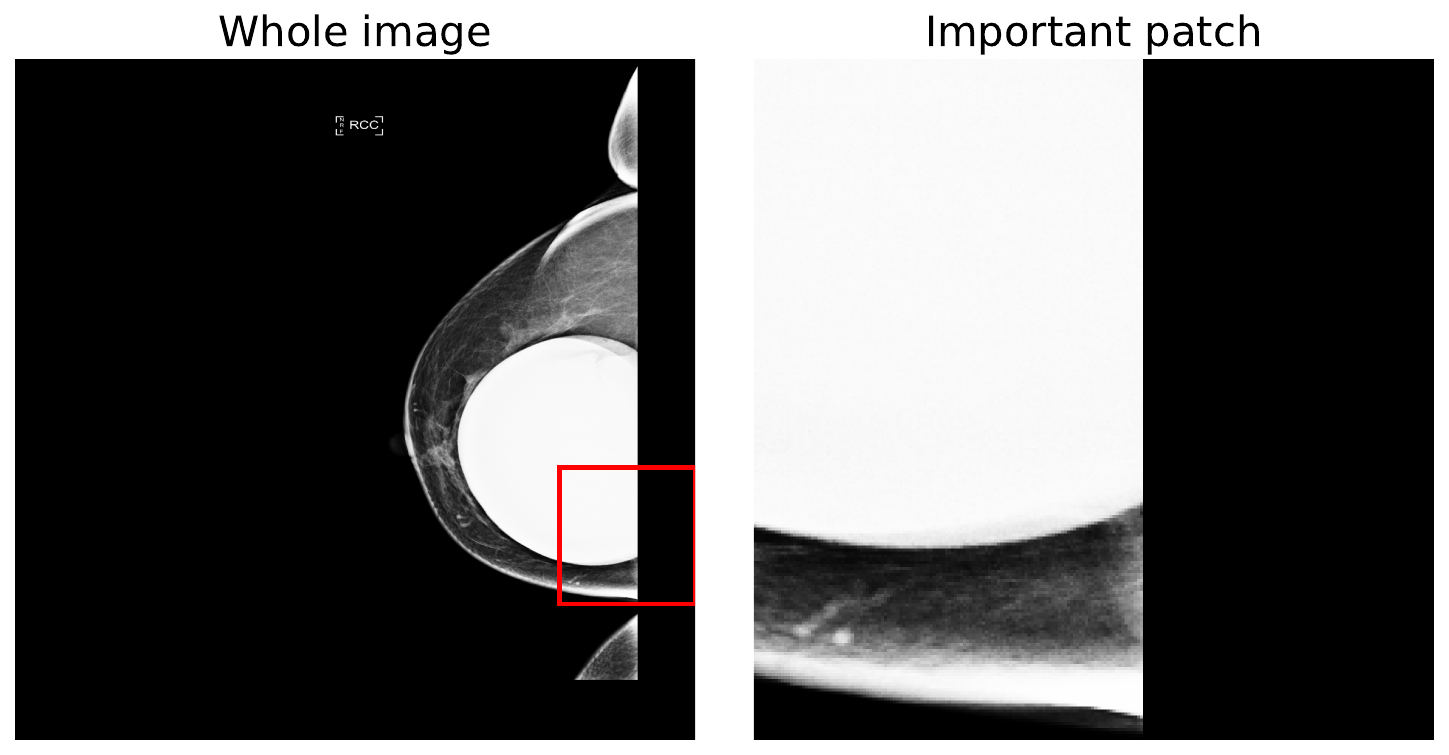}
        \caption{Image patch 6}
    \end{subfigure}

    \caption{Indicating image patches influential for detecting cancer cell of a whole slide image from training set of RSNA-SMBC dataset.}
    \label{fig:image_patch_rsna_breast_important_patch}
\end{figure}

Similarly, we can perform the same inspection for random iamges of the RSNA-SMBC dataset in Figure \ref{fig:image_patch_rsna_breast_important_patch}. Note that this dataset is not as obvious to non-medical professionals but each instance is a true-positive identification.

\begin{figure}[!h]
    \centering

    \begin{subfigure}{0.9\textwidth}
        \centering
        \includegraphics[width=\textwidth]{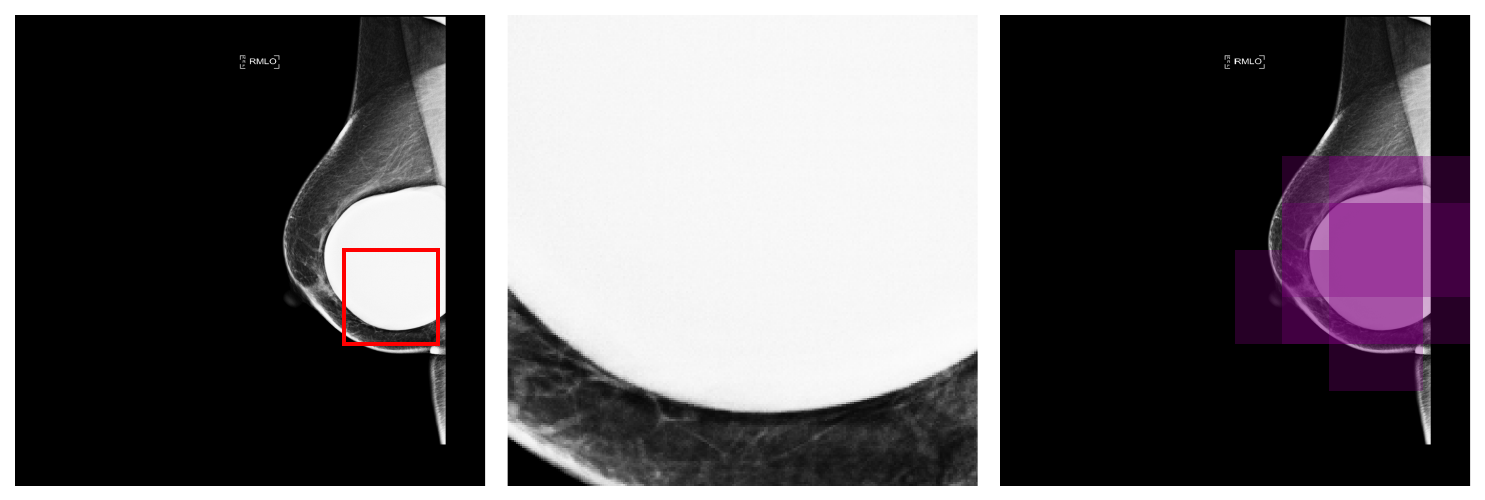}
        \label{fig:subfig4_1}
    \end{subfigure}
    
    \vspace{1em} %

    \begin{subfigure}{0.9\textwidth}
        \centering
        \includegraphics[width=\textwidth]{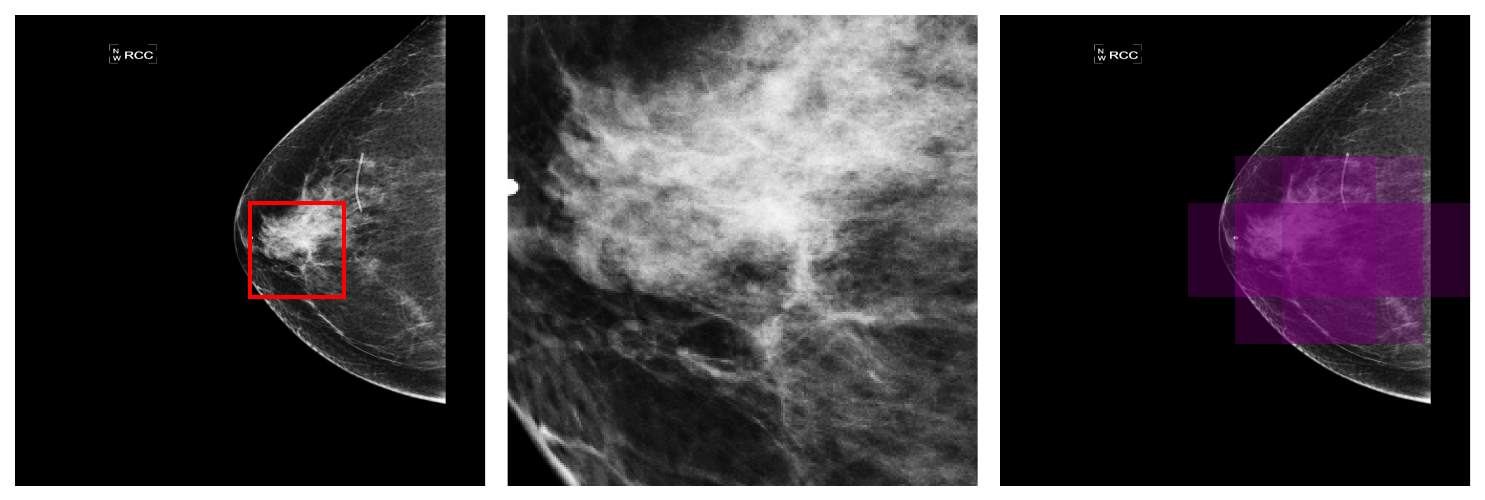}
        \label{fig:subfig4_2}
    \end{subfigure}

    \vspace{1em}

    \begin{subfigure}{0.9\textwidth}
        \centering
        \includegraphics[width=\textwidth]{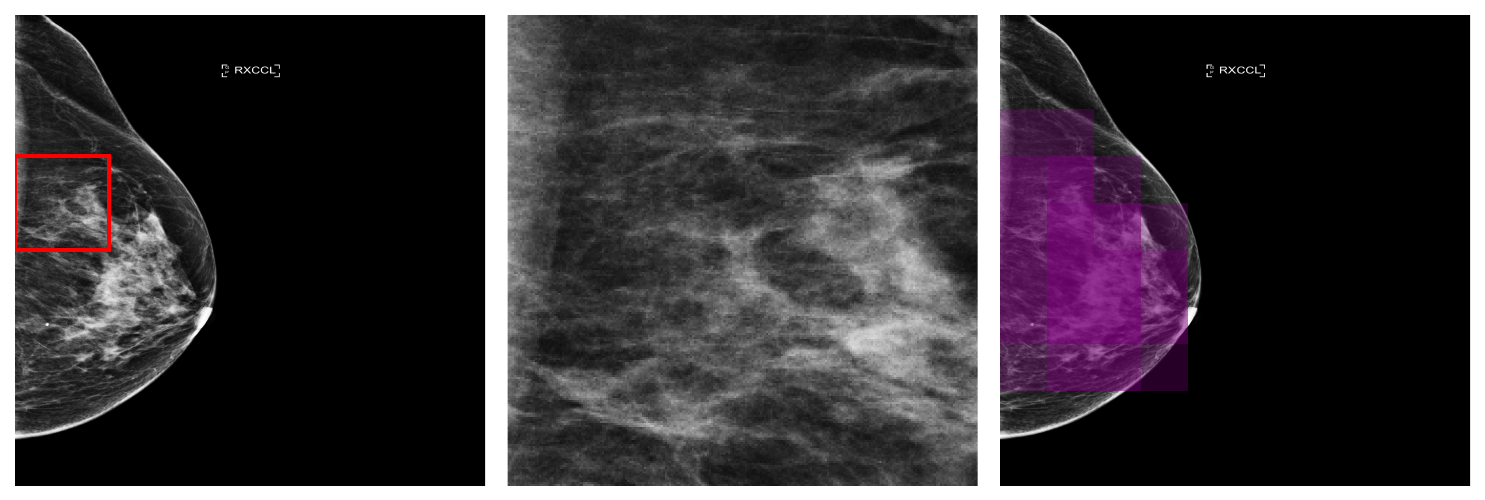}
        \label{fig:subfig4_3}
    \end{subfigure}

    \caption{Indicating image patches influential for detecting cancer cell of a whole slide image from training set of RSNA-SMBC dataset. The left most figure is the top most influential patch with a red bounding box in a whole slide image and the middle figure is the zoomed view of it. The right most figure is the overlapping patches influential for detecting tumor cell indicated by pink bounding boxes.}
    \label{fig:image_patch_breast_cancer_important_patches_overlap1}
\end{figure}

VSA-MIL further succeeds in detecting more than just the singular region of malignancy. This is confirmed by visualizing all overlapping sub-regions that are detected as positive malignancy, demonstrated in Figure \ref{fig:image_patch_breast_cancer_important_patches_overlap1}. We make note that in some cases adjusting the size of the patch may yield more intuitive results depending on the size of what needs to be detected, but did not alter this to be aligned with prior articles and the patch size previous used.

\begin{figure}[!h]
    \centering

    \begin{subfigure}[b]{0.22\textwidth}
        \centering
        \includegraphics[width=\linewidth]{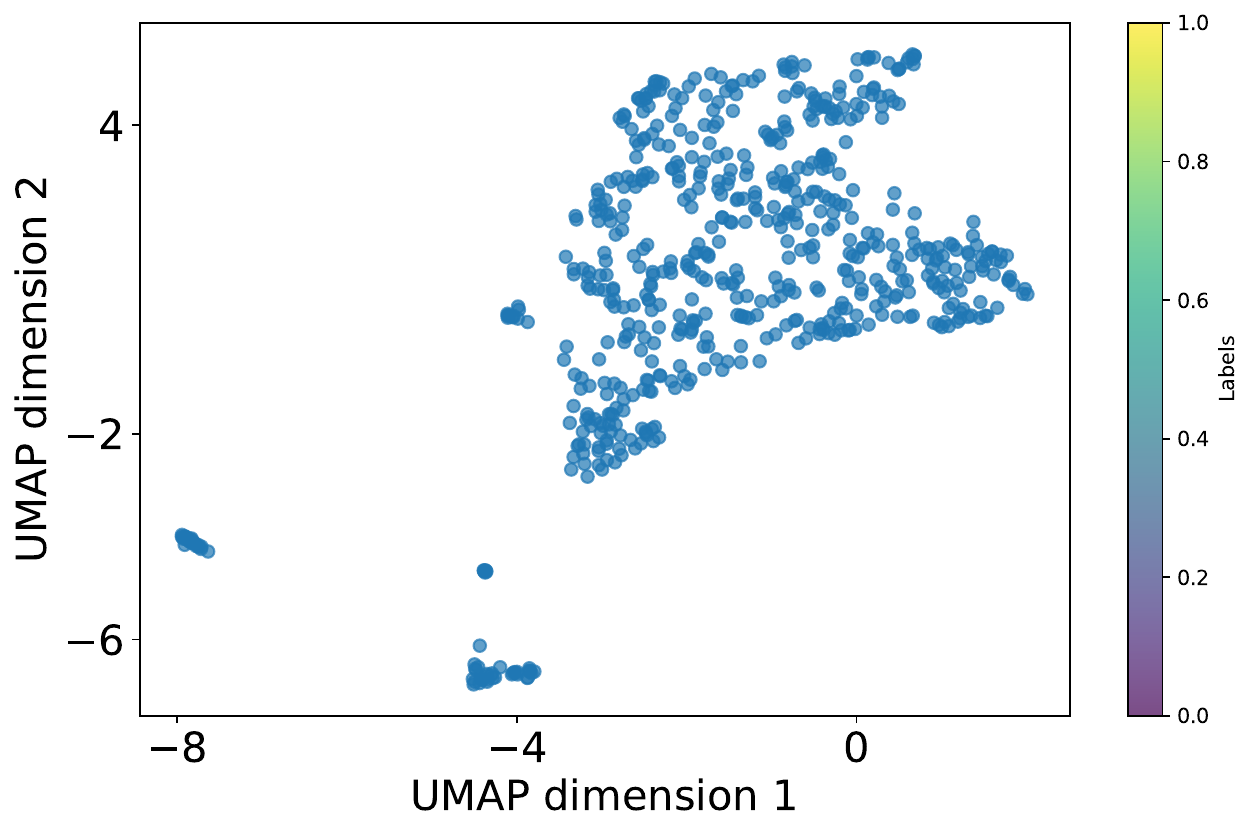}
        \caption{Elephant}
    \end{subfigure}
    \hfill
    \begin{subfigure}[b]{0.22\textwidth}
        \centering
        \includegraphics[width=\linewidth]{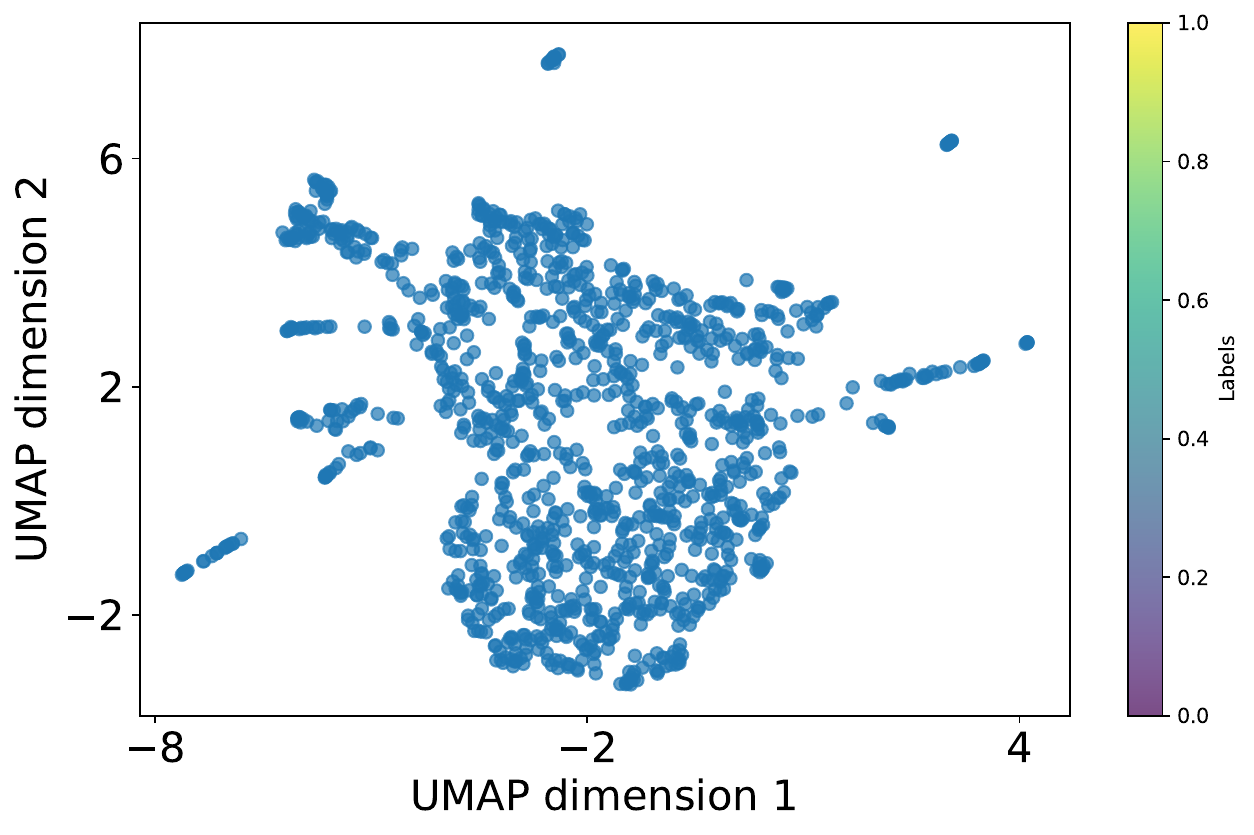}
        \caption{Protein}
    \end{subfigure}
    \hfill
    \begin{subfigure}[b]{0.22\textwidth}
        \centering
        \includegraphics[width=\linewidth]{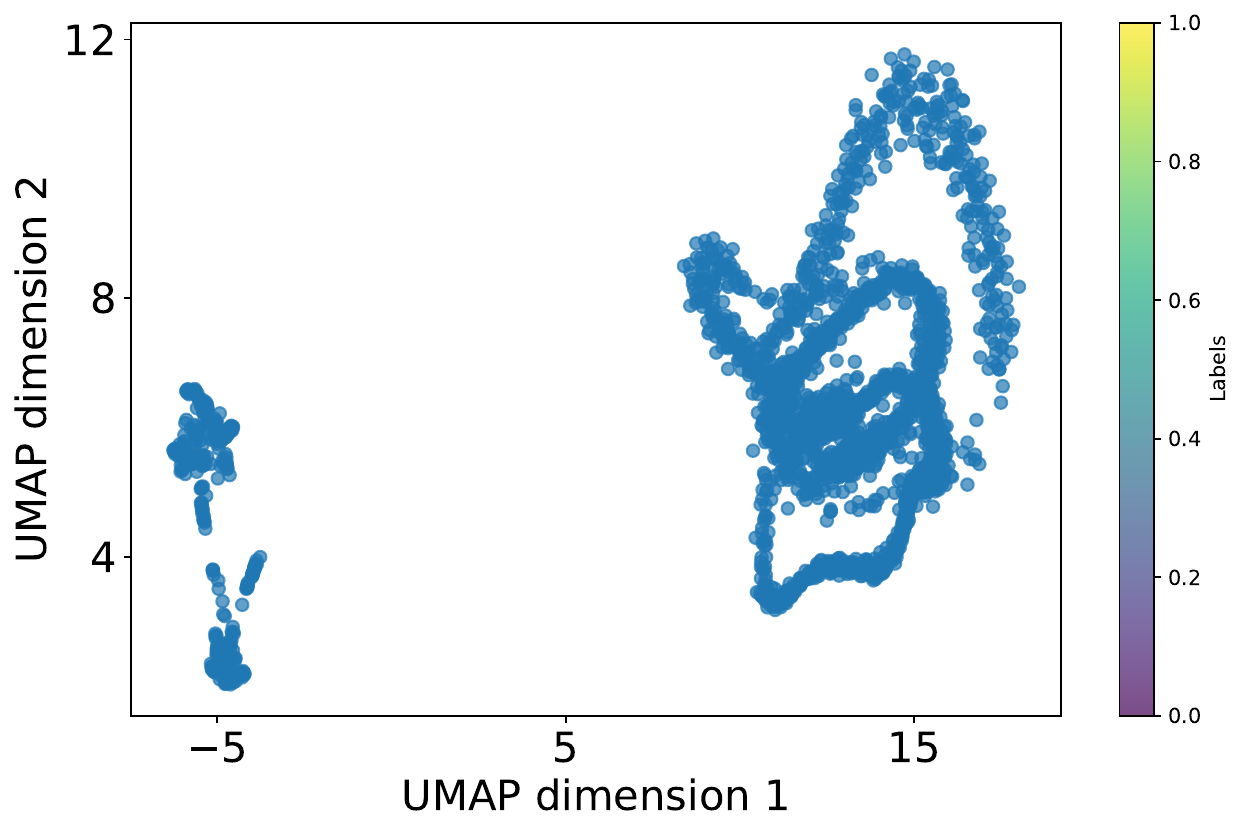}
        \caption{MUSK1}
    \end{subfigure}
    \hfill
    \begin{subfigure}[b]{0.22\textwidth}
        \centering
        \includegraphics[width=\linewidth]{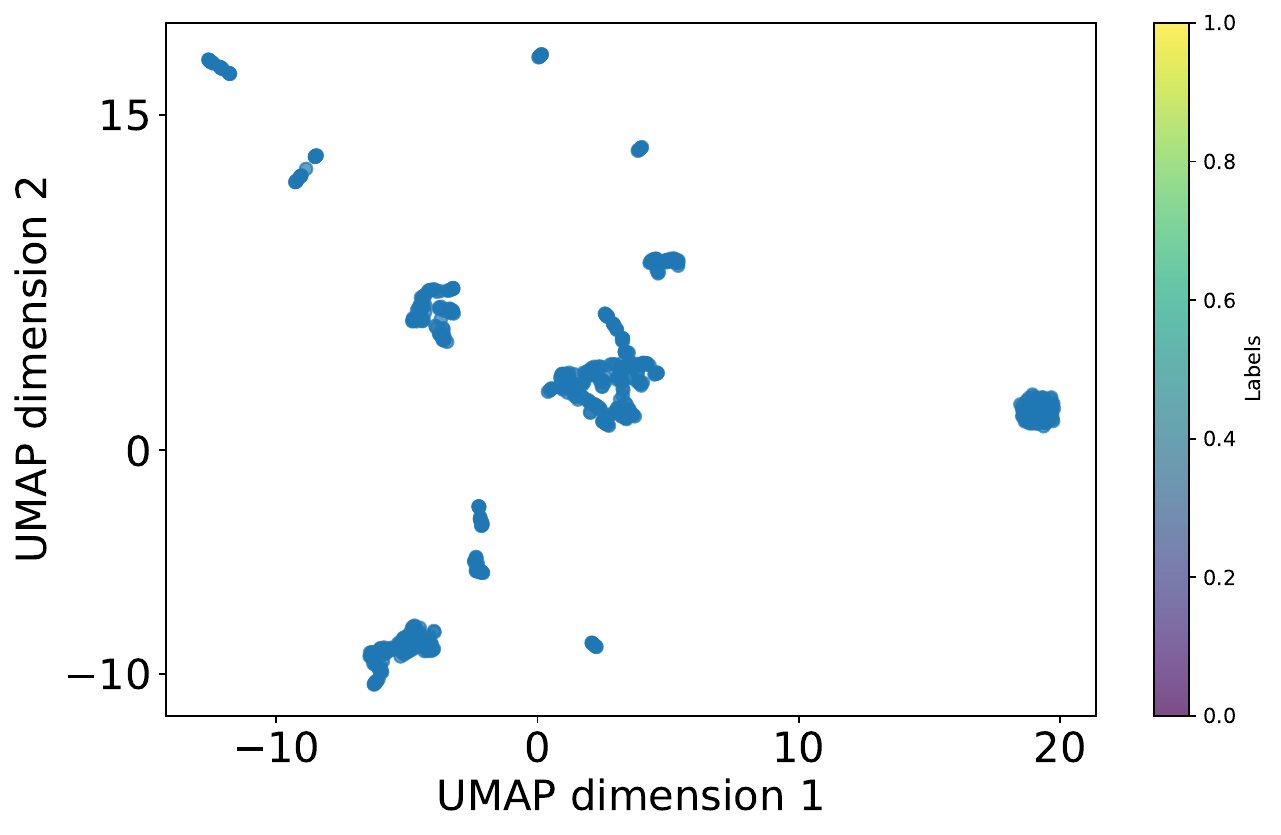}
        \caption{MUSK2}
    \end{subfigure}

    \vspace{1em}

    \begin{subfigure}[b]{0.22\textwidth}
        \centering
        \includegraphics[width=\linewidth]{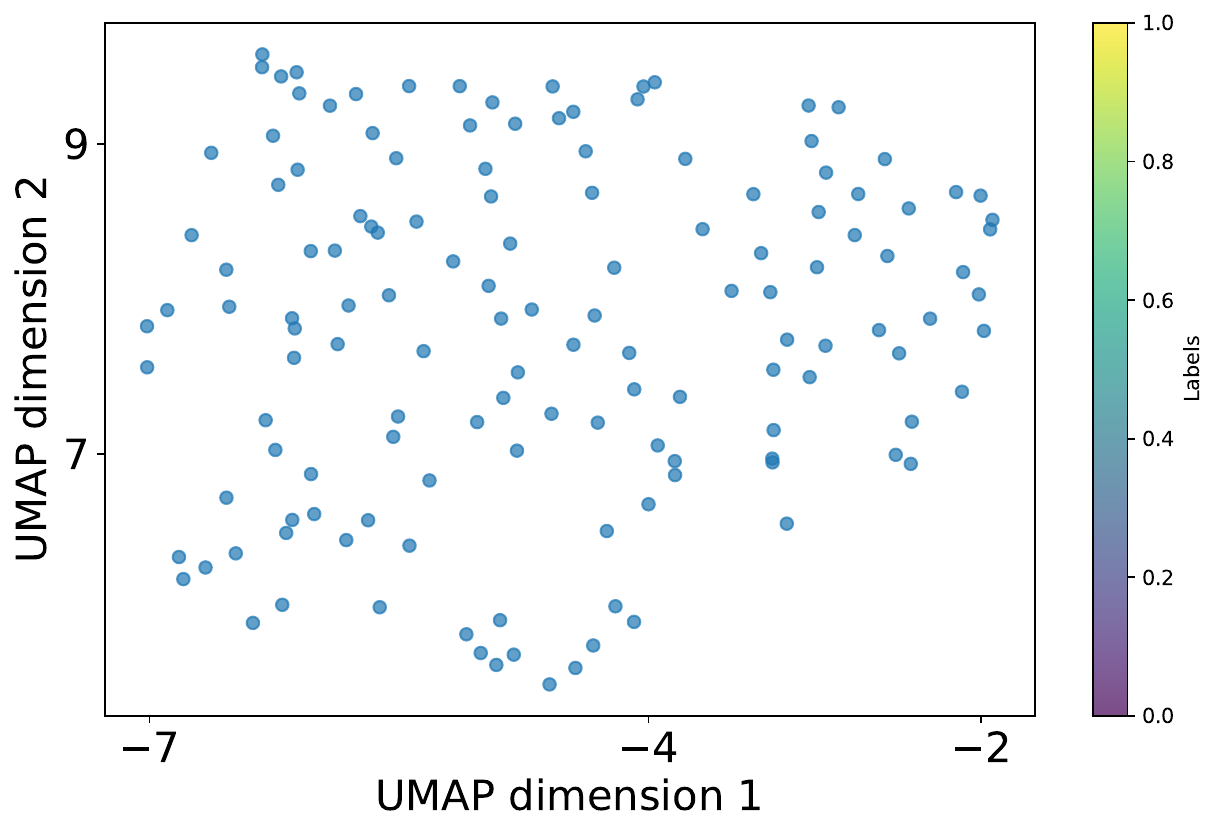}
        \caption{UCSB Breast Cancer}
    \end{subfigure}
    \hfill
    \begin{subfigure}[b]{0.22\textwidth}
        \centering
        \includegraphics[width=\linewidth]{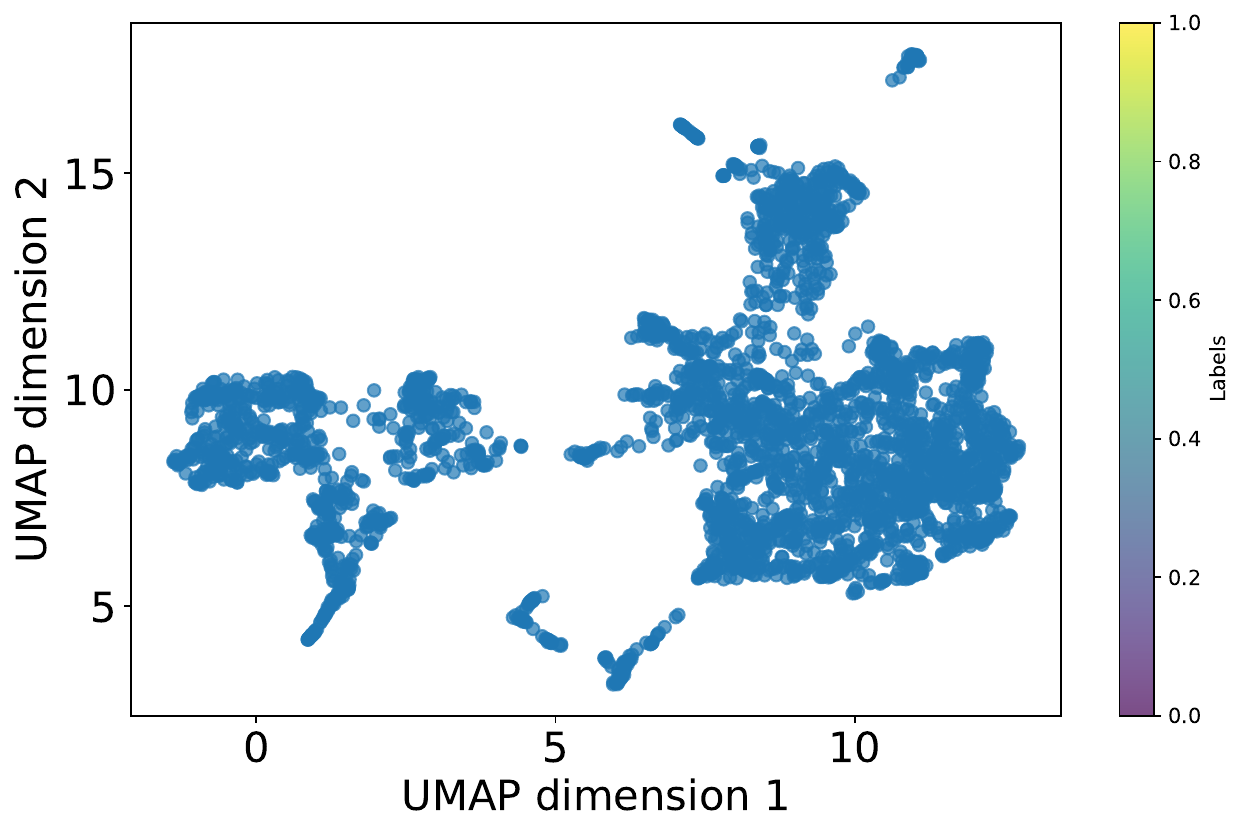}
        \caption{Birds Brown Creeper}
    \end{subfigure}
    \hfill
    \begin{subfigure}[b]{0.22\textwidth}
        \centering
        \includegraphics[width=\linewidth]{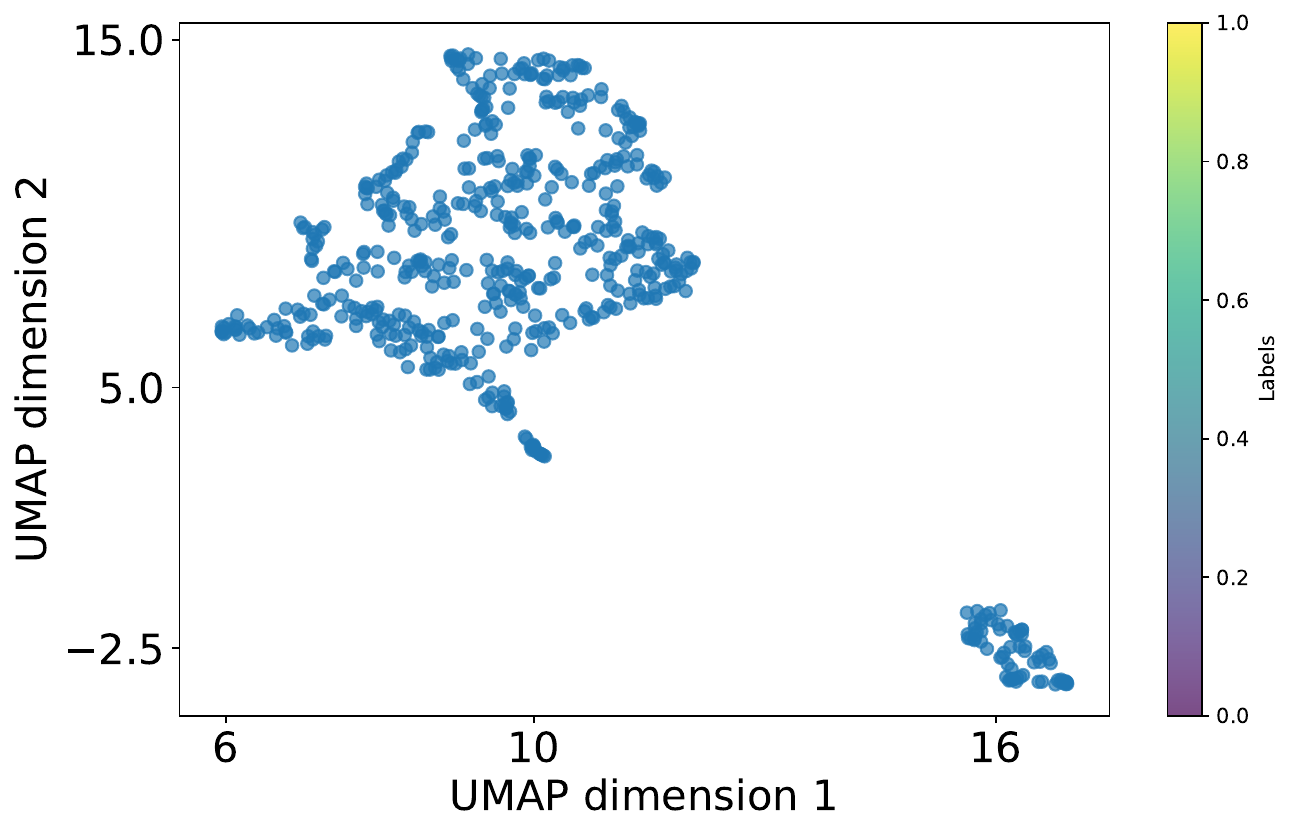}
        \caption{Web Recommend}
    \end{subfigure}
    \hfill
    \begin{subfigure}[b]{0.22\textwidth}
        \centering
        \includegraphics[width=\linewidth]{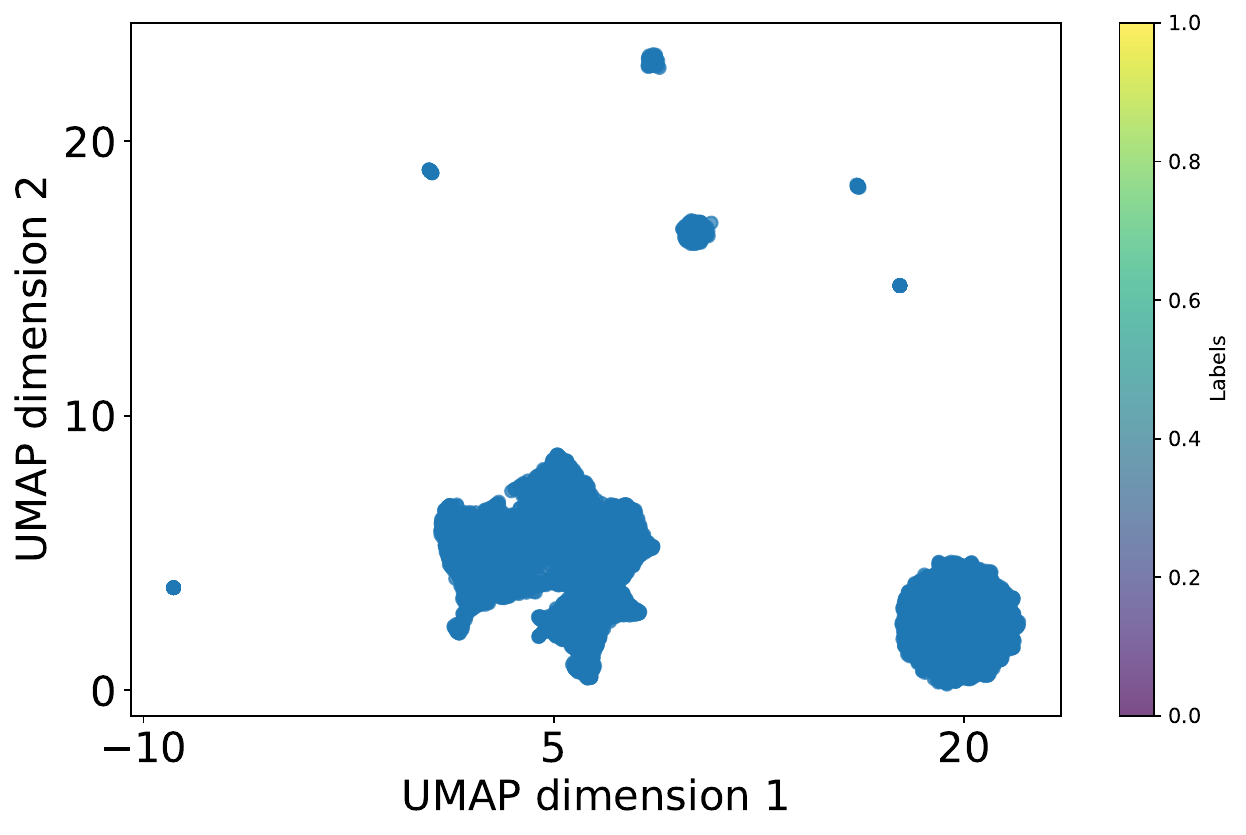}
        \caption{Corel Dogs}
    \end{subfigure}
    
    \caption{UMAP projection of VSA embeddings from the training dataset of the traditional benchmark MIL datasets.}
    \label{fig:umap_projection_all_MIL_datasets}
\end{figure}

\begin{figure}[!h]
    \centering

    \begin{subfigure}[b]{0.22\textwidth}
        \centering
        \includegraphics[width=\linewidth]{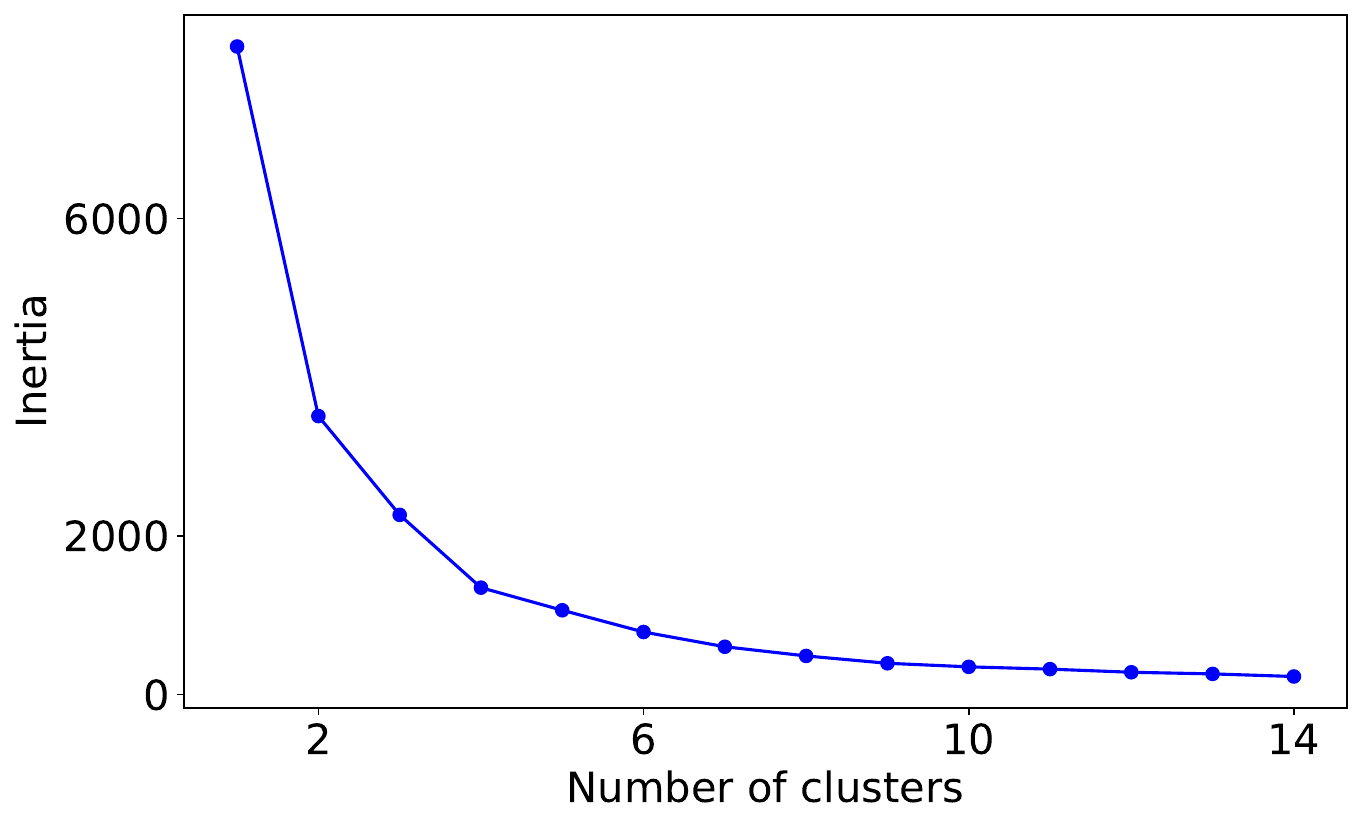}
        \caption{Elephant}
    \end{subfigure}
    \hfill
    \begin{subfigure}[b]{0.22\textwidth}
        \centering
        \includegraphics[width=\linewidth]{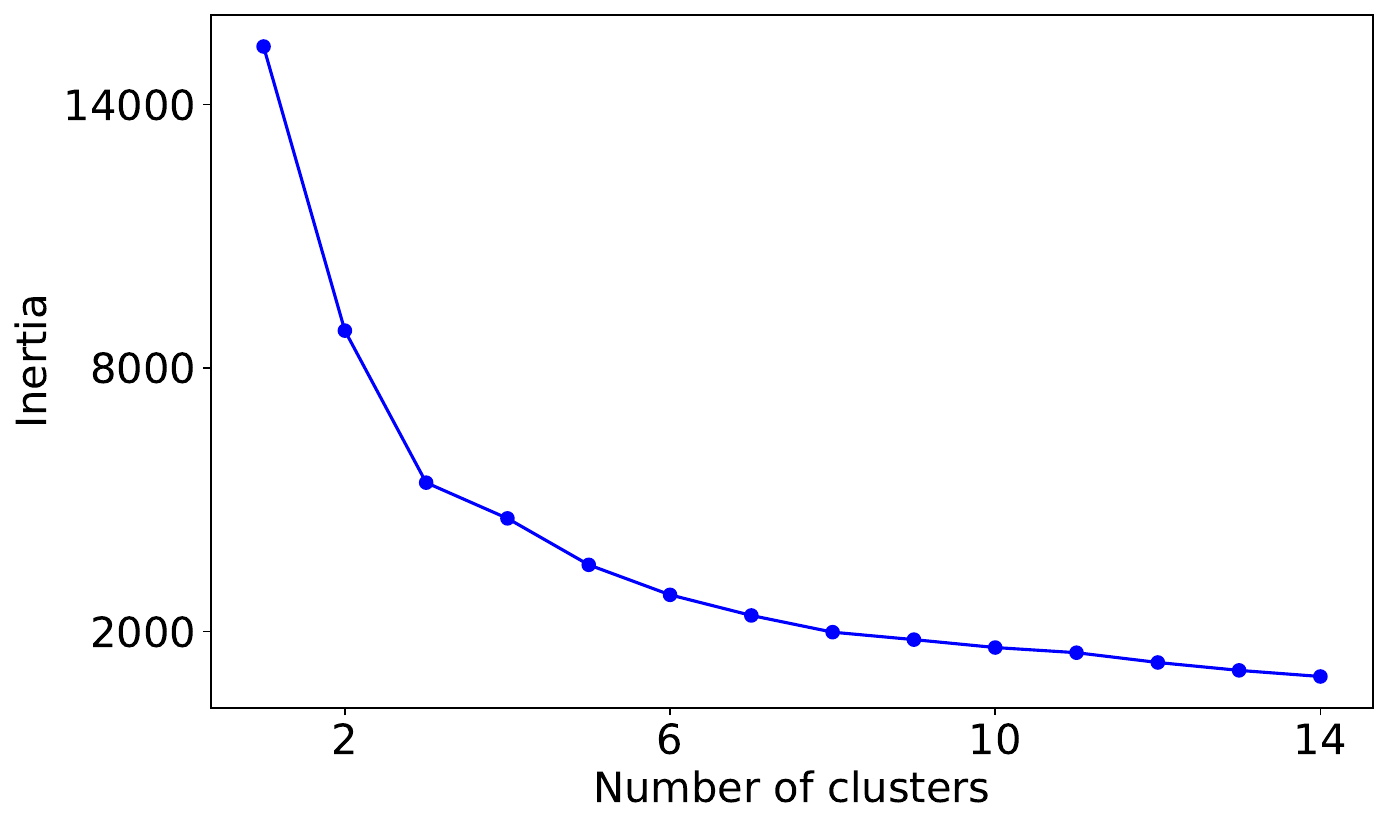}
        \caption{Protein}
    \end{subfigure}
    \hfill
    \begin{subfigure}[b]{0.22\textwidth}
        \centering
        \includegraphics[width=\linewidth]{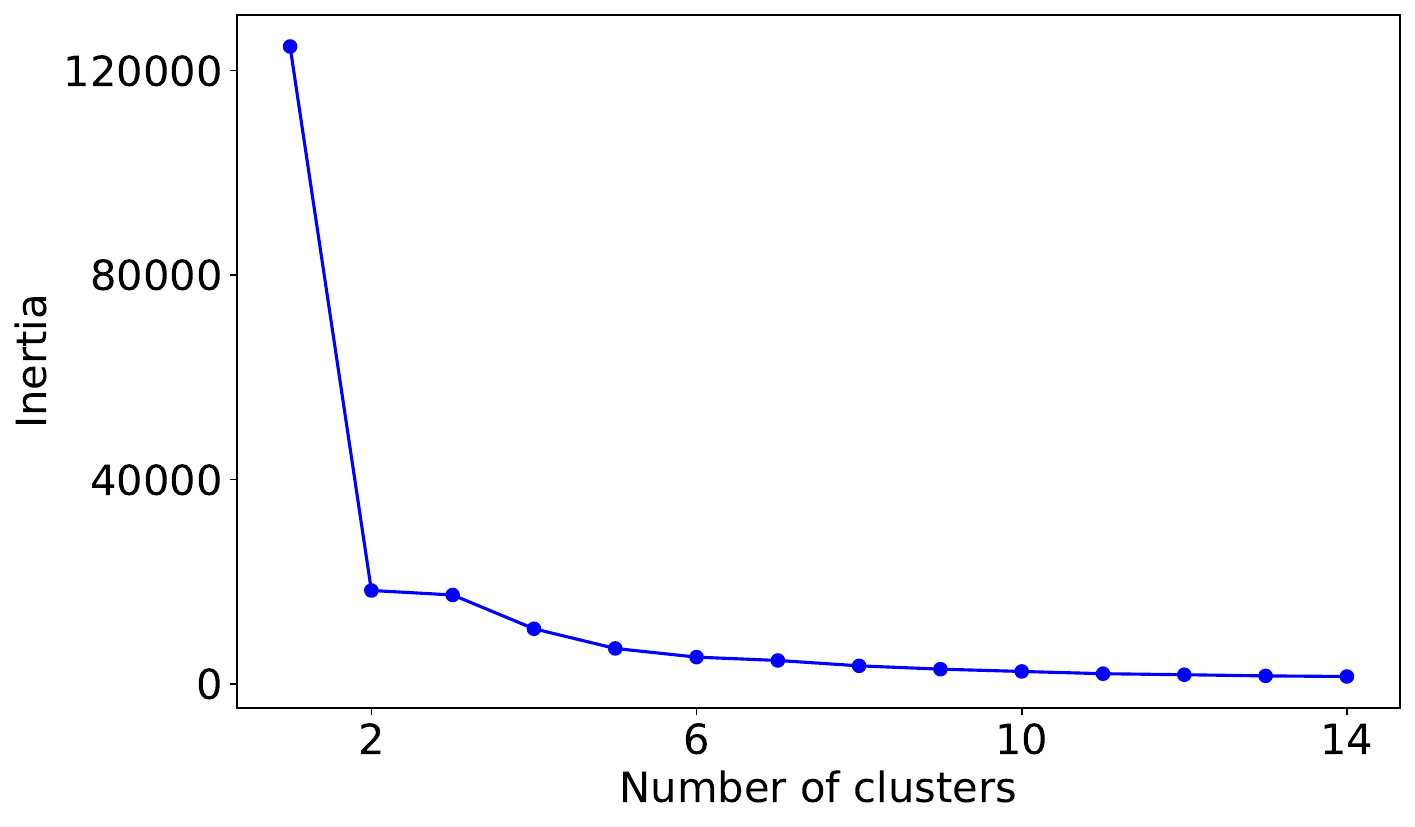}
        \caption{MUSK1}
    \end{subfigure}
    \hfill
    \begin{subfigure}[b]{0.22\textwidth}
        \centering
        \includegraphics[width=\linewidth]{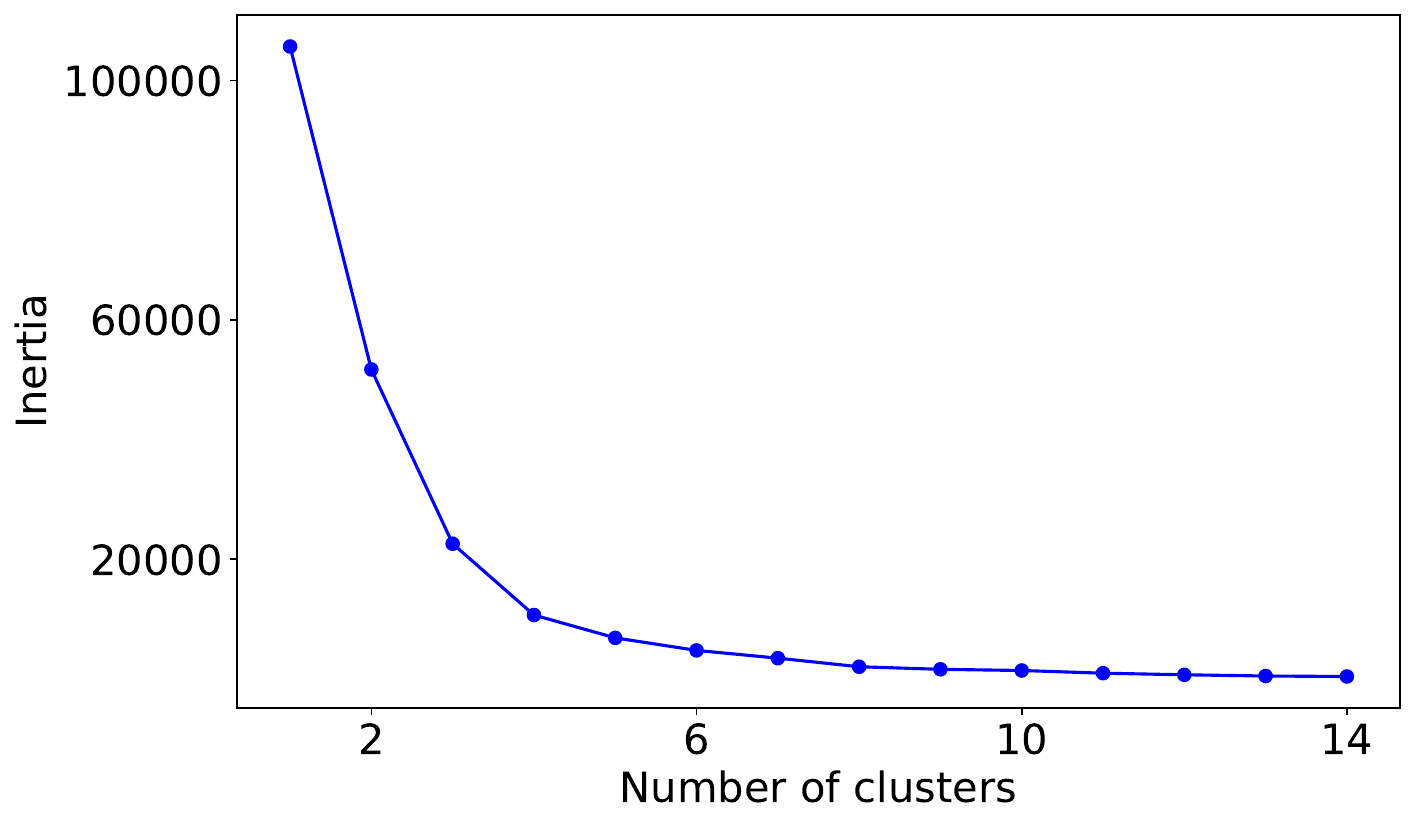}
        \caption{MUSK2}
    \end{subfigure}

    \vspace{1em}

    \begin{subfigure}[b]{0.22\textwidth}
        \centering
        \includegraphics[width=\linewidth]{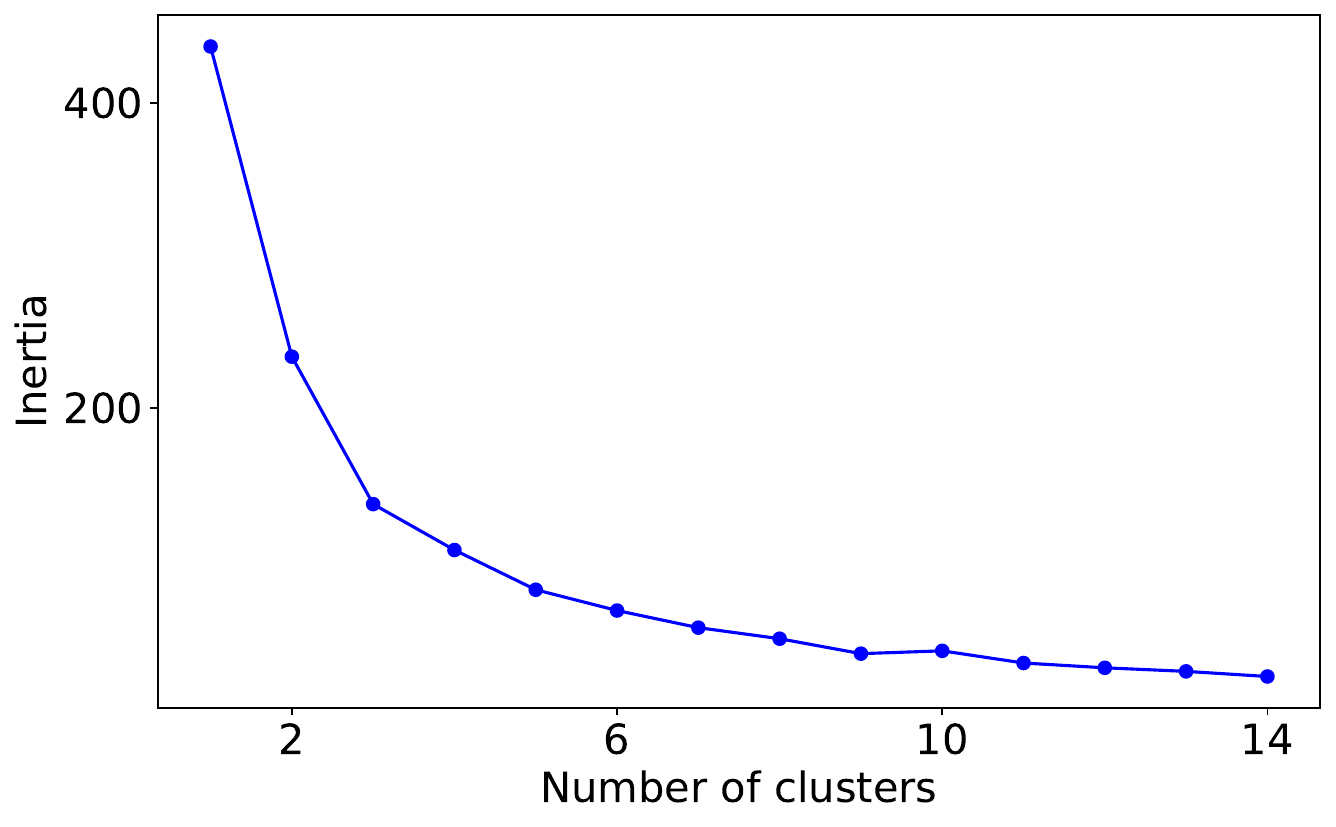}
        \caption{UCSB Breast Cancer}
    \end{subfigure}
    \hfill
    \begin{subfigure}[b]{0.22\textwidth}
        \centering
        \includegraphics[width=\linewidth]{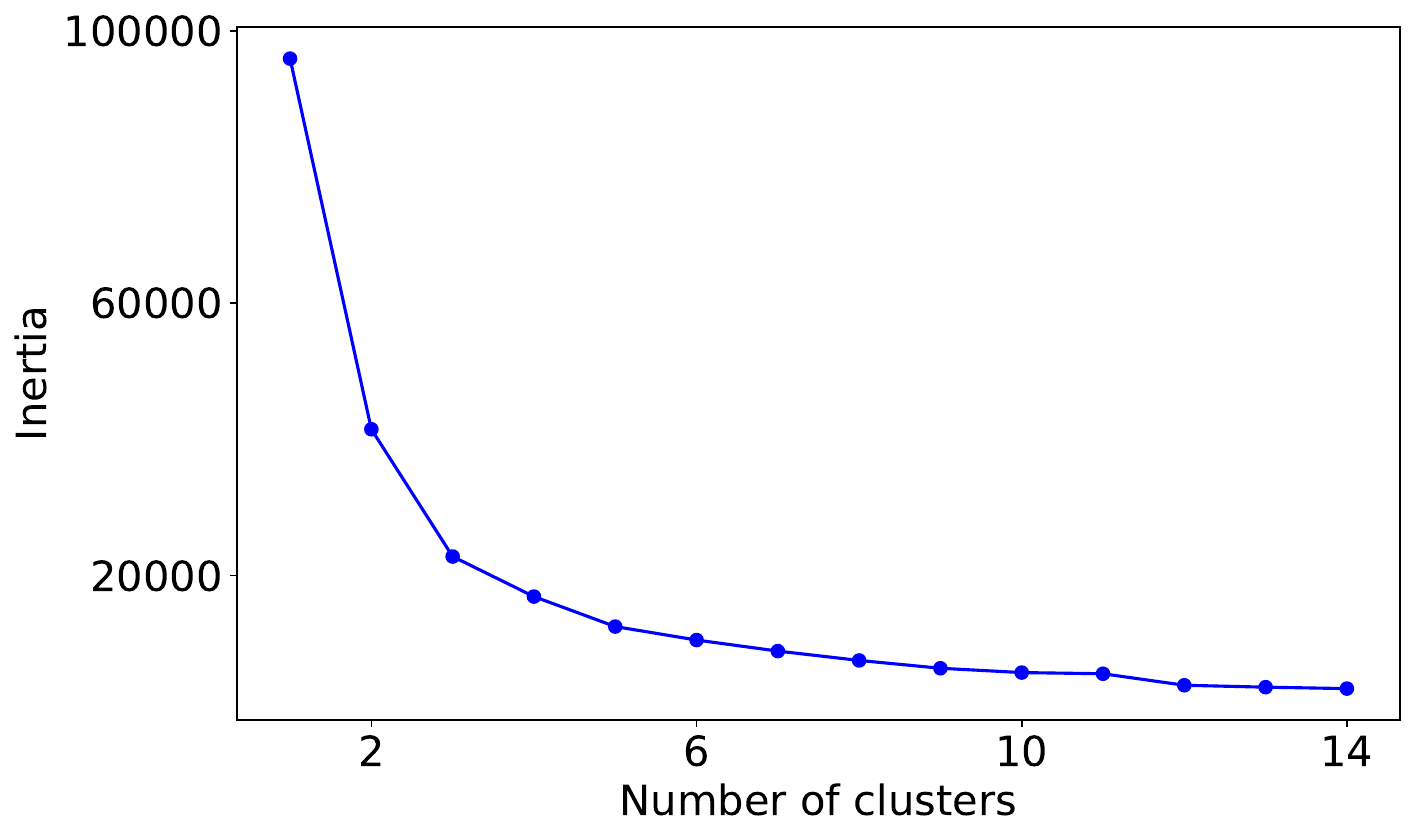}
        \caption{Birds Brown Creeper}
    \end{subfigure}
    \hfill
    \begin{subfigure}[b]{0.22\textwidth}
        \centering
        \includegraphics[width=\linewidth]{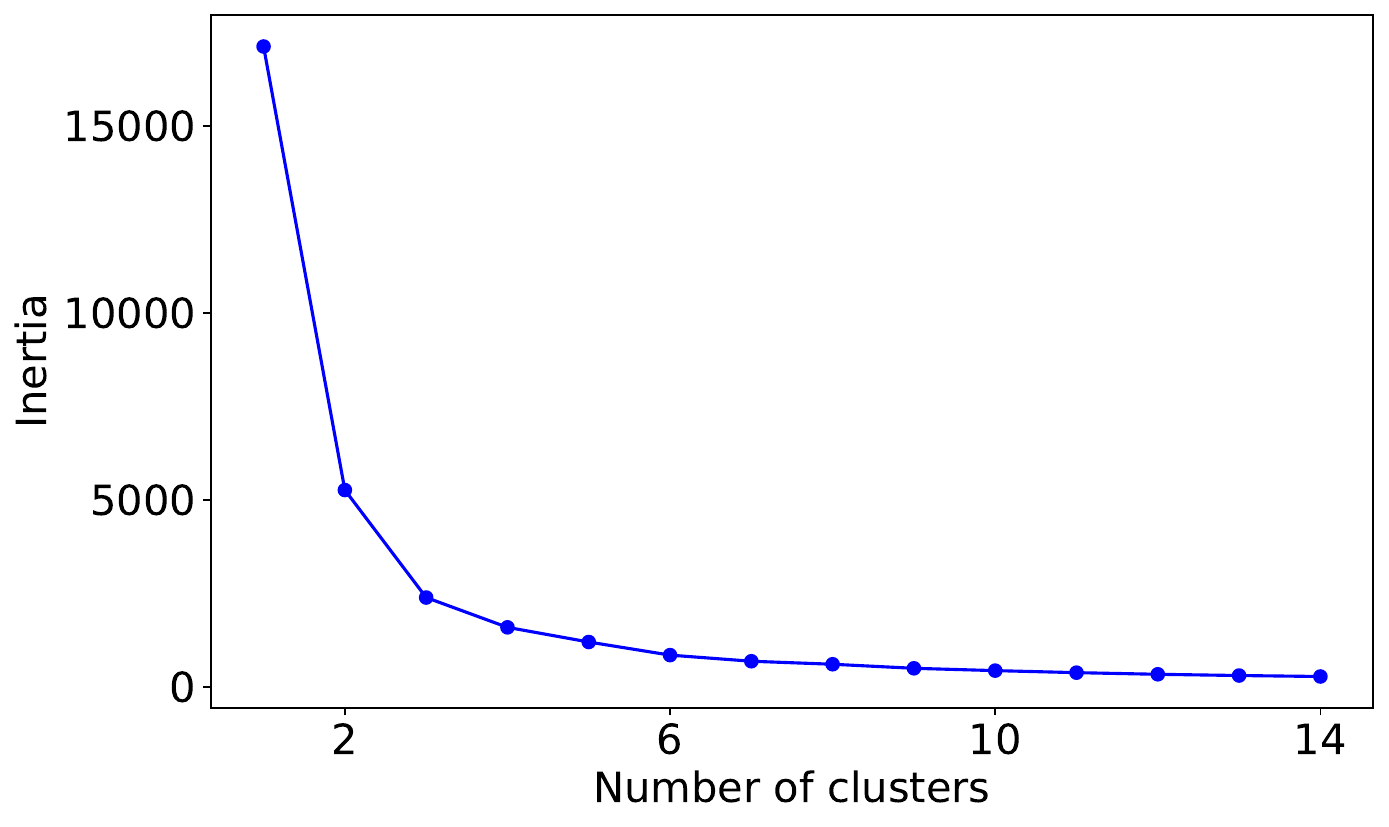}
        \caption{Web Recommend}
    \end{subfigure}
    \hfill
    \begin{subfigure}[b]{0.22\textwidth}
        \centering
        \includegraphics[width=\linewidth]{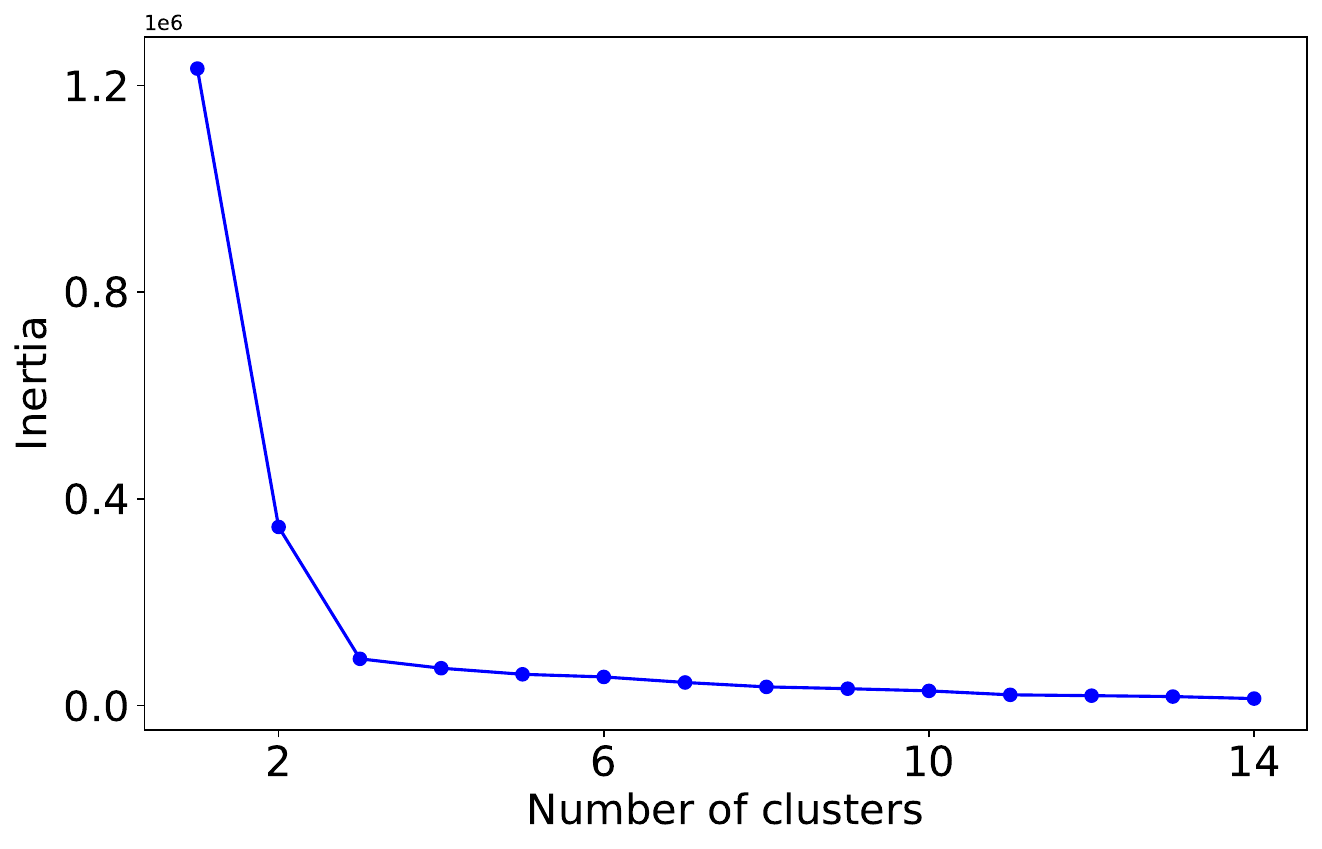}
        \caption{Corel Dogs}
    \end{subfigure}
    \caption{Analysis of the optimal cluster number using the elbow method of the k-means clustering algorithm for the VSA embeddings from the training dataset of the traditional benchmark MIL datasets.}
    \label{fig:Elbow_method_all_MIL_datasets}
\end{figure}

\begin{figure}[!h]
    \centering

    \begin{subfigure}[b]{0.22\textwidth}
        \centering
        \includegraphics[width=\linewidth]{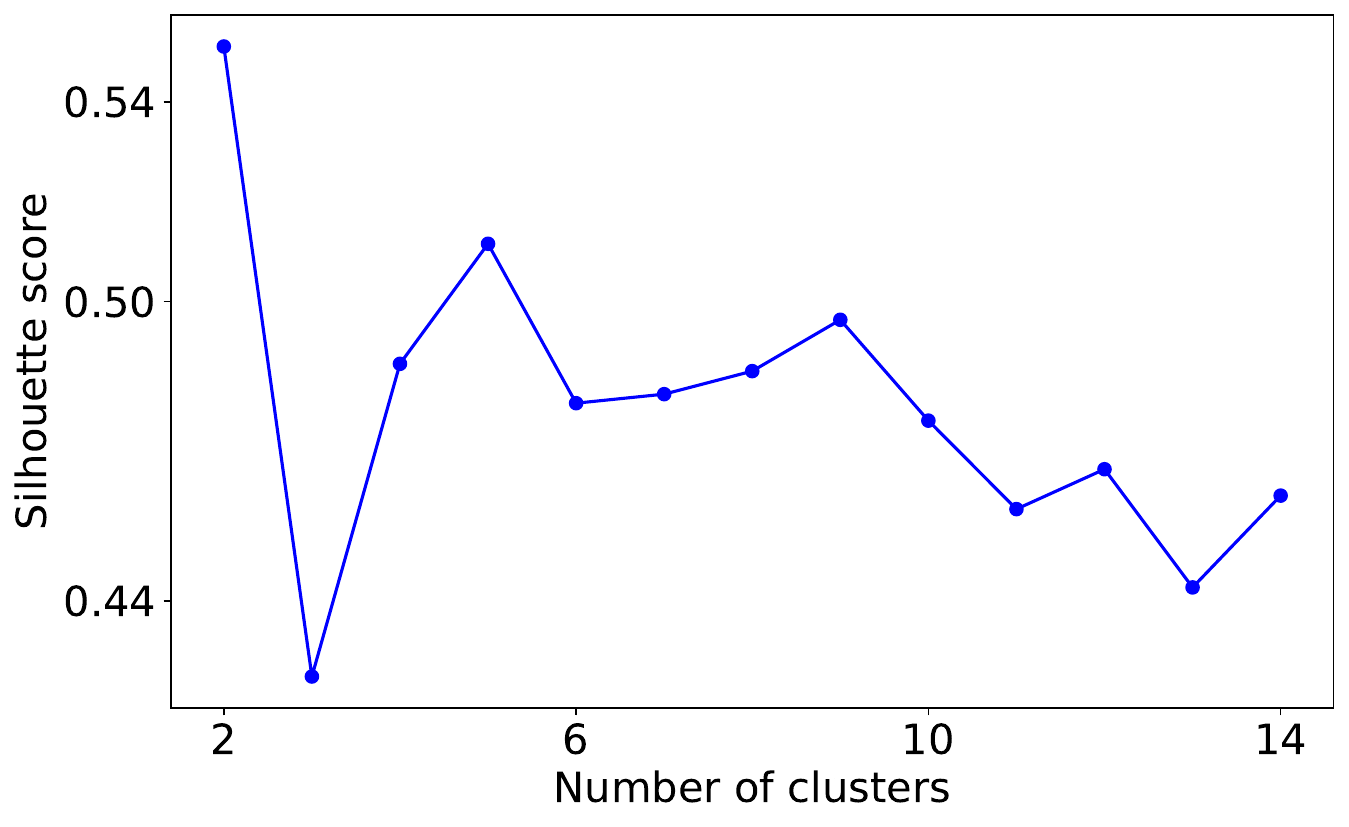}
        \caption{Elephant}
    \end{subfigure}
    \hfill
    \begin{subfigure}[b]{0.22\textwidth}
        \centering
        \includegraphics[width=\linewidth]{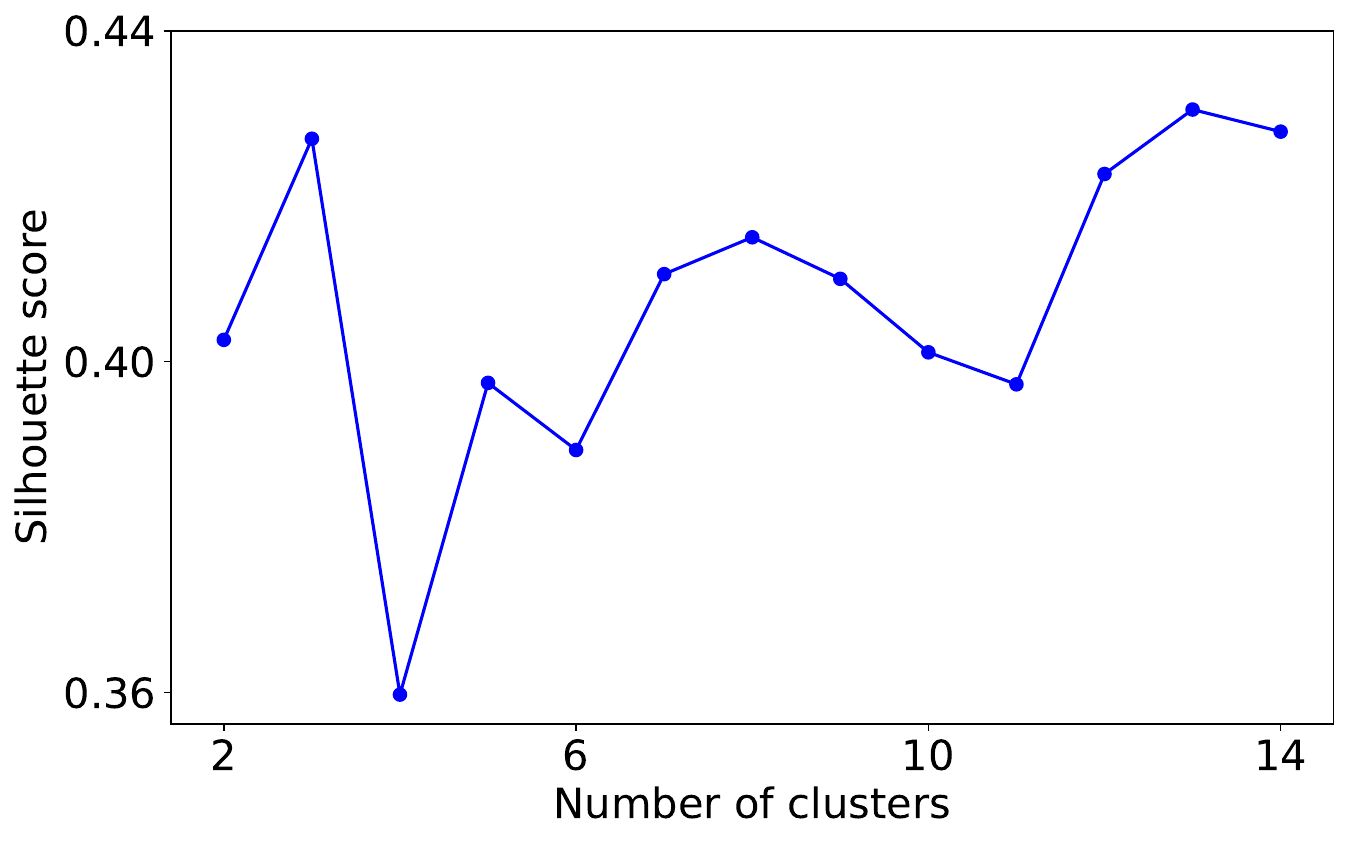}
        \caption{Protein}
    \end{subfigure}
    \hfill
    \begin{subfigure}[b]{0.22\textwidth}
        \centering
        \includegraphics[width=\linewidth]{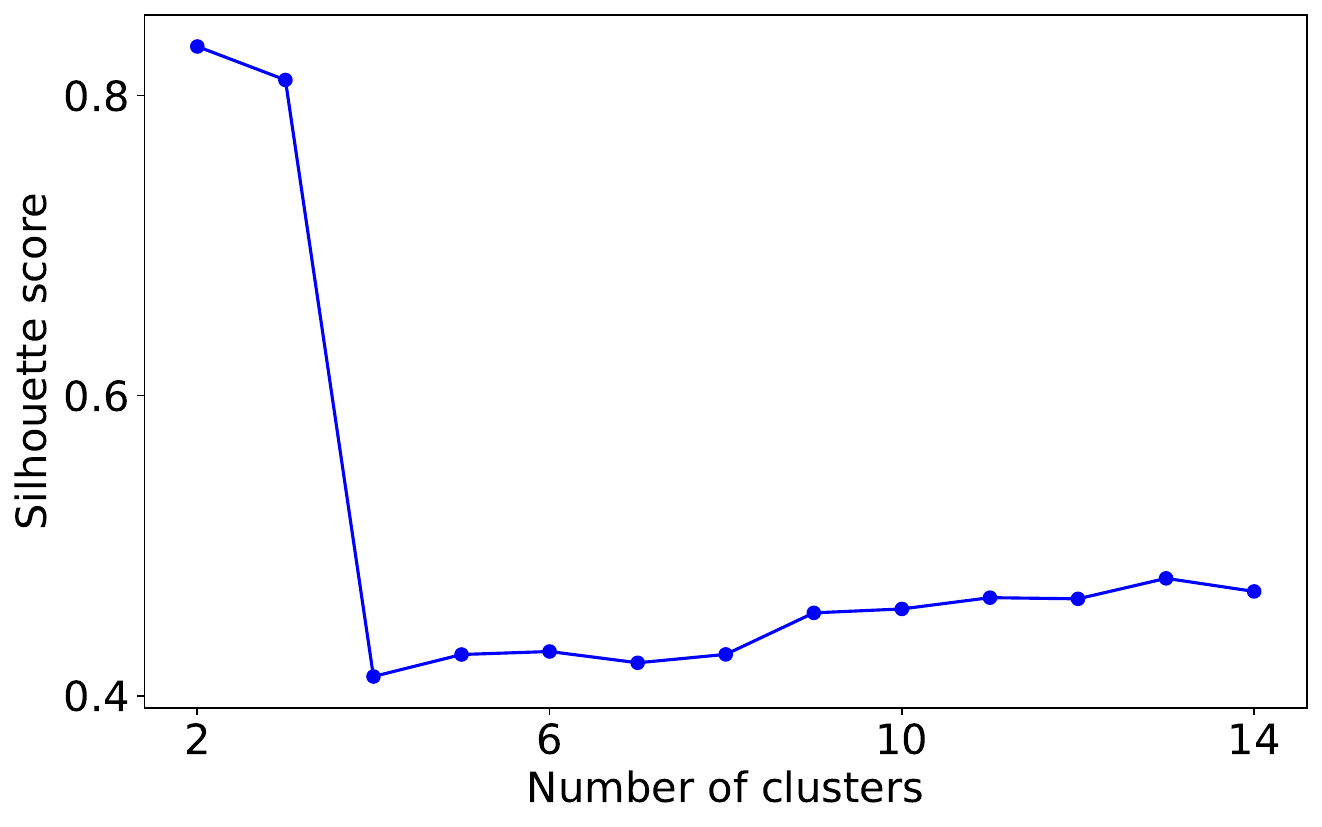}
        \caption{MUSK1}
    \end{subfigure}
    \hfill
    \begin{subfigure}[b]{0.22\textwidth}
        \centering
        \includegraphics[width=\linewidth]{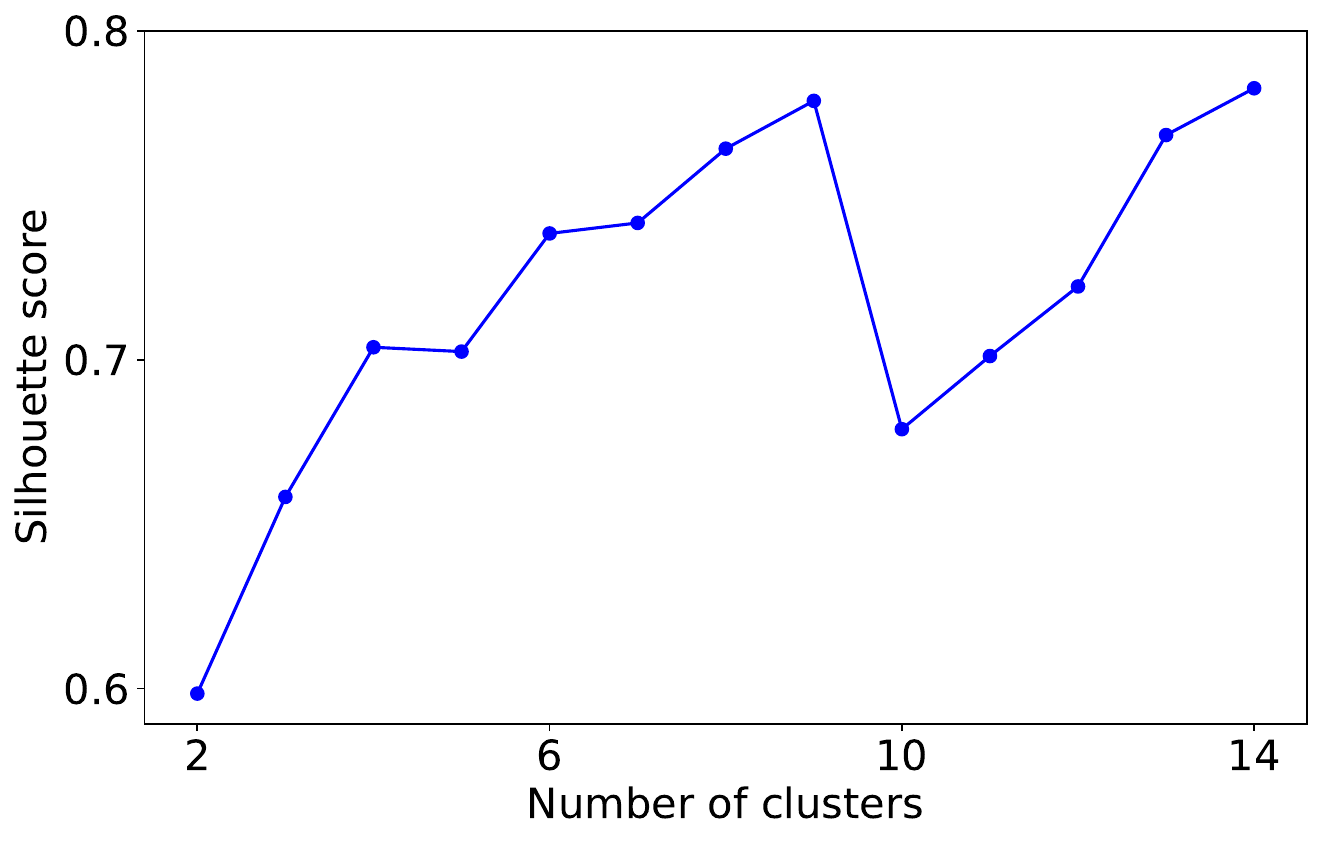}
        \caption{MUSK2}
    \end{subfigure}

    \vspace{1em}

    \begin{subfigure}[b]{0.22\textwidth}
        \centering
        \includegraphics[width=\linewidth]{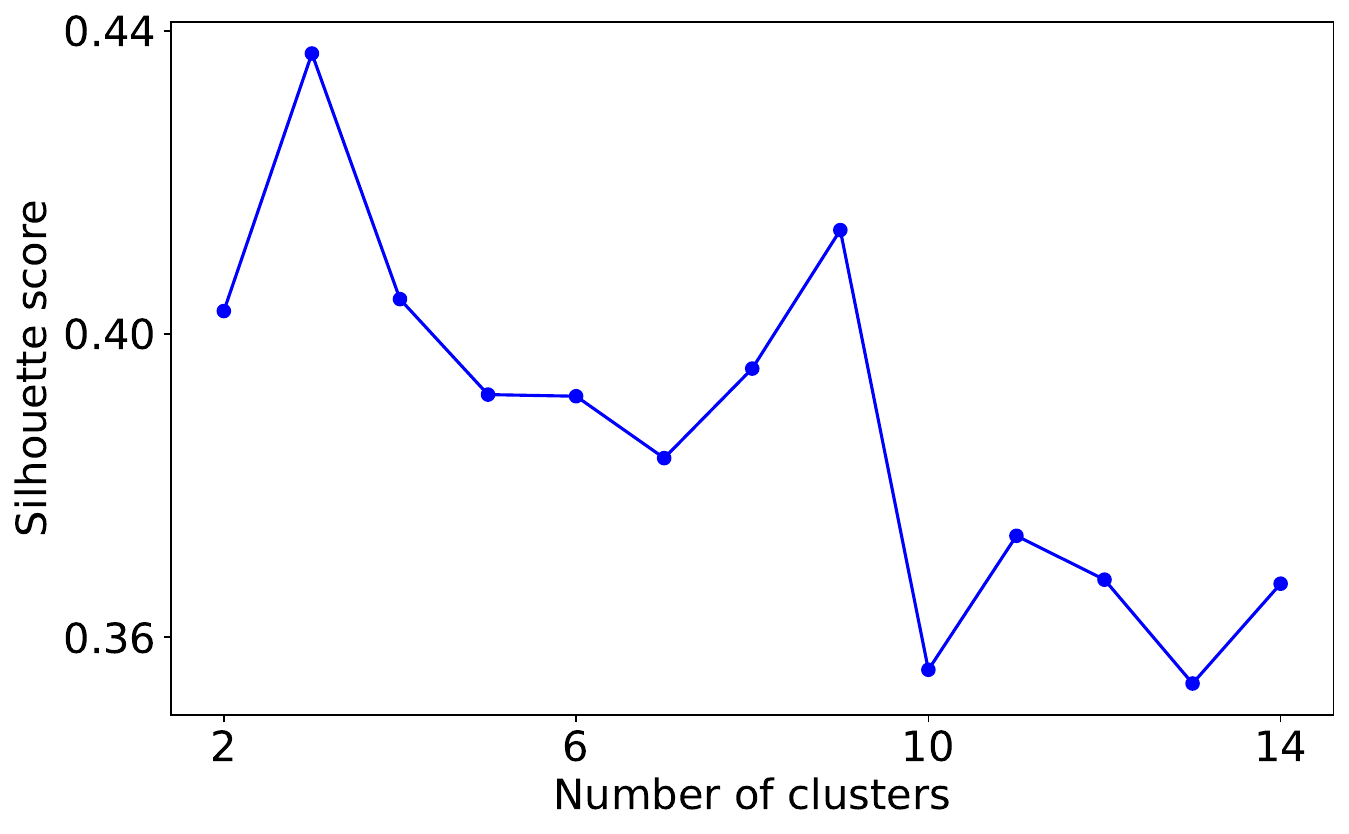}
        \caption{UCSB Breast Cancer}
    \end{subfigure}
    \hfill
    \begin{subfigure}[b]{0.22\textwidth}
        \centering
        \includegraphics[width=\linewidth]{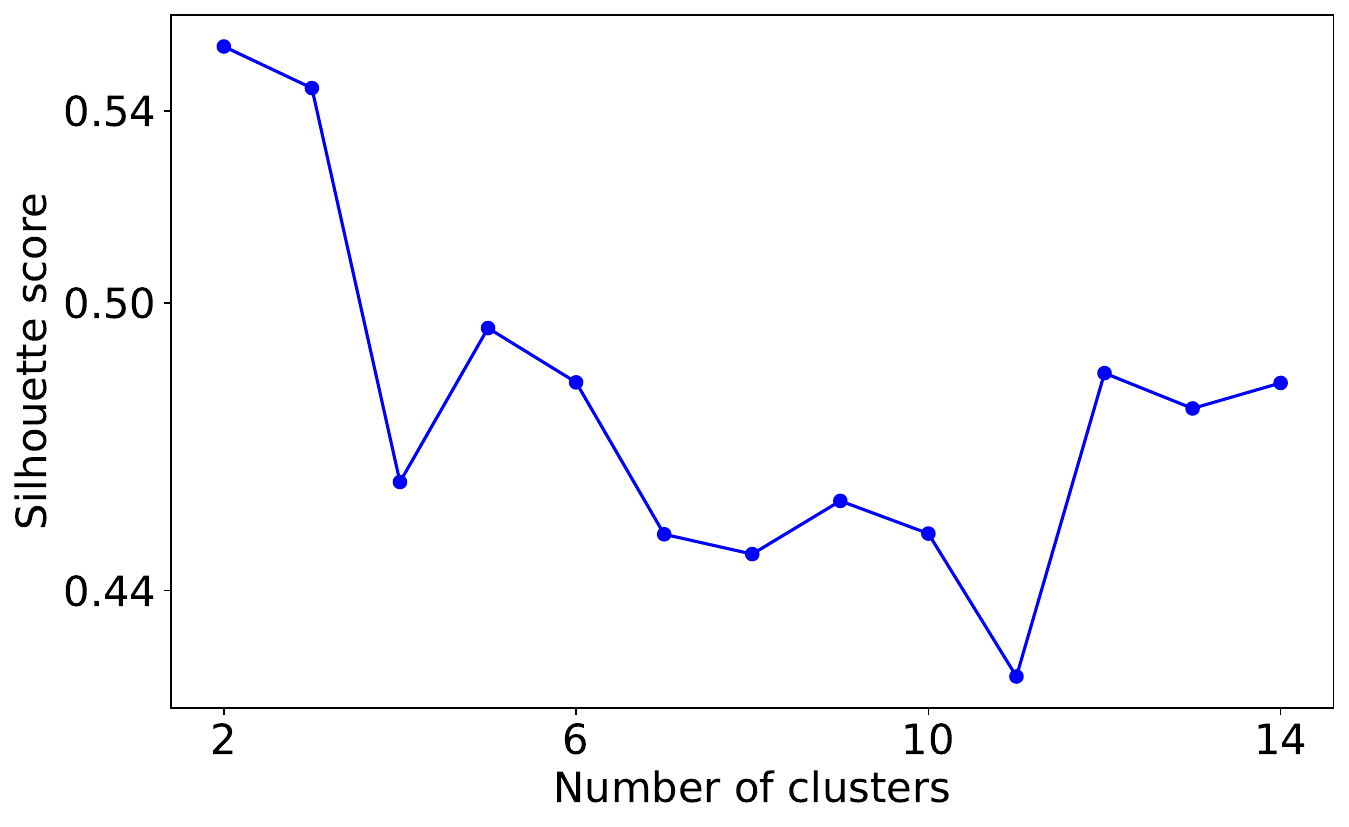}
        \caption{Birds Brown Creeper}
    \end{subfigure}
    \hfill
    \begin{subfigure}[b]{0.22\textwidth}
        \centering
        \includegraphics[width=\linewidth]{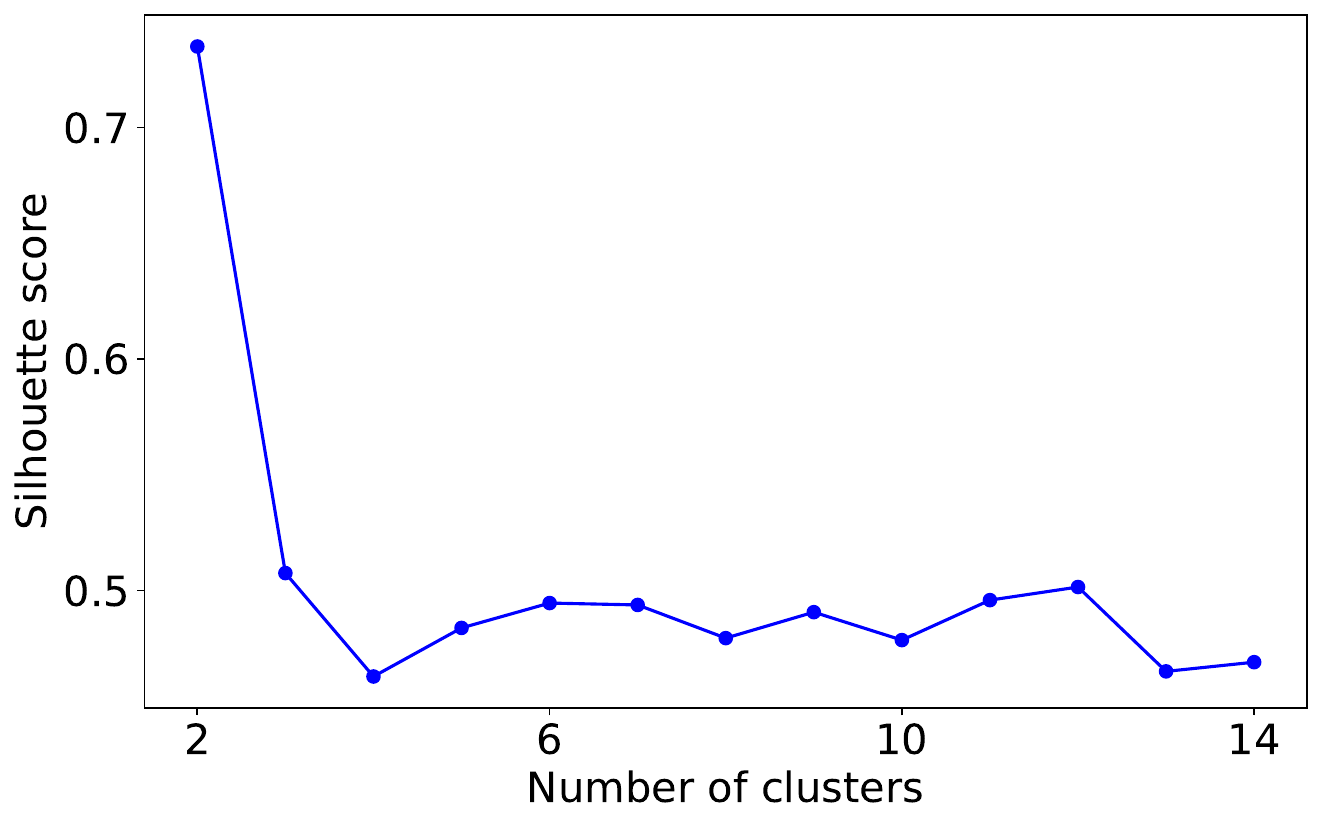}
        \caption{Web Recommend}
    \end{subfigure}
    \hfill
    \begin{subfigure}[b]{0.22\textwidth}
        \centering
        \includegraphics[width=\linewidth]{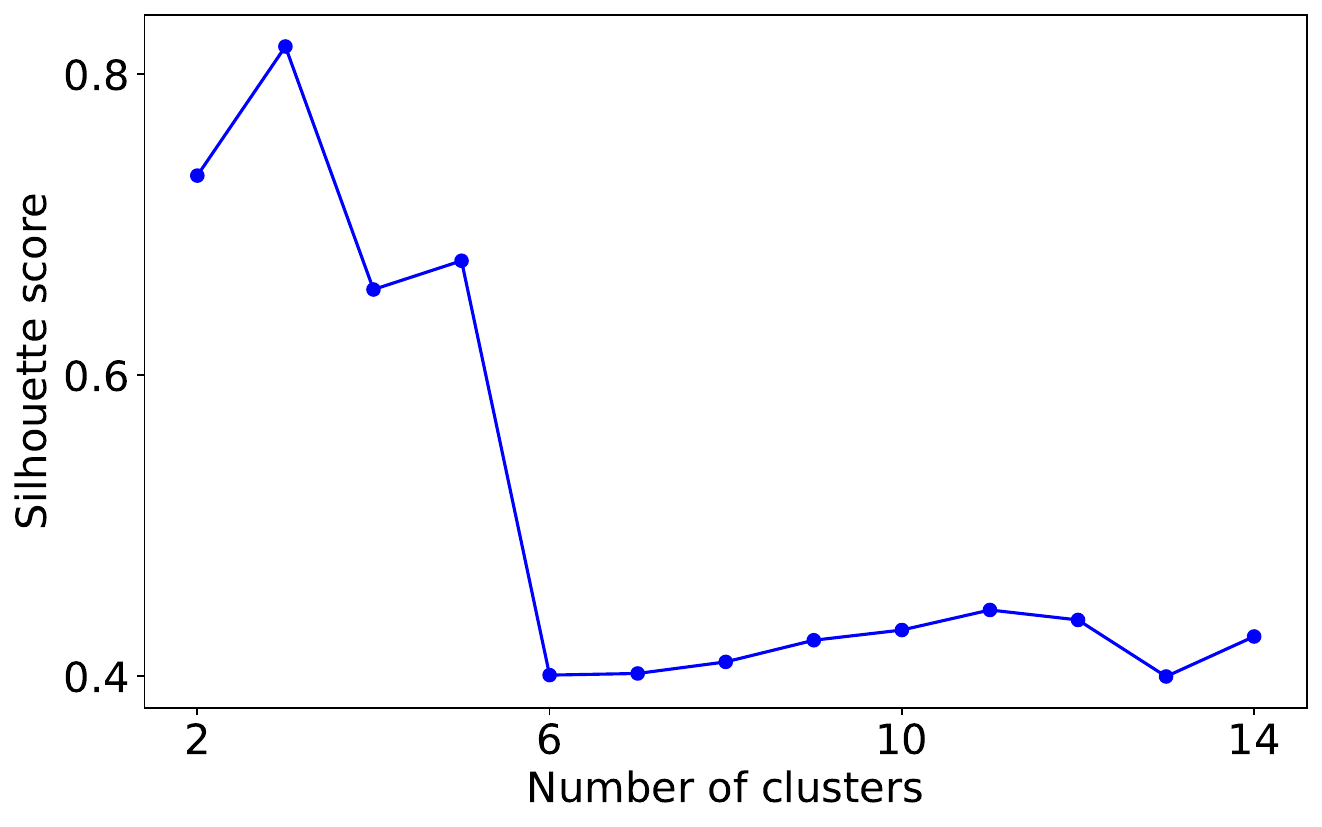}
        \caption{Corel Dogs}
    \end{subfigure}
    \caption{Analysis of the optimal cluster number using the silhouette score method of the k-means clustering algorithm for the VSA embeddings from the training dataset of the traditional benchmark MIL dataset.}
    \label{fig:silhouette_score_all_MIL_datasets}
\end{figure}

 We further work on the visualization of the VSA embeddings. Figure \ref{fig:umap_projection_all_MIL_datasets} demonstrates the VSA embeddings projected into 2D space using UMAP from the training set of the traditional benchmark MIL datasets. Figure \ref{fig:Elbow_method_all_MIL_datasets} depicts the optimal cluster number of the VSA embeddings using the elbow method of the k-means clustering algorithm from the training set of the traditional benchmark MIL datasets. Similar work is repeated using the silhouette score method of the k-means clustering algorithm for the VSA embeddings from the training set of the traditional benchmark MIL dataset in Figure \ref{fig:silhouette_score_all_MIL_datasets}.

 Analysis of the optimal cluster number using the silhouette score method of the k-means clustering algorithm for the VSA embeddings from the training dataset of the traditional benchmark MIL dataset.

Table \ref{table:mean_VSA_MIL_benchmark} describes the statistical distribution of the VSA vectors collected from the training and test sets of traditional MIL benchmark datasets. The mean and absolute mean are close to the target measures constrained by the HLB VSA properties. Here, the standard deviation along with the mean indicates the distribution and fluctuation of the VSA vectors compared to the target ones.  

\begin{table}[!h]
\caption{Mean and absolute mean of the VSA vectors from the training and test sets for traditional MIL benchmark datasets are given where target mean and target absolute mean are 0 and 0.5 respectively}
\label{table:mean_VSA_MIL_benchmark}
\centering
\adjustbox{max width=\columnwidth}{%
\begin{tabular}{lccccc}
\toprule
Dataset & Absolute & Absolute & Mean(Train) & Mean(Test) \\

 & Mean(Train) & Mean(Test) &  &  \\
\midrule

Protein & 0.318 $\pm$ 0.508 & 0.309 $\pm$ 0.487 & 0.037 $\pm$ 0.598 & 0.041 $\pm$ 0.575 \\
MUSK1 & 0.715 $\pm$ 1.194 & 0.674 $\pm$ 1.065 & 0.171 $\pm$ 1.381 & 0.086 $\pm$ 1.258 \\  
MUSK2 & 0.994 $\pm$ 1.193 & 0.955 $\pm$ 1.187 & -0.008 $\pm$ 1.553 & -0.024 $\pm$ 1.523 \\  
UCSB Breast Cancer & 0.656 $\pm$ 0.500 & 0.746 $\pm$ 1.164 & 0.039 $\pm$ 0.540 & 0.047 $\pm$ 1.214 \\ 
Birds brown creeper & 0.465 $\pm$ 0.529 & 0.483 $\pm$ 0.559 & -0.013 $\pm$ 0.704 & -0.011 $\pm$ 0.739 \\

\bottomrule
\end{tabular}
}
\end{table}

We make a detailed comparison of accuracy and AUC score between our approach and different baseline MIL methods for the traditional benchmark MIL datasets in Table \ref{table:cmp_baseline_acc_MIL_benchmark} and \ref{table:cmp_baseline_auroc_MIL_benchmark}. Here, valid MIL models are in the upper half and non-valid MIL models are in the lower half (Valid MIL models are those models that have passed the Algorithm 1 test \cite{raff_reproducibility_2023}). We can see that our VSA-MIL approach performs best in terms of accuracy and AUC score among all valid MIL models.

\begin{table}[!h]
\caption{Comparison of accuracy between different methods for the traditional benchmark MIL datasets}
\label{table:cmp_baseline_acc_MIL_benchmark}
\centering
\adjustbox{max width=\columnwidth}{%
\begin{tabular}{lcccccccc}
\toprule
Baseline & Elephant & Protein & MUSK1 & MUSK2 & UCSB & Birds & Web & Corel \\
 &  &  &  &  & Breast & Brown & Recommendation & Dogs \\
 &  &  &  &  & Cancer & Creeper & 1 &  \\
\midrule

mi-SVM & 0.825 & 0.871 & 0.789 & 0.666 & 0.666 & 0.918 & 0.800 & 0.930  \\
MI-SVM & 0.775 & 0.829 & 0.578 & 0.809 & 0.614 & 0.842 & 0.866 & 0.930  \\
SIL & 0.775 & 0.871 & 0.736 & 0.761 & 0.500 & 0.520 & 0.571 & 0.930     \\ 
MICA & 0.575 & 0.500 & 0.578 & 0.523 & 0.500 & 0.879 & \textbf{0.933} & 0.930    \\
MissSVM & 0.775 & 0.685 & 0.736 & 0.761 & 0.500 & 0.863 & 0.785 & 0.930  \\
mi-Net & 0.825 & 0.871 & 0.736 & 0.761 & 0.833 & 0.909 & 0.466 & 0.930  \\
CausalMIL & 0.810 & 0.872 & 0.737 & 0.800 & 0.667 & 0.918 & 0.866 & 0.930 \\
VSA-MIL & \textbf{0.900} & \textbf{0.948} & \textbf{0.894} & \textbf{0.952} & \textbf{0.833} & \textbf{0.945} & \textbf{0.933} & 0.930 \\

\midrule

NSK & 0.850 & 0.871 & 0.842 & 0.761 & 0.685 & 0.945 & 0.785 & 0.930     \\
MI-Net & 0.775 & 0.897 & 0.894 & 0.857 & 0.833 & 0.927 & 0.600 & 0.930  \\ 
MIL-Pool & 0.800 & 0.974 & 0.789 & 0.954 & 0.833 & 0.927 & 0.933 & 0.937  \\
Tran-MIL & 0.850 & 0.794 & 0.894 & 0.857 & 0.916 & 0.672 & 0.666 & 0.930  \\

\bottomrule
\end{tabular}
}
\end{table}

\begin{table}[!h]
\caption{Comparison of auroc score between different methods for the traditional benchmark MIL datasets}
\label{table:cmp_baseline_auroc_MIL_benchmark}
\centering
\adjustbox{max width=\columnwidth}{%
\begin{tabular}{lcccccccc}
\toprule
Baseline & Elephant & Protein & MUSK1 & MUSK2 & UCSB & Birds & Web & Corel \\
 &  &  &  &  & Breast & Brown & Recommendation & Dogs \\
 &  &  &  &  & Cancer & Creeper & 1 &  \\
\midrule

mi-SVM & 0.878 & 0.773 & 0.727 & 0.982 & 0.657 & \textbf{0.987} & 0.733 & 0.850  \\
MI-SVM & 0.929 & 0.708 & 0.591 & 0.791 & 0.714 & 0.945 & 0.785 & 0.831  \\
SIL & 0.906 & 0.694 & 0.807 & 0.791 & 0.685 & 0.932 & 0.357 & 0.873     \\ 
MICA & 0.906 & 0.811 & 0.750 & 0.927 & 0.714 & 0.968 & 0.785 & 0.815    \\
MissSVM & 0.906 & 0.792 & 0.806 & 0.791 & 0.685 & 0.955 & 0.785 & 0.829  \\
mi-Net & 0.866 & 0.829 & 0.852 & 0.918 & 0.885 & 0.971 & 0.500 & 0.927  \\
CausalMIL & 0.881 & 0.741 & 0.807 & 0.872 & 0.743 & 0.964 & 0.927 & 0.836 \\
VSA-MIL & \textbf{0.954} & \textbf{0.988} & \textbf{0.965} & \textbf{1.000} & \textbf{0.942} & 0.984 & \textbf{0.964} & \textbf{0.941} \\

\midrule

NSK & 0.909 & 0.762 & 0.818 & 0.881 & 0.685 & 0.987 & 0.785 & 0.801     \\
MI-Net & 0.833 & 0.758 & 0.886 & 1.000 & 0.885 & 0.978 & 0.642 & 0.930  \\
MIL-Pool & 0.873 & 0.994 & 0.954 & 1.000 & 0.771 & 0.978 & 0.928 & 0.952  \\
Tran-MIL & 0.866 & 0.794 & 0.965 & 0.900 & 0.942 & 0.829 & 0.714 & 0.591  \\

\bottomrule
\end{tabular}
}
\end{table}

\end{document}